\newif\iffigs\figstrue
\documentclass[10pt,a4paper]{article}
\usepackage{latexsym,amssymb,lscape,graphics,setspace}
\usepackage{amsmath}
\usepackage{amsthm}
\usepackage{graphicx}        
\usepackage{longtable}
\usepackage[vcentermath]{youngtab}
\usepackage{multirow}
\usepackage{booktabs}
\usepackage{multirow}
\usepackage{color}
\usepackage{slashed,epsfig}
\usepackage{amsfonts}
\usepackage{cite}
\usepackage{verbatim}
\usepackage{colortbl}
\usepackage[table]{xcolor}
\usepackage{hyperref}       
\usepackage{array}
\usepackage{fancyhdr}
\usepackage{tikz}
\usepackage{multicol}
\usepackage{tikz-cd}
\newcommand{\mathsym}[1]{{}}

\newtheorem{definizione}{Definition}[section]
\newtheorem{teorema}{Theorem}[section]

\newtheorem{statement}{Statement}[section]

\newtheorem{lemma}{Lemma}[section]
\newtheorem{prooflem}{Proof}[lemma]

\newtheorem*{remark}{Remark}
\newcommand{\bd}{\begin{definizione}}
\newcommand{\ed}{\end{definizione}}

\def\IC{\relax\,\hbox{$\inbar\kern-.3em{\rm C}$}}
\def\IG{\relax\,\hbox{$\inbar\kern-.3em{\rm G}$}}
\def\IB{\relax{\rm I\kern-.18em B}}
\def\ID{\relax{\rm I\kern-.18em D}}
\def\IL{\relax{\rm I\kern-.18em L}}
\def\IF{\relax{\rm I\kern-.18em F}}
\def\IH{\relax{\rm I\kern-.18em H}}
\def\II{\relax{\rm I\kern-.17em I}}
\def\IN{\relax{\rm I\kern-.18em N}}
\def\IP{\relax{\rm I\kern-.18em P}}
\def\IQ{\relax\,\hbox{$\inbar\kern-.3em{\rm Q}$}}
\def\bfzero{\relax\,\hbox{$\inbar\kern-.3em{\rm 0}$}}
\def\IK{\relax{\rm I\kern-.18em K}}
\def\IG{\relax\,\hbox{$\inbar\kern-.3em{\rm G}$}}
 \font\cmss=cmss10 \font\cmsss=cmss10 at 7pt
\def\IR{\relax{\rm I\kern-.18em R}}
\def\ZZ{\relax\ifmmode\mathchoice
{\hbox{\cmss Z\kern-.4em Z}}{\hbox{\cmss Z\kern-.4em Z}}
{\lower.9pt\hbox{\cmsss Z\kern-.4em Z}} {\lower1.2pt\hbox{\cmsss
Z\kern-.4em Z}}\else{\cmss Z\kern-.4em Z}\fi}
\def\bfone{\relax{\rm 1\kern-.35em 1}}

\def\Solv{\mathop{\rm Solv}\nolimits}

\def\inbar{\vrule height1.5ex width.4pt depth0pt}
\def\bfzero{\relax{\rm I\kern-.18em 0}}
\def\bfone{\relax{\rm 1\kern-.35em 1}}

\DeclareFontFamily{U}{rsf}{} \DeclareFontShape{U}{rsf}{m}{n}{
  <5> <6> rsfs5 <7> <8> <9> rsfs7 <10-> rsfs10}{}
\DeclareMathAlphabet\Scr{U}{rsf}{m}{n}

\renewcommand{\baselinestretch}{1.0}
\newcommand{\ft}[2]{{\textstyle\frac{#1}{#2}}}
\def\tilde{\widetilde}

\def\1bar{1\hskip -.275cm -}
\def\2bar{2\hskip -.275cm -}
\def\3bar{3\hskip -.275cm -}

\newsavebox{\uuunit}
\sbox{\uuunit}
                 {\setlength{\unitlength}{0.825em}
                      \begin{picture}(0.6,0.7)
                                      \thinlines
                                      \put(0,0){\line(1,0){0.5}}
                                      \put(0.15,0){\line(0,1){0.7}}
                                      \put(0.35,0){\line(0,1){0.8}}
                                     \multiput(0.3,0.8)(-0.04,-0.02){10}{\rule{0.5pt}{0.5pt}}
                      \end {picture}}

\makeatletter \@addtoreset{equation}{section} \makeatother

\def\bfone{\relax{\rm 1\kern-.35em 1}}

\def\bfone{\relax{\rm 1\kern-.35em 1}}
\font\cmss=cmss10 \font\cmsss=cmss10 at 7pt

\newcommand{\so}{\mathfrak{so}}

\newcommand{\slal}{\mathfrak{sl}}

\begin{document}
\begin{titlepage}
\begin{center}
\vskip 0.2cm
{{\large {\sc  Navigation through Non-Compact Symmetric Spaces: \\
\vskip 0.2 cm  a mathematical perspective on Cartan  Neural Networks  ${}^\dagger$
}} }\\
 \vskip 1cm {\sc Pietro Giuseppe Fr\'e\,$^{a,c}$, Federico 
Milanesio\,$^{b,e}$, Guido Sanguinetti\, $^{d}$,  Matteo Santoro\,$^{d}$} \vskip 0.5cm 
\smallskip
{\sl \small \frenchspacing
${}^a\,$ {\tt Emeritus Professor of} ${}^b\,$Dipartimento di Fisica, Universit\`a di Torino, Via P. Giuria 1, I-10125 Torino, Italy \\[2pt]
${}^{c}\,${\tt Consultant of } Additati\&Partners Consulting s.r.l, Via Filippo Pacini 36, I-51100 Pistoia, Italy \\[2pt]
${}^d\,$ SISSA (Scuola Internazionale Superiore di Studi Avanzati), Via Bonomea 265, I-34136 Trieste, Italy \\[2pt]
${}^e\,$INFN, Sezione di Torino\\[2pt]
E-mail:  {\tt pietro.fre@unito.it, federico.milanesio@unito.it, gsanguin@sissa.it, msantoro@sissa.it,\\
} } \vskip 2cm 
\begin{abstract}  
Recent work has identified \textit{non-compact 
symmetric spaces} $\mathrm{U/H}$ as a promising class of homogeneous manifolds to develop a geometrically consistent theory of neural networks. An initial implementation of these concepts has been presented in a twin paper under the moniker of Cartan Neural Networks, showing both the feasibility and the performance of these geometric concepts in a machine learning context. The current paper expands on the mathematical structures underpinning Cartan Neural Networks, detailing the geometric properties of the layers and how the maps between layers interact with such structures to make Cartan Neural Networks covariant and geometrically interpretable. Together, these twin papers constitute a first step towards a fully geometrically interpretable theory of neural networks exploiting group-theoretic structures.
\end{abstract}
\vfill
\end{center}
\noindent \parbox{175mm}{\hrulefill} 
\par
${}^\dagger$ P.G. Fr\'e acknowledges  support  by the Company \textit{Additati\&Partners Consulting s.r.l} during all the time of development of 
the present research. Furthermore, the Ph.D. fellowships of F. Milanesio and M. Santoro are cofinanced by the Company \textit{Additati\&Partners 
Consulting s.r.l} at Torino University and SISSA, respectively. 
\\[5pt]
\end{titlepage}
{\small \tableofcontents} \noindent {}
\newpage
\section{Introduction}\label{intro}
Machine Learning algorithms have mostly utilized affine Euclidean spaces $\mathbb{E}^n$ as the geometrical environment where data points are transformed and grouped. The natural choice of maps between such spaces is the affine space homomorphism, but, since such maps are linear, to create a sufficiently rich function class, it is necessary to add point-wise activation functions \cite{dubnaexpress}, non-linearities applied to each entry of the data vector.
The indisputable success of classical neural networks conceived in this way 
should not obscure the fact that, because of the ad-hoc introduction of pointwise activation functions, the geometric 
interpretability of the {learned} network parameters is lost, along with geometric properties of the networks (e.g. the equivariance of single homomorphism layers).

This problem, which is intrinsic when we define deep networks using Euclidean geometry, motivates the investigation of the remaining, vast playground provided by the non-compact symmetric spaces. Non-compact symmetric spaces raise more hopes of unfolding new perspectives since homomorphisms between such spaces are non-linear. These Riemannian spaces $(\mathcal{M},g)$  present a dual nature since the manifold $\mathcal{M}$ supporting the metric $g$ can be alternatively viewed as a solvable group manifold $\mathcal{S}$ or as a coset manifold $\mathrm{U/H}$, so that maps pertaining to each of the two algebraic structures pertaining to the same differentiable manifold  can be successively applied in order to increase the expressivity of the network. Indeed our goal is to show how, adopting the Paint group - Tits - Satake framework (PGTS theory) developed in \cite{pgtstheory}, one can construct functioning Neural 
Networks free from point-wise activation functions that are universally substituted with a sequence of operations that amount to a 
\textbf{homomorphism} between two solvable Lie Groups in the map from one layer to the next one, followed by \textbf{isometry} transformations of 
each layer into itself.

\subsection{Symmetric spaces as a natural framework}\label{sse:choiceofsymmetricspaces}
As mentioned, the notion of the traditionally used Euclidean affine spaces $\mathbb{E}^n$ is too restrictive for our aims. While it is possible to equip such spaces with both a Riemannian and group structure, the function class of isometries and homomorphisms is still the same, namely that of affine transformations. The choice of \textbf{non-compact symmetric spaces} as the appropriate mathematical framework for the modeling of neural network hidden layers 
\footnote{As we will explain, the input layer and the output layer are an exception since the first 
map, the injection, is the linear map from the datum vector to the coordinates of the first layer that is a symmetric space $\mathrm{U_1/H_1}$, 
while the last layer can be either the value of a function or a section of a bundle over the last layer $\mathrm{U_N/H_N}$ or a probability measure 
on a chamber decomposition of the last hidden layer, as it happens in the classification task.} was highlighted in \cite{pgtstheory}, and is based 
on the following principles.
\begin{enumerate}
  \item \textbf{The spaces on which we operate must be differentiable, Riemannian manifolds}.
  \item \textbf{It should be possible to embed isometrically any Riemannian manifold in our chosen manifold class}, provided the dimension is high enough (similarly to Nash's embedding theorem).
  This general requirement follows from the hidden manifold hypothesis, and that the initial \textbf{injection map} $\iota$ of the hidden manifold 
 can always match relative distances.
 \item \textbf{Each manifold $(\mathcal{M}_n,g)$ should be a Cartan-Hadamard manifold.} By definition, such Riemannian manifolds are simply connected, geodesically complete, and have a \textit{nowhere positive} sectional curvature, not necessarily constant. Thanks to a theorem named after \textit{Hopf-Rinow},  any Cartan-Hadamard manifold admits a unique well-defined distance function $d(p_1,p_2)$ between any pair of its points $p_{1,2}$, provided by the length of the unique geodesic arc that joins them. This critical and unique property is the very reason why one must restrict one's attention to Cartan-Hadamard manifolds in Machine Learning. Furthermore, because of another theorem due to Cartan-Hadamard themselves (see, for instance, \cite{Helgasonobook}), the manifolds named after them are diffeomorphic to $\mathbb{R}^n$ and constitute the natural generalization of $\mathbb{R}^n$ from the point of view of Riemannian geometry.
\item \textbf{Each $\mathcal{M}_n$ of the class should admit a translation group with free isometric transitive action on $\mathcal{M}_n$}, just as it happens for flat Euclidean space. All points of $\mathcal{M}_n$ should be equivalent so that only relative locations are relevant,  as many machine learning algorithms aim at grouping points according to their similarities and differences, 
 which are always conceived in terms of relative distance.
\end{enumerate} 

The properties above single out, apart from the infinite family of $n$-dimensional Euclidean spaces $\mathbb{E}^n$, only the families of 
 \textit{non-compact symmetric spaces $\mathrm{U/H}$ where $\mathrm{U}$ is a simple Lie group}. These spaces,  classified by the monumental work of Èlie Cartan, via Cartan involution, are in one-to-one correspondence with the 
 non-compact real sections of complex simple Lie Algebras, on their turn classified first by the pioneering work of Killing \cite{kalingo}, later 
 refined in Cartan's doctoral thesis \cite{cartatedesca} and finally encoded in the classification of Dynkin diagrams (see \cite{historicbook} for the historical details of the simple Lie algebra classification and the classical textbook \cite{Helgasonobook} for a complete
 description of symmetric spaces and of their Cartan classification).  

 The key mathematical result which allows the use of non-compact symmetric spaces in our construction is the \textbf{metric equivalence} of all non-compact symmetric spaces 
$\mathrm{U/H}$ with an appropriate \textbf{solvable Lie subgroup} $\mathrm{\mathcal{S}_{U/H}}\subset \mathrm{U}$\footnote{At first sight the universal existence of such solvable 
subgroup may seem miraculous, yet it is just the generalization to all non-compact simple Lie groups of the well known Borel subgroup of maximally 
split simple Lie groups (see \cite{pgtstheory} for details). Notation-wise, we point out that when we talk about a generic Lie group we name it 
$\mathrm{G}$ and we denote $\mathbb{G}$ its Lie algebra. When the Lie group has a Lie algebra which is a non-compact real section of a 
(semi)-simple complex Lie algebra we name it $\mathrm{U}$ and the corresponding  Lie algebra $\mathbb{U}$. The reason is historical. The modern 
mathematical theory of non-compact symmetric spaces $\mathrm{U/H}$ was mostly developed within Supergravity Theory where such spaces occur as the 
scalar manifold containing all scalar fields of the specifically considered Supergravity Lagrangian. The isometry group $\mathrm{U}$ is realized 
as a group of electric-magnetic duality transformations \cite{mylecture} which unifies the so-named target space dualities with the 
non-perturbative strong-weak dualities of string theory. Because of that unification, it is traditionally named $\mathrm{U}$.} (see Theorem \ref{thm:metricequivalence} and 
\cite{pgtstheory} for a thorough theoretical discussion). This equivalence on one side matches the mathematical conception introduced by Alekseevsky of \textit{normal Riemannian spaces} 
\cite{Alekseevsky1975,Cortes,Alekseevsky:2003vw}, on the other was instrumental in introducing \textit{solvable coordinates} in Supergravity 
theories that lead to results reviewed in \cite{pgtstheory}, in particular in relation to special geometries.
\par
 \par

\subsection{Previous work}
As we mentioned in \cite{naviga}, in machine learning literature, some of the non-compact symmetric spaces, namely the 
\textit{hyperbolic spaces}:
\begin{equation}\label{lenisco}
  \mathbb{H}^{1+q} \, \equiv \, \dfrac{\mathrm{SO(1,1+q)}}{\mathrm{SO(1+q)}} 
\end{equation}
have already been considered \cite{nickel_poincare_2017, 
ganea_hyperbolic_2018, klimovskaia_poincare_2020, peng_hyperbolic_2022, 
bdeir_fully_2024} for the heuristic motivation that data 
corresponding to tree-structured graphs have an optimal embedding in the manifolds $\mathbb{H}^{1+q}$. In relation to the set of Cartan-Hadamard manifolds in eq. (\ref{lenisco}), it is quite appropriate to point out the following. The Euclidean spaces $\mathbb{E}^{1+q}$
originally considered in ML constructions are also coset manifolds, namely $\mathrm{ISO(1+q)/SO(1+q)}$ and the numerator group
$\mathrm{ISO(1+q)}$, \textit{i.e.} the Euclidean group of motions, is just the Inonu-Wigner contraction of
$\mathrm{SO(1,1+q)}$. Hence, the hyperbolic spaces
(\ref{lenisco}) degenerate into Euclidean spaces in the limit where the negative scalar curvature goes to zero (group-theoretically, the Inonu-Wigner contraction). This is an important point. The absence of an intrinsic non-linearity affecting the classical setup of neural networks and requiring the use of point-wise activation functions is a limiting characteristic of the Euclidean space due to the vanishing of curvature. As soon as the negative curvature of the $\mathrm{U/H}$ symmetric spaces is restored,  the intrinsic non-linearity is retrieved and the point-wise activation functions can be discarded.
Early work on hyperbolic deep learning focused on hyperbolic embeddings for hierarchical data. Nickel et al. \cite{nickel_poincare_2017} introduced \textit{Poincaré embeddings}, showing superior hierarchical representation compared to Euclidean embeddings. Ganea et al. \cite{ganea_hyperbolic_2018} and subsequent works \cite{shimizu_2021, chen-etal-2022-fully, bdeir_fully_2024, peng_hyperbolic_2022}, extended hyperbolic geometry to deep learning by developing \textit{hyperbolic neural networks}, using Möbius operations \cite{ungarGyrovectorSpaceApproach2009}. Various generalizations of hyperbolic networks have been explored. Convolutional networks \cite{skliar_hyperbolic_2023, ghosh2024universalstatisticalconsistencyexpansive}, graph neural networks \cite{chami_2019}, and attention mechanisms \cite{gulcehre2018hyperbolic} hyperbolic variants were introduced to handle different datasets. 

Hyperbolic spaces, depending on the utilized 
coordinate system, obtained various denominations, including the most popular denomination of \textit{Poincaré balls}, but also \textit{Lorentz 
model of Hyperbolic space} or simply \textit{Hyperbolic Space}. 

\par We argue instead that all noncompact symmetric spaces are hyperbolic since they are all Cartan-Hadamard manifolds satisfying the stronger condition that the \textbf{sectional curvature is everywhere negative}. Indeed all classical\footnote{By the wording classical, following Weyl's tradition (see his book Classical Groups) we mean symmetric spaces 
      $\mathrm{G/H}$ where the numerator Group has a Lie algebra $\mathbb{G}_\mathbb{R}$ given by a real section of a simple complex Lie Algebra 
      $\mathbb{G}_\mathbb{C}$ } symmetric spaces $\mathrm{G/H}$, compact and non-compact, are Einstein spaces with respect to their canonical 
      $\mathrm{G}$-invariant metric. Utilizing the intrinsic components of the curvature $2$-form in vielbein formalism:
      \begin{itemize}
      \item the 
  Riemann tensor is made of constants (see for instance eq.(I.3.115) in volume I of \cite{castdauriafre}). 
  \item The Ricci tensor is positive definite for the compact $\mathrm{G/H}$ and it is instead negative definite 
      for all the non-compact ones.
      \item The scalar curvature is constant and positive for compact symmetric spaces, while it is constant and 
      negative for all the non-compact symmetric spaces $\mathrm{U/H}$.
      
      \end{itemize}
      In this sense all the non-compact  symmetric  spaces are 
      \textbf{hyperbolic Riemannian manifolds}\footnote{The \textbf{hyperbolic spaces} (\ref{lenisco}) have the additional property that the Weyl tensor is identically zero and hence they are 
      conformally flat in all dimensions. The particular coordinates utilized in \cite{nickel_poincare_2017, klimovskaia_poincare_2020} for the 
      manifolds (\ref{lenisco}), dubbed by the authors \textit{Poincaré balls}, are the well known projective off-diagonal coordinates for coset 
      manifolds (see eq.s (5.28) and following ones in the classical book \cite{gilmore_lie} and, in particular, the explicit formula (5.2.43) in the 
      second volume of \cite{pietrobook}). }
      , while \textbf{the compact ones are elliptic}. For this reason, Hyperbolic Learning should be utilized as the denomination of all neural networks that are composition of maps between \textit{non 
compact symmetric spaces}.

\par

\subsection{General aims of the PGTS programme}\label{sse:contributions}
The present paper constitutes the second step in the development of the PGTS programme, the first being provided by the foundational paper \cite{pgtstheory}. 

This paper is accompanied by another work by the same authors \cite{naviga} which introduced Cartan neural networks (CaNNs), a novel neural 
network architecture based on group-theoretic concepts. Paper \cite{naviga} was aimed at a machine learning audience, and 
focused on implementation and performance measurements, while the current one adopts the classical theoretical physics style to discuss in depth the 
mathematical foundations of the CaNN construction and the first principles inspiring their formulation. The explicit architectures presented in \cite{naviga}  just correspond to the simplest instance 
(\textit{hyperbolic spaces}) in the much wider class of \textit{non-compact symmetric spaces} $\mathrm{U/H}$.

The basic guidelines of the PGTS programme are: 
\begin{description}
  \item[a)] Removal of the \textit{point-wise activation functions} in order to obtain \text{covariance} with respect to the change of basis in 
      the vector space corresponding to $i$-th layer, that turns out to be the solvable Lie algebra $Solv_{i}$ of the $i$-th solvable Lie group 
      $\mathcal{S}_{i}$. 
  \item[b)] \textit{Geometrical interpretability} of all the learned parameters that just correspond to the parameters determining the $i$th 
      solvable \textit{group homomorphism}. 
  \item[c)] Proper consideration of the natural organization of hyperbolic symmetric spaces into \textit{Tits-Satake universality classes} 
      (see \cite{pgtstheory} for all definitions, concepts, and original literature quotations) and the determination of appropriate procedures 
      to monitor the working 
      of neural networks about the distribution of data in \textit{Paint Group level surfaces} through the learning epochs (see 
      \cite{pgtstheory} for definitions and theoretical explanations about the Paint Group). 
\end{description}

In the present paper, as we sketch in \cite{naviga}, we consider the general issue a), focusing, as a case study, on the already explored case of 
manifolds (\ref{lenisco}) that constitute the Tits-Satake universality class of the Poincaré-Lobachevsky plane 
$\dfrac{\mathrm{SO(1,2)}}{\mathrm{SO(2)}} \sim\dfrac{\mathrm{SL(2,\mathbb{R})}}{\mathrm{SO(2)}}$. The aim is to illustrate the precise 
recipe for the construction of the maps:
\begin{equation}\label{carnevaledirio}
        \mathfrak{h}_{i}(W) \quad : \quad \mathcal{S}_{i} \, \longrightarrow \, \mathcal{S}_{i+1}
      \end{equation} 
mapping one layer into the next, recipe that has an immediate generalization to the case of other Tits-Satake universality classes. 

\section{Critical Analysis of Classical neural networks}
\label{criticollo}

In the previous section, we specified the goal of the present paper and briefly mentioned the obtained results. Before presenting the mathematical details of our architectures proposed and tested in \cite{naviga}, we  summarize the basics of classical neural networks described in a more mathematical abstract formulation. This is purposely done to highlight the critical issue of the \textit{point-wise activation 
functions} and emphasize their conflict with covariance and geometrical interpretability, as well as introducing the reader not already familiar with machine learning terminology to its basics.

\subsection{The perceptron}
\label{precettore} 

\begin{figure}[h!]
 \begin{center}
\includegraphics[width=9cm]{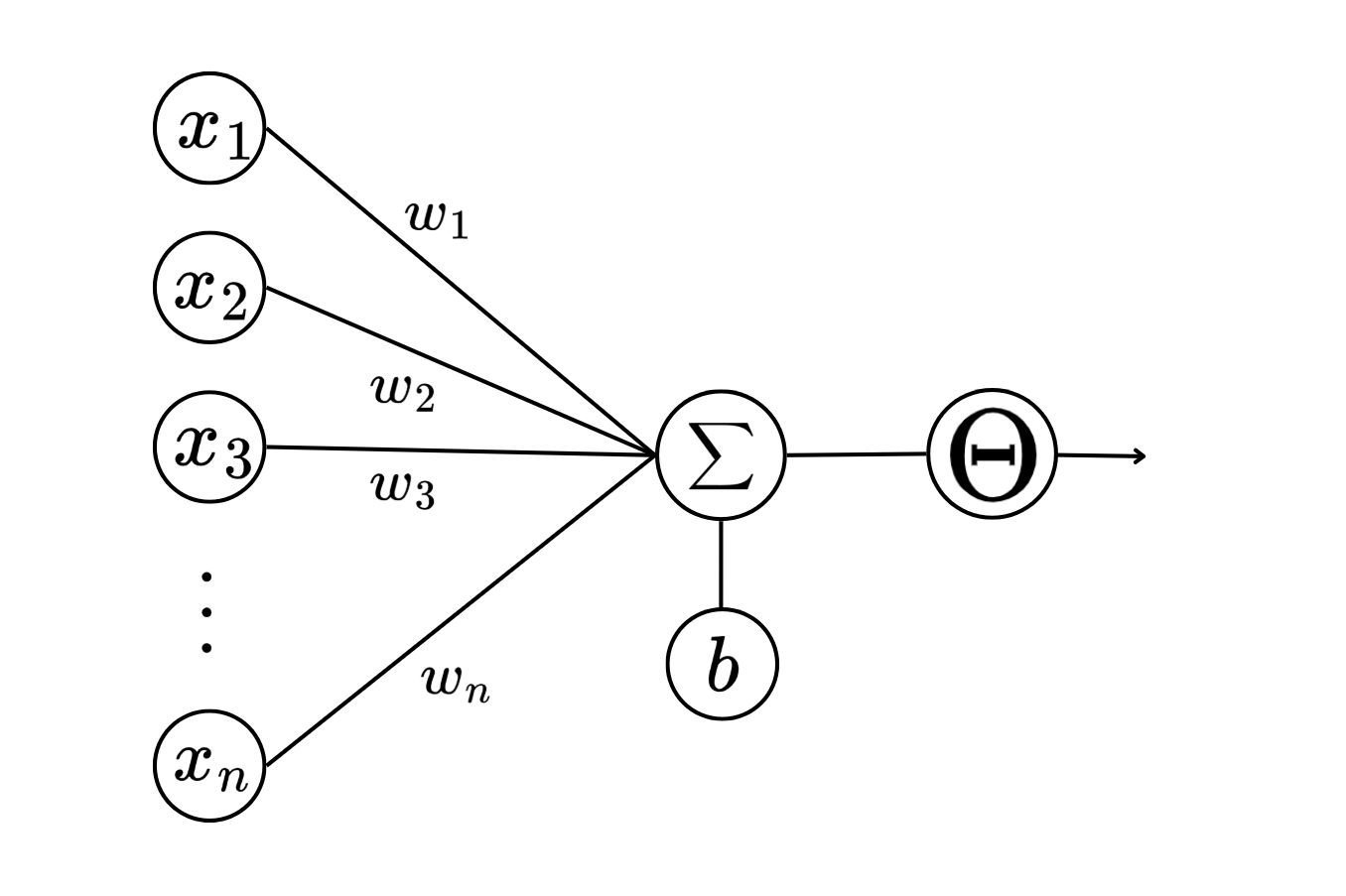}
 \caption{\label{percettone} 
  Simplified conceptual diagram of a perceptron or neuron. The neuron receives an input composed of a datum (the \textbf{training 
 set}) $\mathbf{x}$, and a pair of weight $\mathbf{w}$ and bias $b$ in order to elaborate that datum.} 
\end{center}
\end{figure}

The denomination of the components of a neural network (fig. \ref{mlpreti}) has historical reasons,  tracing back to the seminal work of McCulloch and Pitts \cite{mcculloch_logical_1943}, who in 1943  introduced the notion of a mathematical atomic computational unit simulating the 
functionality of a biological neuron, later known as \textit{artificial neuron} or \textit{perceptron}. The latter specifically refers to the first algorithm used to train an artificial neuron for a classification task, introduced by Rosenblatt in 1958 \cite{cimicerosa}. In the original idea of Rosenblatt, a perceptron would be realized concretely with electronic circuits, but it is also an abstract mathematical object consisting of an algorithm that processes some input 
data and returns more output data: it is the atomic unit of the learning process. The conceptual diagram of a \textit{perceptron} is shown in fig. \ref{percettone}. 

\par
In its basic version, the perceptron has an input formed by a vector of data 
\begin{equation}\label{inputdat}
 \mathbf{x}=\left\{x_1,x_2,\dots,x_n\right \}\in\mathbb{R}^n
\end{equation}
which has to be classified into one of two classes, labeled 0 and 1. This kind of problem is called \textbf{binary classification}, and, together with regression, where each data point has to be associated with a scalar value, is the most common task in supervised machine learning.
The perception is a scalar function of the \textbf{weighted average} of such input data 
\begin{equation}\label{returndata}
  f(\mathbf{x})=F\left(\langle \mathbf{w}\mid\mathbf{x} \rangle\right)
\end{equation}
where $\langle \mathbf{w}\mid\mathbf{x} \rangle$ denotes the usual Euclidean scalar product in the $n$-dimensional vector space $\mathbb{R}^n$:
\begin{equation}\label{scalproduct}
  \langle \mathbf{w}\mid\mathbf{x} \rangle \equiv \sum_{i=1}^n\, w_i\, x_i  
\end{equation}
of the vector $\mathbf{x}$ with a vector $\mathbf{w}\in\mathbb{R}^n$. The original idea was that the perception could emphasize or de-emphasize certain features, with the emphasis being decided by the explicit form of the weight vector $\mathbf{w}$. Correspondingly the simplest form chosen for the scalar function $F$, mentioned in (\ref{returndata}) was the 
following:  
\begin{equation}\label{gromellina}
 F\left(\langle \mathbf{w}\mid\mathbf{x} \rangle\right)\, =\, \Theta\left(\langle \mathbf{w}\mid\mathbf{x} \rangle\, - \, b\right) \, 
\end{equation}
where $ b >0$ and $\Theta(\cdot)$ is the well-known \textit{Heaviside step function}, which is equal to zero for negative values of its argument $y$ and equal to one for positive values of the same. The interpretation of (\ref{gromellina}) is rather obvious. It implies that the neuron \textit{fires} its output signal if the filtered input signal $\langle \mathbf{w}\mid\mathbf{x} \rangle$ is larger than a threshold $b$, also known as \textbf{bias}. In this case, firing corresponds to predicting class 1 for the data point, and not firing class 0. The function $F(x)$ can be chosen in different ways according to the intended purposes.

The training of a perception proceeds through a supervised learning algorithm. Given a dataset of labeled input-output pairs $\{\mathbf{x}_{k}, \,y_{k}\}_{k=1}^P$, where $y_{k}\in\{0,1\}$ is the target output for input $\mathbf{x}_{k}$, the algorithm updates the weights according to the perceptron learning rule:

\begin{equation}
\begin{aligned}
\mathbf{w}_{t+1} &\leftarrow \mathbf{w}_t + \eta \left( y_{k} - f(\mathbf{x}_{k}) \right) \mathbf{x}_{k} \\
b_{t+1} &\leftarrow b_t - \eta \left( y_{k} - f(\mathbf{x}_{k}) \right)
\end{aligned}
\end{equation}

where $\eta > 0$ is the learning rate and $\mathbf{w}_0$ is randomly initialized. This rule adjusts the weight vector in the direction that reduces the error between the actual output and the desired output. If the perceptron’s prediction is correct, the weights remain unchanged; otherwise, they are nudged toward producing the correct result in future iterations. This update is repeated for each sample in the training set, often across multiple passes (epochs), until convergence is achieved, either the perceptron correctly classifies all training samples, or a predefined stopping criterion is met.

\subsection{The multilayer perceptron, namely the classical neural networks} 

\begin{figure}
 \begin{center}
\includegraphics[width=9cm]{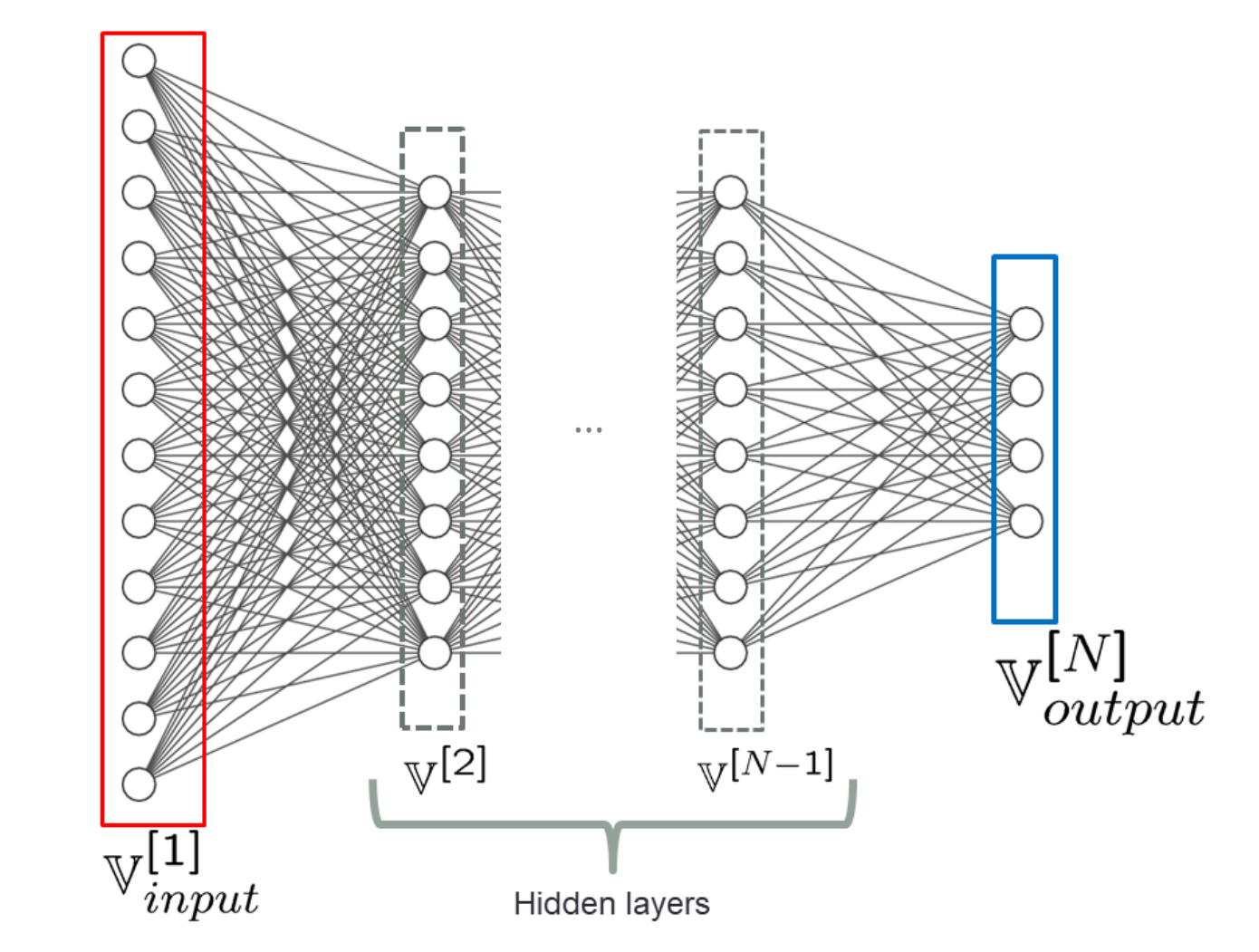}
 \caption{\label{mlpreti} Conceptual scheme of the Multi-Layer Perceptron, original name 
 for what was later named a Deep Neural Network, or simply a Neural Network, with various qualifications. }
\end{center}
\end{figure}

 The perceptron, as discussed in the previous section, represents the most elementary model of a neuron: it receives an input vector, computes a weighted sum of its components, applies a non-linear activation function, and produces a scalar output. Despite its simplicity and historical importance, the perceptron had a fundamental limitation: it could only solve linearly separable problems. The term Multi-Layer Perceptron (MLP) (see fig. \ref{mlpreti}), later known as \textit{neural network} (NN), refers to an architecture composed of several layers of interconnected neurons. Although neural networks originated trying to imitate the functioning of a biological brain, from a mathematical standpoint, they are just 
sophisticated maps from an \textbf{input space}, to an \textbf{output space}, where the complexity lies in the large number of parameters that have to be set through what in machine learning jargon is called \textbf{learning}\footnote{Once again the iterative redefinition of the parameters, named learning, is assimilated to the neuroplasticity of the brain.}.

The general operation of a multi-layer neural network can be viewed as a sequence of transformations:

\begin{equation}
\mathbf{x}^{(0)} \rightarrow \mathbf{x}^{(1)} \rightarrow \mathbf{x}^{(2)} \rightarrow \cdots \rightarrow \mathbf{x}^{(N)} \rightarrow\hat y
\end{equation}

where $\mathbf{x}^{(0)}$ is the data corresponding to a point, and each intermediate vector $\mathbf{x}^{(i)}$ is computed through a function similar in spirit to the perceptron’s transformation, an affine map followed by a non-linearity.

\begin{equation}
 \mathbf{x}^{(\ell)} = \phi^{(\ell)}\left(W^{(\ell)} \mathbf{x}^{(\ell-1)}+\mathbf{b}^{(\ell)}\right)
\end{equation}

where $W^{(\ell)}$ is a matrix of learnable weights, $\mathbf{b}^{(\ell)}$ is a bias vector, and $\phi^{(\ell)}$ is a (typically non-linear) activation function applied component-wise. Popular activation functions are summarized below:
 
\begin{table}[htb]
 \centering
\begin{tabular}{|c|c|}
\hline
  \textbf{Name}
& \textbf{Given by}
\\ \hline
  sigmoid / logistic
& $\frac{1}{1+\exp(-x)}$
\\ \hline
  tanh
& $\tanh(x)$
\\ \hline
  rectified linear unit (ReLU)
& $\max\{0,x\}$
\\ \hline
  $a$-leaky ReLU
& $\max\{ax,x\}$ for some $a \geq 0$, $a \neq 1$
\\ \hline
  exponential linear unit
& $x\cdot \chi_{x\geq 0}(x)+ (\exp(x)-1)\cdot \chi_{x<0}(x)$
\\ \hline
\end{tabular}
\caption{Commonly-used activation functions }
 \label{azionetavoletta}
\end{table}

The final layer representation $\mathbf{x}^{(N)}$ is then transformed into the output $\hat{y}$. This mapping is given by:

\begin{equation}
 \hat{y} = \sigma( W^{(\text{out})}\mathbf{x}^{(N)} \, + \, \textbf{b}^{(\text{out})})
\end{equation}

where $\sigma$ is usually either the sigmoid or the identity function, depending on the task.

Before discussing the mathematical backbones of this map, we will briefly summarize how a multi-layer neural network is trained. Given a dataset of pairs $\{\mathbf{x}_k,\,y_k\}_{k=1}^P$, we define a \textbf{loss function} (or cost function) $\mathcal{L}$ that quantifies the discrepancy between the network's predictions and the true target outputs over the training dataset. A common choice in regression tasks is the mean squared error (MSE):

\begin{equation}
\mathcal{L}(\boldsymbol{\theta}) = \frac{1}{P} \sum_{k=1}^P \left\| \hat y_{k} - y_{k} \right\|^2,
\end{equation}

where $P$ is the number of training samples. Calling $\boldsymbol{\theta}$ the set of learnable parameters $\{W^{(\ell)},\mathbf{b}^{(\ell)}\}_{i=1}^N$, the goal of training is to find the optimal parameter vector $\boldsymbol{\theta}^\ast$ that minimizes $\mathcal{L}$:

\begin{equation}
\boldsymbol{\theta}^\ast = \arg\min_{\boldsymbol{\theta}} \, \mathcal{L}(\boldsymbol{\theta}).
\end{equation}

To solve this optimization problem, one typically employs gradient descent, an iterative method that updates the parameters in the direction opposite to the gradient of the loss:

\begin{equation}
\boldsymbol{\theta}_{t+1} = \boldsymbol{\theta}_t - \eta \, \nabla_{\boldsymbol{\theta}} \mathcal{L}(\boldsymbol{\theta}_t),
\end{equation}

where $\eta > $ is the learning rate, a hyperparameter that controls the step size. In practice, especially for large datasets, one uses \textbf{stochastic gradient descent (SGD)}, where the gradient is estimated using a small batch of samples rather than the entire training set, reducing computational cost and often improving generalization.

The maps used in the construction of classical neural networks are affine maps, and by definition, affine maps act on a vector space. Summarizing, we can say that in classical neural networks, each layer corresponds to a real vector space, and each neuron represents a basis vector of that space. Formally we have a sequence $n=0,\dots,N$ of real vector spaces of variable dimension \begin{eqnarray}\label{sequenzavectspace}
&&\mathbb{V}^{[n]} \quad : \quad \text{dim} \, \mathbb{V}^{[n]} \, = \, d^{[n]}
\end{eqnarray} 
and in each space, we have the basis vectors 
\begin{eqnarray}\label{sequenzavecdim}
&&\mathbf{e}_{[n]}^i \quad ; \quad i=1,\dots , d^{[n]}
\end{eqnarray}
The input for all neurons in layer $n+1$ is the same, namely a vector $\mathbf{x}^{[n]}$ of the previous space: 
\begin{eqnarray}\label{vectorVn}
&&\mathbf{x}^{[n]} \, \in \, \mathbb{V}^{[n]} \quad ; \quad \mathbf{x}^{[n]} \, = \, \sum_{j=1}^{d^{[n]}} \, x^{[n]}_i \, \mathbf{e}_{[n]}^i 
\end{eqnarray}

Now, each layer in a neural network can be conceptually divided into two stages: an affine transformation followed by a non-linear activation. The affine transformation is given by:

\begin{equation}
\begin{split}
A^{(n)}&:\mathbb{V}^{[n-1]} \to \mathbb{V}^{[n]}\\  A^{(n)}\left(\mathbf{x}\right) & = \, a_j \, \mathbf{e}^j_{[n]}  \quad;\quad  \forall \mathbf{x} \,=\, x_i \mathbf{e}^i_{[n-1]} 
  \in \mathbb{V}^{[n-1]}\\
  a_j & =\left(W^{[n]}\right)_{j}^{\phantom{j}i} \,x_i + b_j^{[n]}
\end{split}
\end{equation}

while the non-linear map $\phi$ has the following form: 
\begin{equation}
\begin{split}\label{sigmusorror}
  &\phi^{[n]}  : \, \mathbb{V}^{[n]} \, \longrightarrow \, \mathbb{V}^{[n]}\\
  &\forall \mathbf{z} \, = \, z_i  \mathbf{e}^i_{[n]} \quad \quad \phi(\mathbf{z} ) = \varphi(z_i) \, \mathbf{e}^i_{[n]} 
\end{split}
\end{equation}

where common choices for the function $\varphi \, : \, \mathbb{R} \, \to \, \mathbb{R}$ are described in tab. \ref{azionetavoletta}.

Now, each layer can be thought of as the map:

\begin{equation}\label{platowillo}
\begin{split}
\mathcal{K}^{[n]}_{\left[{W}^{[n]},\mathbf{b}^{[n]}\right]} &:  \mathbb{V}^{[n-1]}\, \rightarrow \, \mathbb{V}^{[n]}\\
\mathcal{K}^{[n]}_{\left[{W}^{[n]},\mathbf{b}^{[n]}\right]}& \,\equiv \phi^{[n]}\circ A^{[n]}
\end{split}
\end{equation}

And we have that the overall map from the input space $\mathbb{V}^{[0]}$ to the last hidden layer $\mathbb{V}^{[N]}$, realized as follows: 

\begin{equation}\label{cospirazione}
\begin{split}
\mathcal{U}_{[\pmb{\mathfrak{\mathfrak{W}}},\boldsymbol{\mathfrak{b}}]}&: \mathbb{V}^{[0]} \, \to \, \mathbb{V}^{[N]}\\
\mathcal{U}_{[\pmb{\mathfrak{\mathfrak{W}}},\boldsymbol{\mathfrak{b}}]}& =   \mathcal{K}^{[N]}_{\left[{W}^{[N]},\mathbf{b}^{[N]}\right]}\circ \dots   
\circ
\mathcal{K}^{[2]}_{\left[{W}^{[2]},\mathbf{b}^{[2]}\right]} \circ 
\mathcal{K}^{[1]}_{\left[{W}^{[1]},\mathbf{b}^{[1]}\right]}\\
\mathcal{K}^{[n]}_{\left[W^{[n]},\mathbf{b}_{n}\right]} & =  
  \phi^{[n]}\circ A^{[n]}
\end{split}
\end{equation}

where, by $\pmb{\mathfrak{\mathfrak{W}}}$ we have denoted the collection of all rectangular matrices $W^{[n]}$   and 
by $\boldsymbol{\mathfrak{b}}$ we have denoted the collection of all bias vectors $\mathbf{b}_{[n]}$. 

Mathematically, the definition of the non-linear map (\ref{sigmusorror}) is anomalous, independently of the choice 
of $\varphi$, since it is not covariant with respect to a change of basis in the vector space $\mathbb{V}^{[n]}$. The definition of $\phi$ makes sense only on a specific basis, and there is no a priori understanding of how such a basis is singled out. Furthermore, the list of functions 
displayed in table \ref{azionetavoletta} has no intrinsic characterization of any algebraic, analytic, or functional space nature. It is just a 
collection of  "ad hoc" objects whose justification is merely given by their usefulness and efficiency in numerical simulations.  We will see how 
in the new geometric setup, where the maps from one layer to the next are a homomorphism as in eq.(\ref{carnevaledirio}), this \textbf{severe conceptual problem} could be overcome from the very start. 

We postpone to section \ref{cicero} a detailed description of the classification task,  focusing now on the regression task where:

\begin{equation}
\begin{split}
    l_{[W^{(\text{out})},\, \textbf{b}^{(\text{out})}]}&:\quad\mathbb{V}^{[N]}\to\mathbb{R}^{n_{\text{out}}}\\
    l_{[W^{(\text{out})},\, \textbf{b}^{(\text{out})}]}& =W^{(\text{out})}\mathbf{x}^{(N)} \, + \, \textbf{b}^{(\text{out})}
\end{split}
\end{equation}

with $n_{\text{out}}$ being the number of predictions.

The entire network is then a function:

\begin{equation}
\begin{split}
    f&:\mathbb{V}^{[0]}\to\mathbb{R}^{n_{\text{out}}}\\
    f& = l_{[W^{(\text{out})},\, \textbf{b}^{(\text{out})}]}\circ\mathcal{U}_{[\pmb{\mathfrak{\mathfrak{W}}},\boldsymbol{\mathfrak{b}}]}
\end{split}
\end{equation}

\subsection{The universal approximator issue and the activation functions} 
Before turning to the next topics, let us briefly discuss the \textit{universal approximator issue}. By the end of the 1980s, the neurophysiological viewpoint on neural networks 
had already silently glided away, and the scientists working on them regarded their target of study in the same way we presented it in 
eq.(\ref{cospirazione}), namely the construction of a complicated map from an input space to an output one. In particular, if the input space 
$\mathcal{M}^0=K$ is a compact subregion of the standard $n$-dimensional Euclidean space, $K \subset \mathbb{V}_0$, where $\mathbb{V}_0 := \mathbb{R}^n$, for instance the 
hypercube $[0,1]^n$, and if the output space is just $\mathbb{R}$, then the overall network can be regarded as the construction of a real function 
of $n$ real variables: $f(\mathbf{x}) \, = \,f(x_1,\dots,x_n)$. In particular, reducing the architecture to a single hidden layer, the multivariate real 
function $f(\mathbf{x})$ takes the following form: 
\begin{equation}\label{carnevalata}
  f\left(\mathbf{x} \mid W,\mathbf{b},\mathbf{w} \right)\, = 
  \, \sum_{i=1}^N \, w^i \,\sigma
  \left(\,\sum_{a=1}^{n} \,W_i^a\, x_a \, + \, b_i\right)
\end{equation}
where $\sigma(\cdot)$ is the non-linear activation function, typically the sigmoid mentioned in table \ref{azionetavoletta} and discussed both in 
section \ref{cicero} and in appendix \ref{historsigma} within the different context of logistic regression. Equipping the space $\mathcal{C}(K)$ 
of differentiable functions on $K$ with the uniform convergence squared distance 
\begin{equation}\label{uniformconv}
  \forall f(\mathbf{x}),\,g(\mathbf{x}) \in \mathcal{C}(K) \quad ; \quad \| f-g \|^2 \, \equiv \, 
  \underbrace{\text{sup}}_{\mathbf{x}\in K}|f(\mathbf{x})-g(\mathbf{x})|^2 
\end{equation}
there was an extended literature around the late 1980s and early 1990s, of which we reference a few of the principal representatives
\cite{Cybenko1989,CHUI1992131,FUNAHASHI1989,hornik1991approximation,carroll1989construction}, aimed at showing 
that the set of functions $f\left(\mathbf{x} \mid W,\mathbf{b},\mathbf{w}\right)$ defined in eq. (\ref{carnevalata}) is dense in $\mathcal{C}(K)$, with a fixed choice of $\sigma(\cdot)$. The positive results obtained in such a setup are named the \textbf{universal approximation theorems} and contributed to a 
radical change of attitude towards the activation functions. The first approximation theorems were obtained for the sigmoid activation \cite{Cybenko1989}. Subsequently, the conditions on $\sigma(\cdot)$ were progressively relaxed: in a first step, it was just requested that 
$\sigma(\cdot)$ should be monotonic and bounded. Then a further relaxation occurred with the result of \cite{PinkusUniversalApproximation} stating 
that any non-linear, non-polynomial function might realize this universal approximation. The quoted studies concentrated on so-called \textbf{shallow networks} where there is only one hidden layer and the parameterized ansatz for the unknown function is precisely of the form of eq. (\ref{carnevalata}). In due time, the investigations were extended to multi-layer networks and more general multi-component output functions (see the review \cite{dubnaexpress}), yet the general idea is captured 
sufficiently by equation (\ref{carnevalata}) to allow a conceptual discussion at its level. 
While universal approximation theorems are existence theorems, and thus do not guarantee that an algorithm will reach a solution, they still show why pointwise nonlinearities have been important in the functioning of neural networks. Thus, in our framework, such theorems should be revised in terms of the injection map and of a suitable covariant description of the space of functions on the last layer. Such a study is in our 
agenda for the near future. 

\section{Maps between symmetric spaces}
In Section \ref{intro}, we have summarized the general conception of classical neural networks, emphasizing the critical role played by the  
\textit{point-wise activation functions}. Their final mathematical structure is that encoded in eq.(\ref{cospirazione}). In this paper, we describe the alternative architecture, named by us \textit{fixed $r$ Cartan neural network}, obtained as follows:
\begin{equation}\label{ago}
  \mathcal{M}_0 \, \,\underbrace{\stackrel{\iota}{\hookrightarrow}}_{\text{injection}} \,\, \underbrace{\mathcal{M}^{[1]}\,
  \stackrel{\mathcal{U}_{[\pmb{\mathfrak{\mathfrak{W}}},\boldsymbol{\Psi}]}}{\longrightarrow}
  \mathcal{M}^{[N]}}_{\text{hidden layers}}\,\,
  \rightarrow \underbrace{\mathcal{M}_{out}}_{\mathrm{output\:space}}
\end{equation}
where $\mathcal{U}_{[\pmb{\mathfrak{\mathfrak{W}}},\boldsymbol{\Psi}]}$ denotes the composition of several maps from one layer to the next one, as 
follows: 
\begin{eqnarray}\label{fisarmonica}
\mathcal{U}_{[\pmb{\mathfrak{\mathfrak{W}}},\boldsymbol{\Psi}]}&:& \mathcal{M}^{[r,q_0]} \, \to \, \mathcal{M}^{[r,q_N]} \nonumber \\
\mathcal{U}_{[\pmb{\mathfrak{\mathfrak{W}}},\boldsymbol{\Psi} ]}& = &  \hat{\mathcal{K}}^{[N-1]}_{\left[\mathcal{W}^{[N-1]},\Psi_{N}\right]}\circ \dots   
\circ
\hat{\mathcal{K}}^{[2]}_{\left[\mathcal{W}^{[2]},\Psi_{3}\right]} \circ 
\hat{\mathcal{K}}^{[1]}_{\left[\mathcal{W}^{[1]},\Psi_{2}\right]}\nonumber\\
\hat{\mathcal{K}}^{[n]}_{\left[\mathcal{W}^{[n]},\Psi_{n+1}\right]} & : & \mathcal{M}^{[r,q_n]} \, \longrightarrow \, \mathcal{M}^{[r,q_{n+1}]}
\end{eqnarray} 
The essential difference from classical neural networks is that each layer is, instead of a vector space, a member of the same Tits-Satake universality class (see \cite{pgtstheory} and Equation \ref{Mrq} further down) selected by 
the choice a definite non-compact rank $r_{n.c.}\, = \, r$. It is important to stress immediately that using as layers  manifolds of the same TS universality class is not forced. Indeed, as we explain later on in the main text and illustrate in appendix \ref{patagonia} with some toy examples, we can also include geometrical maps that move from 
a TS class to another one, thus providing additional expressivity. We name all such architectures \textit{Cartan neural networks}, so that those specified in eq.(\ref{fisarmonica}) are a distinguished subclass of  Cartan neural networks, those at fixed $r$ and those based on Hyperbolic Spaces, treated in the twin paper \cite{naviga} are the sub-subclass corresponding to $r=1$.
\par
This being duly emphasized, we first clarify the geometrical properties of the layers, and then we  discuss the nature of the parameters 
$\mathcal{W}^{[n]},\Psi_{n+1}$ determining the map $\hat{\mathcal{K}}^{[n]}_{\left[\mathcal{W}^{[n]},\Psi_{n+1}\right]}$ from the $n$-th layer to next.  
\par

\subsection{Preliminaries and notation}
We recall here some preliminary notions and results, extracted for reader's convenience from textbooks and from \cite{pgtstheory}. 

\begin{teorema}[Classification of simple complex Lie algebras \cite{humphreys_semisimple_1972,pietro_discrete}]
Every simple complex algebra is either a member of the infinite series
 $\mathfrak{a}_l, \mathfrak{b}_l, \mathfrak{c}_l, \mathfrak{d}_l$ or one of 
the exceptional Lie  algebras $\mathfrak{e}_6$, $\mathfrak{e}_7$, $\mathfrak{e}_8$, $\mathfrak{f}_4$, $\mathfrak{g}_2$. 
\end{teorema}

\begin{remark}
The simple algebras $\mathfrak{b}_l,  \mathfrak{d}_l,\mathfrak{c}_l$ correspond to the complex orthogonal, respectively  odd and even dimensional, 
algebras $\mathfrak{so}(2l+1, \mathbb{C})$, $\mathfrak{so}(2l, \mathbb{C})$, and to the symplectic algebra $\mathfrak{sp}(2l, \mathbb{C})$. 
Because of that, the fundamental representation of these three Lie Algebras is given by the set of matrices that are either $\eta$-antisymmetric 
in the orthogonal case 
$$A \eta+\eta A^T = 0$$
where $\eta=\eta^T$ is any symmetric non degenerate matrix   or $C$-symmetric in the symplectic case 
$$A C+C A^T = 0$$ 
where $C= - C^T$ is any antisymmetric non degenerate matrix. 
\end{remark}

\begin{definizione}[Def. 1.1, \cite{pgtstheory}]\label{def:normedalgebra}
    A \emph{normed solvable Lie algebra} is a pair $(\mathfrak{g}, \langle \:, \: \rangle)$ where $\mathfrak{g}$ is a finite-dimensional solvable 
    Lie algebra and $\langle \:, \: \rangle$ is a positive definite symmetric quadratic form satisfying invariance with respect 
    to the adjoint action, i.e.
    $$\langle [Z,X], Y\rangle + \langle X, [Z,Y]\rangle=0 \quad \forall X,Y,Z\in \mathfrak{g}$$
\end{definizione}

\begin{teorema}[Th. 1.1, \cite{pgtstheory}]\label{thm:metricequivalence}
Let $\mathrm{U}$ be a simple non compact finite-dimensional Lie group, denote by $\mathbb{U}$ its Lie algebra. Then  $\mathbb{U}$ is necessarily a 
non-maximally compact real section of the simple Lie algebra $\mathbb{U}_\mathbb{C}$ obtained as complexification of  $\mathbb{U}$. Let 
$\mathrm{H}\subset \mathrm{U}$ be the unique (up to conjugation) maximal compact subgroup, whose  Lie Algebra we denote $\mathbb{H}$. Then the 
coset manifold $\mathrm{U/H}$ is necessarily a symmetric space endowed with a unique (up to homothety) $\mathrm{U}$-invariant Einstein metric. 
Furthermore the Riemannian manifold $\left(\mathrm{U/H},g_{Einst}\right)$ is metrically equivalent to a solvable group manifold 
$\mathcal{S}_{\mathrm{U/H}} \subset \mathrm{U}$. 
\end{teorema} 

\begin{definizione}[See \cite{pgtstheory}]\label{alekshomo}
    A Riemannian manifold $(\mathcal{M},g)$ is metrically equivalent to an Alekseevskian \textbf{normal solvable group} $\mathcal{S}$ if
    \begin{itemize}
      \item $\mathcal{S}$ has a \textbf{free transitive action} on $\mathcal{M}$ that is an isometry for $g$. In this way $\mathcal{M}$ is 
          diffeomorphic to $\mathcal{S}$. 
      \item There exists on the solvable Lie algebra $Solv$ of the solvable group $\mathcal{S}$ a positive definite symmetric quadratic form 
          $\langle \:, \: \rangle$ whose $\mathcal{S}$ transport from the origin to any point of $\mathcal{M}\sim \mathcal{S}$ coincides with 
          the metric $g$.  
    \end{itemize}     
\end{definizione}
\par
\begin{remark}
    With the assumptions of Theorem \ref{thm:metricequivalence}, the metric is just the one induced from 
    the unique $\mathrm{U}$-invariant Einstein metric on $\mathrm{U/H}$ (see formula 1.6 of \cite{pgtstheory}.
\end{remark}
Metric equivalence can be extensively used  to construct relevant maps in group theoretical terms. 
\par
Following the notations and the definitions of \cite{pgtstheory} we consider the TS universality classes of pseudo-orthogonal type, namely: 
\begin{equation}\label{Mrq}
  \mathcal{M}^{[r,q]} \, \equiv \, \frac{\mathrm{SO(r,r+q)}}{\mathrm{SO(r)} \times\mathrm{SO(r+q)}}
\end{equation}  
the hyperbolic spaces (\ref{lenisco}) being the case $r=1$ in the above list. As extensively illustrated in \cite{pgtstheory}, \textbf{for each 
fixed integer value of $r$, the corresponding infinite family of manifolds obtained by varying $q\in \mathbb{N}$ in eq.(\ref{Mrq}) constitutes a 
Tits-Satake universality class}, in the sense that the Tits-Satake projection $\pi_{TS}$,  explained in \cite{pgtstheory}, projects every manifold 
of the class onto the same and unique maximally split submanifold (see also the papers \cite{Fr__2013,bacchihollititti} and the book 
\cite{advancio})\footnote{ Actually, as we will also explain in the forthcoming paper \cite{tassella}, the entire universality class constitutes a 
sequence of vector bundles on the same base manifold, namely the Tits-Satake $\mathcal{M}_{TS}$ with \textbf{structural group} the corresponding 
\textbf{subPaint Group} (see \cite{pgtstheory} for the explanation of the subPaint group). The concept of Paint and subPaint groups was discovered 
and developed in the years 2006-2007 \cite{fre_cosmic,tittusnostro}. 
\par
Indeed relying on the \textbf{PGTS theory} exposed in \cite{pgtstheory} we can consider the Tits-Satake projection:
\begin{equation}\label{proietta}
  \pi_{TS} \quad : \quad \mathcal{M}^{[r,q]} \, \longrightarrow \, \mathcal{M}^{[r,1]}
\end{equation}
The fibers $\pi_{TS}^{-1}(p)$ over each point $p\in \mathcal{M}^{[r,1]}$ of the base manifold have the structure of a $r\times (q-1)$ vector space 
$\mathbb{V}^{(r\mid q-1)}$ that corresponds to the carrier space for the representation $(r\mid q-1)$ of the subgroup $\mathrm{SO(r) \times 
SO(q-1)}\subset \mathrm{SO(r,r+q)}$. Correspondingly, the original manifold $\mathcal{M}^{[r,q]}$ can be viewed as the total manifold 
\begin{equation}\label{totalspazio}
  \text{tot}\left[\mathcal{E}^{[r(q-1)]}\right] \, = \, \mathcal{M}^{[r,q]}
\end{equation}
of a vector bundle:
\begin{equation}\label{carneadtwo}
  \mathcal{E}^{[r(q-1)]} \, \stackrel{\pi_{TS}}{{ \longrightarrow}}\, \mathcal{M}^{[r,1]}
\end{equation}
that has the Tits-Satake projection as projection, the vector $\mathbb{V}^{(r\mid q-1)}$ as standard fiber:
\begin{equation}\label{carneadthree}
  \mathrm{F}  \, = \, \mathbb{V}^{(r\mid q-1)}
\end{equation}
and the subpaint group $\mathrm{G_{subPaint}} \, = \, \mathrm{SO(q-1)}$, multiplied by its normalizator $\mathrm{SO(r)}$ in $\mathrm{H_c}$, as 
structural group: 
\begin{equation}\label{carneadstruc}
  \mathrm{G_{struc}} \, = \, \mathrm{SO(r)\times SO(q-1)}
\end{equation}
\par
As one sees, in the $r=1$ case of Hyperbolic Spaces $\mathbb{H}^p$, the normalizer $\mathrm{SO(1)}$ does not exist and the structural group is 
just the subPaint group as stated above.    
}
\begin{equation}\label{tittosatakko}
  \mathcal{M}_{TS} \, = \, \mathcal{M}^{[r,1]} \, \equiv \, \frac{\mathrm{SO(r,r+1)}}{\mathrm{SO(r)} \times\mathrm{SO(r+1)}}
\end{equation}
Finally, we recall a well-known
\begin{lemma}[Theorem 5.6 of \cite{hall_lie_2015}]
\label{omomorfista} Let $G, G'$ be simply connected matrix Lie groups and $\mathbb{G}, \mathbb{G}'$ their Lie algebras. Let $h: \mathbb{G} \rightarrow \mathbb{G}'$ be a Lie algebra homomorphism. Then there exists a unique group homomorphism $\tilde{h}: G \rightarrow G'$ such that

\[
\begin{tikzcd}
\mathbb{G} \arrow[r, "h"] \arrow[d, "\mathrm{exp}_\mathbb{G}"'] & \mathbb{G}' \arrow[d, "\mathrm{exp}_{\mathbb{G}'}"] \\
G \arrow[r, "\tilde{h}"'] & G'
\end{tikzcd}
\]

commutes.
\end{lemma}
\begin{remark}
As the group homomorphism $\tilde{h}$ is in general nonlinear, it might be hard to express it analytically. However, as the spaces $\mathcal{M}^{[r,s]}$ are solvable, the general expression of Lie algebra homomorphisms can be computed iteratively for high non-compact rank cases, as we show in the examples of Appendix \ref{patagonia}.
\end{remark}

\subsection{The nested embedding of non compact symmetric spaces \texorpdfstring{$\mathrm{U/H}$}{U/H}}
\label{generalemapposo} The set of non-compact symmetric spaces $\mathrm{U/H}$ has been shown in \cite{pgtstheory} and Section \ref{sse:choiceofsymmetricspaces} to be the natural candidate for new neural network architectures. In these spaces, the Lie algebra $\mathbb{U}$ of the numerator 
group is a non-compact real section of a complex simple Lie algebra and the Lie algebra of the denominator group is the maximal compact subalgebra $\mathbb{H}\subset \mathbb{U}$ of the former. The multifaceted theory of these manifolds, including their general \textbf{metric equivalence} with a solvable Lie subgroup $\mathcal{S}\subset \mathrm{U}$, was assembled and systematically exposed in \cite{pgtstheory}, in which we prepared the mathematical foundations of the PGTS programme stated in Section \ref{sse:contributions}. We recall some results from \cite{pgtstheory} which are particularly relevant for our constructions. 
\subsubsection{The general triangular embedding of Lie algebras \texorpdfstring{$\mathbb{U}$}{U}}
\par
To properly parameterize the manifolds of interest, we recall the hierarchical embedding of groups and Lie algebras described in  
\cite{pgtstheory}, of which we reproduce Statement 3.1. For simplicity we assume $q$ to be even. The odd case is completely analogous with obvious 
changes. 
\begin{statement}\label{statamento}
Let $\mathrm{N}$ be the real dimension of the fundamental representation of the Lie Algebra $\mathbb{U}$ of
the simple non compact Lie group $\mathrm{U}$ and $\mathbb{H}\subset \mathbb{U} $ its maximal
compact subalgebra. Moreover, let $\mathbb{U}$ be the real section of either $\mathfrak{b}_l, \mathfrak{c}_l$ or $\mathfrak{d}_l$. 
Then there exists a suitable matrix $\eta_t$ with the proper signature and a canonical embedding:
\begin{eqnarray}
 \mathbb{ U} & \hookrightarrow & \slal(\mathrm{N},\mathbb{R})~\,\nonumber\\
\mathbb{ U} \, \supset \, \mathbb{H} & \hookrightarrow &
\so\mathrm{(N)} \, \subset \,
\slal(\mathrm{N},\mathbb{R})~. \label{eq:cantriaembed}
\end{eqnarray}
such that the image of $\Solv\left(\mathrm{U/H} \right)$ is made of upper triangular
matrices.
Let $\mathbb{K}$ be the orthogonal complement of $\mathbb{H}$ in $\mathbb{U}$. Then 
\begin{eqnarray}\label{symeta}
    \forall \, \mathrm{K}\, \in \, \mathbb{K} & : & \quad \, \mathrm{K} \,- \,  \mathrm{K}^T  \, = \, 0\nonumber\\
    \forall \, \mathrm{H}\, \in \, \mathbb{H} & : & \quad \, \mathrm{H} \, + \,   \mathrm{H}^T   \, = \, 0
\end{eqnarray}
\end{statement}

\begin{remark}
The condition (\ref{symeta})  guarantees that $\mathbb{H}$ is mapped to $\so(\mathrm{N})$  while $\mathbb{K}$ is mapped instead to the
orthogonal complement of $\so(\mathrm{N})$ in $\slal(\mathrm{N},\mathbb{R})$.

\end{remark}
\par
A basis of generators for the $\so(r,r+2s)$ Lie Algebra, where the above abstractly described structure is made explicit, is obtained through the 
construction in dimension $\mathrm{N}=2r+2s$ of the matrix $\eta_t$ advocated by the statement \ref{statamento}. By definition of the Lie Algebra, 
the matrix $\eta_t$ must have signature: 
\begin{equation}\label{pincuspallus}
 \text{sign}_{r,s} \, = \, \left\{\underbrace{+,\dots, +}_{r+2s \text{ times}},\underbrace{-,\dots,-}_{r \text{ - times}} \right\}
\end{equation}
The Lie algebra $\so(r,r+2s)$ is then made by all $N\times N$ matrices that fulfill the following condition:
\begin{equation}\label{gornayalavandaia}
  X \in \so(r,r+2s) \quad \Leftrightarrow \quad  X^T \, \eta_t \, + \, \eta_t \, X \, = \, 0
\end{equation}
The appropriate choice of $\eta_t$ is the following one:
\begin{equation}\label{etatdefi}
  \eta \, = \, \eta_t \, = \, \left(
\begin{array}{ccc|c|ccc}
 0 & \cdots & 0 & 0 & 0 & \ldots & 1 \\
 \vdots & \vdots & \vdots & \vdots & \vdots & 1 & \vdots \\
 0 & \ldots & 0 & 0 & 1 & \ldots & 0 \\
 \hline
 0 & \ldots & 0 & \mathbf{1}_{2s \times 2s} & 0 & \ldots & 0  \\
 \hline
 0 & \ldots & 1 & 0 & 0 & \ldots & 0 \\
 \vdots & 1 & \vdots & \vdots & \vdots & \vdots & \vdots \\
 1 & \cdots & 0 & 0 & 0 & \cdots & 0 \\
\end{array}
\right)
\end{equation}
The matrix $\eta_t$ defined above is transformed into the standard one:
\begin{equation}\label{etabdefi}
  \eta_b \, \equiv \, \text{diag}\left\{\underbrace{+,\dots, +}_{r+2s \text{ times}},\underbrace{-,\dots,-}_{r \text{ - times}} \right\}
\end{equation}
by conjugation, so that  for all $r,s$ choices there exists $\Omega$ satisfying (see \cite{pgtstheory} for details):
\begin{equation}\label{etatinetab}
  \Omega \, \eta_t \, \Omega^T \, = \, \eta_b
\end{equation}
Utilizing the basis $\eta_t$, the solvable subalgebra is easily identified since, by construction, it is constituted by all those matrices 
satisfying the defining condition \ref{gornayalavandaia} of the Lie algebra $\so(r,r+2s)$,   which in addition are also upper triangular. Like 
this, the procedure to explicitly construct the algebra becomes straightforward and well-suited for computer algorithms. In that construction one 
automatically derives a set of Cartan-Weyl generators well-adapted to the particularly chosen invariant metric $\eta_t$.  In this basis the Cartan 
involution $\theta$ defining the orthogonal decomposition $\mathbb{U}\, = \, \mathbb{H} \oplus \mathbb{K} $ is  simply given by matrix 
transposition: 
\begin{equation}\label{thetatranspo}
 \forall X \in \so(r,r+2s)\, : \quad  \theta (X) \, = \, X^T
\end{equation}

\subsubsection{Maps defined by adjoint action of the linear group}
The canonical embedding of the Lie algebra $\so(r,r+2s)$ in $\slal(2r+2s,\mathbb{R})$ also implies a canonical embedding of the corresponding 
groups. At the Lie algebra level, we have:
\begin{equation}\label{Kannone}
  Solv\left(\mathcal{M}^{r,q}\right)\, \equiv \,  Solv_{[r,q]} \, \stackrel{{K_1}}{\hookrightarrow} \, \so(r,r+q) \, \stackrel{{K_2}}{\hookrightarrow} \, 
  \slal(2r+q,\mathbb{R})
\end{equation}
which provides a Lie algebra homomorphism between the original solvable Lie algebra and its images through $K_1$ and $K_2\circ K_1$. By means of 
abstract exponentiation, eq.(\ref{Kannone}) induces a canonical embedding of groups:
\begin{equation}\label{obice}
  \mathcal{S}_{[r,q]}\, \stackrel{{K_1}}{\hookrightarrow} \, \mathrm{SO(r,r+q)} \, \stackrel{{K_2}}{\hookrightarrow} \, 
  \mathrm{SL(2r+q,\mathbb{R})}
\end{equation}
In other words equation (\ref{obice}) implies that $\mathcal{S}_{[r,q]}$ is a subgroup of $\mathrm{SO(r,r+q)}$ and a fortiori of 
$\mathrm{SL(2r+q,\mathbb{R})}$.
\par
As we know Lie group conjugation induces the adjoint action of a group on its Lie algebra so that we have:
\begin{equation}\label{aggiunta}
  \forall g \in \mathrm{SL(2r+q,\mathbb{R})} \quad ; \quad \text{Adj}_g \left[\slal (2r+q,\mathbb{R})\right] \, \equiv \, 
  g^{-1} \slal (2r+q,\mathbb{R}) \, g \, = \, \slal (2r+q,\mathbb{R})
\end{equation}
Given any basis $\mathfrak{T}_a$ of the Lie algebra, the adjoint action is linear:
\begin{equation}\label{caramella}
   \forall g \in \mathrm{SL(2r+q,\mathbb{R})}\quad ; \quad \text{Adj}_g \left[\mathfrak{T}_a\right] \, = \, \mathcal{A}_{a}^{\,\,b}(g)\,
   \mathfrak{T}_b 
\end{equation}
One can consider the adjoint action of $\mathrm{SL(2r+q,\mathbb{R})}$ on the solvable subalgebra 
$Solv\left(\mathcal{M}^{r,q}\right)$:
\begin{equation}\label{trafficone}
  \forall g \in \mathrm{SL(2r+q,\mathbb{R})}\quad ; \quad \text{Adj}_g\left[Solv_{[r,q]}\right] \, \equiv \, Solv^{g}_{[r,q]} 
  \subset \slal (2r+q,\mathbb{R})
\end{equation}
For any $g \in \mathrm{SL(2r+q,\mathbb{R})}$, the image $Solv^{g}_{[r,q]}$ of $Solv_{[r,q]}$ through the adjoint map ${\mathrm{Adj}}_g$ is a Lie subalgebra of $\slal (2r+q,\mathbb{R})$ 
isomorphic to its preimage. The same happens with the Lie algebra  $\so(r,r+q)$. This means that we can consider four cases:
\begin{equation}\label{terravuoto}
  g \, \in \,\left\{\begin{array}{l|lcl|l|}
  \hline
  \null&\null&\null&\null&\null \\
                      \mathcal{S}_{[r,q]} & Solv^{g}_{[r,q]} & = & Solv_{[r,q]}& \textit{internal automorphism of $\mathcal{S}_{[r,q]}$} \\
                      \null&\null&\null&\null&\null \\
                      \mathrm{G_{Paint}} \subset \mathrm{SO(r,r+q)} & Solv^g_{[r,q]} & = & Solv_{[r,q]}& \textit{external automorphism of 
                      $\mathcal{S}_{[r,q]}$} \\
                      \null&\null&\null&\null&\null \\
                      \dfrac{\mathrm{SO(r,r+q)}}{\mathrm{G_{Paint}}} & Solv^g_{[r,q]} & \subset  & \so(r,r+q)& \textit{ 
                      Grassmannian rotation } \\
                      \null&\null&\null&\null&\null \\
                       \dfrac{\mathrm{SL(2r+q,\mathbb{R})}}{\mathrm{SO(r,r+q)}} & Solv^g_{[r,q]} & \subset & 
                       \text{Adj}_g\left[\so(r,r+q)\right] & \textit{fully external homomorphism}\\
                       \null&\null&\null&\null&\null \\
                       \hline
                    \end{array}
   \right.
\end{equation}
The adjoint action of the solvable group on itself maps the solvable Lie algebra into itself. It is an internal automorphism. However, 
as explained in \cite{pgtstheory}, within the group $\mathrm{U}$, in our case within $\mathrm{SO(r,r+q)}$, there is a group of external 
automorphisms of the solvable Lie algebra and that is precisely the Paint Group. When the group element $g$ belongs to 
$\mathrm{U}=\mathrm{SO(r,r+q)}$ but not to the Paint group, the image of the solvable Lie algebra is a different isomorphic copy of the same 
inside the isometry group $\mathrm{U}$. Finally when the group element $g$ does not belong to the subgroup $\mathrm{SO(r,r+q)} \subset 
\mathrm{SL(2r+q,\mathbb{R})}$ the image of the solvable Lie algebra $Solv_{[r,s]}$ is completely external. It is on the basis of the last case 
that we can construct non-trivial maps from one layer to the next. 

When building maps between layers, we will alternate \textbf{isometries}, i.e. the first three cases of Equation \ref{terravuoto}, that is $g\in \mathrm{SO}(r,r+q)$, and fully \textbf{general homomorphisms}, which can be considered an extension of the last case $g\in \mathrm{SL}(2r+q, \mathbb{R})$ to targets with different dimensions.

\subsection{Isometries}

\subsubsection{The solvable parameterization}
Given the Lie algebra embeddings detailed above, we are ready to move to the group parameterization and the action of the isometry group 
$\mathrm{U}$. We recall some issues thoroughly discussed in the foundational paper \cite{pgtstheory}, starting from the general concept of 
\textbf{coset representative}. A coset manifold $\mathrm{U/H}$ is by definition a  manifold whose points are the equivalence classes of 
$\mathrm{U}$ elements with respect to the equivalence relation: 
 \begin{equation}\label{lateralclass}
   \forall g_{1,2} \in \mathrm{U} \quad \quad g_1 \,\sim\, g_2 \quad \text{iff} \quad g_2^{-1}\cdot g_1\in \mathrm{H}
 \end{equation}
 A coset parameterization is obtained by choosing a family of \textbf{coset representatives} $\mathbb{L}(\boldsymbol{Y})\in \mathrm{U}$ 
 that are group elements continuously depending on a set of $d$ parameters $Y^I$ ($I=1,\dots,d$), where
 $d=\text{dim} \mathrm{U/H}$, in such a way that: 
 \begin{equation}\label{muccacarolina}
   \mathbb{L}(\boldsymbol{Y}) \sim \mathbb{L}(\boldsymbol{X}) \quad \Rightarrow \quad \boldsymbol{Y} = \boldsymbol{X}
   \end{equation}
   and as $\boldsymbol{Y}$ varies in its domain of definition we cover all lateral classes of $\mathrm{U/H}$. Any different coset
   parameterization corresponds to a different choice of coordinates on the Riemannian manifold $\mathrm{U/H}$. The action of the full group $\mathrm{U}$ on the coset $\mathrm{U/H}$ is given by: 
   \begin{equation}\label{cronosputo}
    \forall  g \in \mathrm{U} \quad : \quad g\cdot \mathbb{L}(\boldsymbol{Y}) \, = \, \mathbb{L}\left(\mathit{f}_g\left[\boldsymbol{Y}\right]\right)\cdot h\left(g,\boldsymbol{Y}\right) \quad ; \quad h\left(g,\boldsymbol{Y}\right) \in \mathrm{H } \subset \mathrm{U}
   \end{equation}
   where the typically non-linear function $\mathit{f}_g\left[\boldsymbol{Y}\right]$ yields the parameters of the new lateral class,
   namely of the new $\mathrm{U/H}$-point, image through $g$ of the original one, while the $\mathrm{H}$-element 
   $h\left(g,\boldsymbol{Y}\right)$, which generically depends both on the group element $g$ acting the transformation and on the point $\boldsymbol{Y}$ in $\mathrm{U/H}$ to which the transformation is applied, is named the \textbf{compensator}. 

 In practice, the metric equivalence of the non-compact symmetric space $\mathrm{U/H}$ with the solvable Lie group $\mathcal{S}_{\mathrm{U/H}}$ amounts to a particularly nice and useful parameterization of the coset. Utilizing the canonical triangular embedding of statement \ref{statamento} for all $\mathrm{U/H}$ we can utilize a coset representative $\mathbb{L}(\boldsymbol{\Upsilon})$ which is an \textbf{upper triangular} $N\times N$ matrix, $N$ being the dimension of the defining fundamental representation of $\mathrm{U}$.
   In the maximally split case $\mathrm{SL(N,\mathbb{R})}/\mathrm{SO(N)}$, the only condition on  $\mathbb{L}(\boldsymbol{\Upsilon})$ is that it should have determinant equal to one and be upper triangular, namely, utilizing a classical notation,  discussed in \cite{pgtstheory} we must have:
\begin{equation}\label{krolik}
 \mathbb{L}(\boldsymbol{\Upsilon}) = \mathbb{L}_{>}(\boldsymbol{\Upsilon}) \quad ; \quad \text{Det} \mathbb{L}(\boldsymbol{\Upsilon}) \, = \, 1
\end{equation}
while in the case of the pseudo-orthogonal  coset manifolds (\ref{Mrq}), the conditions are:
 \begin{equation}\label{lepre}
   \mathbb{L}(\boldsymbol{\Upsilon}) = \mathbb{L}_{>}(\boldsymbol{\Upsilon}) \quad ; \quad \mathbb{L}^T(\boldsymbol{\Upsilon})\cdot \eta_t \cdot \mathbb{L}(\boldsymbol{\Upsilon}) \, = \, 
    \eta_t
 \end{equation}
 Said in simple words $\mathbb{L}(\boldsymbol{\Upsilon})$ should be both an element of $\mathrm{SO(r,r+q)}$ and an upper triangular matrix.
 \begin{lemma}\label{ciurla}
 The condition (\ref{lepre}) implies that also the following equation is satisfied
 \begin{equation}\label{binocchio}
   \mathbb{L}(\boldsymbol{\Upsilon})\cdot \eta_t \cdot \mathbb{L}^T(\boldsymbol{\Upsilon}) \, = \, 
    \eta_t
 \end{equation}
 \end{lemma}
 \begin{prooflem}
 Since $\eta_t^2 \, = \, \mathrm{Id}_{N\times N}$ we have $\eta_t \cdot \mathbb{L}^T \cdot \eta_t \cdot \mathbb{L} \, = \, \mathrm{Id}$. Therefore we have:
 \begin{equation}\label{ciulifischio}
   \eta_t \cdot \mathbb{L}^T \cdot \eta_t  \, = \, \mathbb{L}^{-1}
 \end{equation}
 Since $\mathrm{SO(r,r+q)}$ is a group and $\mathbb{L} \in \mathrm{SO(r,r+q)}$ it follows that its inverse is in the group as well so that $\eta_t \cdot \mathbb{L}^T \cdot \eta_t \in\mathrm{SO(r,r+q)}$. Then by definition of the group $\mathrm{SO(r,r+q)}$ we have:
 \begin{equation}\label{ponedielnik}
   \eta_t \cdot \mathbb{L} \cdot \eta_t \cdot \eta_t \cdot \eta_t \cdot\mathbb{L}^T \cdot \eta_t  \, = \, \eta_t
 \end{equation}
 multiplying the above equation on the left and the right by $\eta_t$ eq. (\ref{binocchio}) follows.
 \end{prooflem}
 As it was explained in \cite{pgtstheory}, a convenient way to label all the points of the coset manifold $\mathrm{U/H}$ is by introducing the symmetric matrix:
 \begin{equation}\label{matraemme}
   \mathcal{M}(\boldsymbol{\Upsilon}) \, \equiv \, \mathbb{L}(\boldsymbol{\Upsilon})\cdot \mathbb{L}^T(\boldsymbol{\Upsilon})
 \end{equation}
 whose great advantage is that of being insensitive to the compensator of equation (\ref{cronosputo}). Indeed we immediately obtain:
 \begin{equation}\label{aulabizantina}
   \forall g \in \mathrm{U} \quad : \quad g\cdot \mathcal{M}(\boldsymbol{\Upsilon})  \cdot g^T \, = \, \mathcal{M}\left(\mathit{f}_g\left[\boldsymbol{\Upsilon}\right]\right) \, \equiv \, \mathcal{M}\left(\Upsilon^\prime\right)
 \end{equation}
 and because of the above lemma \ref{ciurla} we also have:
 \begin{equation}\label{cromoplaccato}
   \mathcal{M}(\boldsymbol{\Upsilon}) \, = \, \mathcal{M}^T(\boldsymbol{\Upsilon}) \quad ; \quad \mathcal{M}(\boldsymbol{\Upsilon})\cdot \eta_t \cdot \mathcal{M}(\boldsymbol{\Upsilon}) \, = \, \eta_t
 \end{equation}
 for the pseudo-orthogonal case while for the maximally split case, we simply have:
 \begin{equation}\label{oroplaccato}
   \mathcal{M}(\boldsymbol{\Upsilon}) \, = \, \mathcal{M}^T(\boldsymbol{\Upsilon}) \quad ; \quad \text{Det} \left[ \mathcal{M}(\boldsymbol{\Upsilon}) \right]\, = \,  1
 \end{equation}
 Note also that for the symplectic case, everything is analogous replacing the symmetric invariant matrix $\eta_t$ with an antisymmetric one $c_t$.
 \par
 Summarizing the above discussion we can say that we have two allied parameterizations of the same coset, the upper triangular one $\mathbb{L}(\boldsymbol{\Upsilon})$ and the symmetric one $\mathcal{M}(\boldsymbol{\Upsilon})$ related by 
 equation (\ref{matraemme}).
 
 \subsubsection{The action of \texorpdfstring{$\mathrm{U}$}{U} on the symmetric space \texorpdfstring{$\mathrm{U/H}$}{U/H}}
 
 The action of the full group of isometries $\mathrm{U}$ is best worked out using the symmetric matrix $\mathcal{M}(\boldsymbol{\Upsilon})$ and eq.(\ref{aulabizantina}). In order to find:
 \begin{equation}\label{bezdomny}
  \boldsymbol{\Upsilon}^\prime \, = \,\mathit{f}_g\left[\boldsymbol{\Upsilon}\right]
 \end{equation}
 we just need the inverse of the relation (\ref{matraemme}) that we can identify with what we name the abstract operation
 $\varpi^{-1}$:
 \begin{eqnarray}\label{varromenuno}
   \varpi^{-1} &\quad : \quad &\mathcal{S} \, \longrightarrow \, \dfrac{\mathrm{U}}{\mathrm{H}}\nonumber\\
   \varpi^{-1}\left[\mathbb{L}(\boldsymbol{\Upsilon})\right] & \quad = \quad & \mathbb{L}^T(\boldsymbol{\Upsilon})\cdot\mathbb{L}(\boldsymbol{\Upsilon})
 \end{eqnarray}
 The inverse of the operation $\varpi^{-1}$, namely $\varpi$ is the well-known algebraic and iterative Cholewsky-Crout algorithm that for each symmetric non-degenerate matrix $\mathcal{M}$  determines the unique upper triangular matrix $\mathbb{L}$ such that $\mathcal{M}\, = \, \mathbb{L} \cdot \mathbb{L}^T$  (see Eq (4.35) of \cite{pgtstheory} that we repeat
 here for completeness):
\begin{align}\label{ciulacavolo}
\mathbb{L}_{jj}&=\sqrt{\mathcal{M}_{jj}-\sum_{k=1}^{j-1} \mathbb{L}_{jk}^2}\nonumber\\
\mathbb{L}_{ji}&=\frac{1}{\mathbb{L}_{jj}}\,\left(\mathcal{M}_{ij}-\sum_{k=1}^{j-1}\,
\mathbb{L}_{jk}\mathbb{L}_{ik}\right)\,\,,\,\,\,i>j\,,
\end{align}
Hence we can set:
\begin{eqnarray}\label{tazzone}
   \varpi &\quad : \quad & \dfrac{\mathrm{U}}{\mathrm{H}}\, \longrightarrow \, \mathcal{S}\nonumber\\
   \varpi\left[\mathcal{M}(\boldsymbol{\Upsilon})\right] & \quad = \quad & \mathbb{L}(\boldsymbol{\Upsilon})
 \end{eqnarray} 
 and the action on the solvable coordinates $\boldsymbol{\Upsilon}$ of any element $g \in \mathrm{U}$ can be formalized as:
 \begin{eqnarray}\label{tarantola}
 \mathit{f}_g\left[\boldsymbol{\Upsilon}\right]\cdot \mathrm{T} & = & \Sigma^{-1}\circ \varpi \circ \mathrm{Adj}_g \circ \varpi^{-1} \circ\Sigma \left(\boldsymbol{\Upsilon}\cdot \mathrm{T}\right) \nonumber\\
 \mathrm{Adj}_g \left(M\right) & = & g\cdot M \cdot g^T   \nonumber\\
 \mathrm{T} &=& \{ T_1,\dots , T_n\} \quad ; \quad \text{basis of generators of $Solv_{\mathrm{U/H}} $}
 \end{eqnarray}
 \par

\subsection{Lie group homomorphisms at variable \texorpdfstring{$r_{n.c}$}{rnc}: TS class switching}
\label{Liegiocaflipper}
Because of the metric equivalence of all the non-compact symmetric spaces with appropriate solvable Lie groups $\mathcal{S}_{[r,q]}$ and of the 
universal triangular embedding in $\mathrm{SL(N,\mathbb{R})}$ guaranteed by statement \ref{statamento} we can consider the general problem of 
homomorphisms between solvable Lie algebras of different rank and different dimensions. Each of such homomorphisms will generate, according to 
lemma \ref{omomorfista}, a corresponding solvable Lie group homomorphism that will also be the overall covariant differentiable map from one layer 
to the next one in a neural network without point-wise activation functions. To maintain the discussion at the general level, it is convenient to 
change the notation for solvable Lie algebras and remove all labels from them. The homomorphisms we want to describe are linear maps of the 
generic form: \begin{equation}\label{generico} 
 \mathcal{W} \quad : \quad   Solv_1 \, \longrightarrow \, Solv_2
\end{equation} We name such map $\Sigma$, since it is going to replace the map $\boldsymbol{\Sigma}$ of eq. 
(\ref{sigmusorror}) realized in terms of point-wise activation functions. Thus we have: 
\begin{alignat}{5}\label{sigmuspulcher} 
   \Sigma_i  & \quad  : \quad &  Solv_i \longrightarrow \, & \quad \mathcal{S}_i .
 \end{alignat}
 \begin{remark}
     $\Sigma$ is invertible for all the manifolds so far considered.
 \end{remark}
 \par
 At this point, we have to recall from the foundational paper \cite{pgtstheory} that the solvable Lie algebras we are discussing 
 are \textbf{normed solvable Lie algebras} and that is the reason why their corresponding solvable Lie group can be seen to be metrically 
 equivalent to the associated non compact symmetric space $\mathrm{U/H}$, as recalled in Theorem \ref{thm:metricequivalence}.  \par
 Relying on this we can consider a further map from the solvable Lie algebra $Solv_i$ to 
 the linear space of left-invariant $1$-forms on the solvable Lie group $\mathcal{S}_i$. 
 More precisely, given any parameterization of the solvable Lie group in particular, that defined as the image of the $\Sigma$  map:  
  \begin{equation}\label{magozurli}
   \forall \boldsymbol{\Upsilon}\cdot \mathrm{T} \in Solv_i \quad : \quad \Sigma_i\left(\boldsymbol{\Upsilon}\cdot \mathrm{T} \right) \, \equiv \, \mathbb{L}\left( \boldsymbol{\Upsilon}\right)
   \, \in \, \mathcal{S}_i
  \end{equation}
 where $ \mathrm{T} = \{T_1,\dots, T_d\}$ denotes an abstract set of generators of the solvable Lie algebra: 
 \begin{equation}\label{commialg}
   \left[ T_j \, , \, T_k \right] \, = \, f^i_{\phantom{i}jk} \, T_i \quad ;\quad  i,j,k\, =\, i,\dots,d
 \end{equation}
We recall that the Maurer-Cartan $1$-form for matrix Lie Algebras can be written as:
 \begin{equation}\label{thetamappata}
   \Theta \left[\mathbb{L}\right] \, \equiv \, \mathbb{L}^{-1} \mathrm{d}\mathbb{L} \
 \end{equation}
where we remind the reader that, by construction, this left-invariant one-form on a group takes values in its Lie algebra
(for a general discussion of the left/right invariant $1$-forms on Lie Groups and coset manifolds see \cite{pietro_discrete}
 in particular chapter 7). 
Combining the Maurer-Cartan form with the operation defined in
\ref{def:normedalgebra}, we obtain a map from the solvable Lie algebra to the left-invariant 
 Maurer-Cartan forms on the solvable group manifold:
\begin{equation}\label{committoni}
   \mathcal{E}^i \equiv \, \langle \mathbb{L}^{-1} \mathrm{d}\mathbb{L}  \, , \,  T_{i} \rangle
 \end{equation}
where $\langle T_j\, , \, T_{i} \rangle \, = \, \delta^i_j$.
 By construction, they satisfy the Maurer-Cartan equations:
 \begin{equation}\label{confetto}
   \mathrm{d}\mathcal{E}^i \, + \, \ft 12 \, f^i_{\phantom{i}jk} \, \mathcal{E}^j \wedge \mathcal{E}^k \, = \, 0\quad ; \quad i,j,k\, =\, i,\dots,d
 \end{equation}
 with the very same structure constants $f^i_{\phantom{i}jk}$ that appear in eq.(\ref{commialg}).
 \par
We make the following observations:
 \begin{description}
   \item[a)] Since the structure constants appearing in Maurer-Cartan equations are the same as those appearing in the Lie bracket description of the same algebra it follows that replacing in equation (\ref{confetto}) the  $1$-forms $\mathcal{E}^i \left( \boldsymbol{\Upsilon}\right)$ explicitly depending on a set of coordinates $\boldsymbol{\Upsilon}$ determined by the specific form of the parametrization $\Sigma$, with abstract one forms $E^i$,  the  formal Maurer-Cartan equations
      \begin{equation}\label{cioccolatino}
   \mathrm{d}{E}^i \, + \, \ft 12 \, f^i_{\phantom{i}jk} \, {E}^j \wedge {E}^k \, = \, 0 \quad ; \quad i,j,k\, =\, i,\dots,d
 \end{equation} 
 are a completely equivalent definition of a Lie algebra as eq.s  (\ref{commialg}).
   \item[b)] Given two solvable Lie algebras of different dimensions $d_{2},d_{1}$ respectively described by two sets of Maurer-Cartan equations:
       \begin{eqnarray}
        \mathrm{d}{E}^i \, + \, \ft 12 \, f^i_{\phantom{i}jk} \, {E}^j \wedge {E}^k  &=& 0 \quad ; \quad i,j,k\, =\,1,\dots,d_2 
        \label{pignatta}\\
         \mathrm{d}{e}^\alpha \, + \, \ft 12 \, g^\alpha_{\phantom{\alpha}\beta\gamma} \, {e}^\beta \wedge {e}^\gamma 
         &=& 0 \quad ; \quad \alpha,\beta,\gamma\, =\,1,\dots,d_1\label{casseruola}
       \end{eqnarray}
       the most general homomorphism (\ref{generico})  is described by a linear map:   
       \begin{equation}\label{cardioaspirina}
         {E}^i \, = \, W^{i}_\alpha \, {e}^\alpha
       \end{equation}
       where $W^{i}_\alpha$ is a generically rectangular constant matrix that must satisfy the constraint that the target Maurer-Cartan equations (\ref{pignatta}) should be identically satisfied if the Maurer-Cartan equations (\ref{casseruola}) are assumed to hold true.
 \end{description}
 
 Once a consistent linear map as in eq.(\ref{cardioaspirina}) is found, by replacing the corresponding explicit $1$-forms 
 $\mathcal{E}^i(\boldsymbol{\Upsilon})$ and $\varepsilon^\alpha(\boldsymbol{X})$ respectively dependent on the solvable parameters 
 $\boldsymbol{\Upsilon}$ of the second solvable group and on the solvable parameters $\boldsymbol{X}$ of the first one, we obtain a system of 
 first-order differential equations determining the functional dependence of the $\boldsymbol{\Upsilon}$ in terms of the $\boldsymbol{X}$ 
 \begin{equation}\label{pergolato} 
         \mathcal{E}^i( \boldsymbol{\Upsilon})\, = \, W^{i}_\alpha \, \varepsilon^\alpha(\boldsymbol{X})
 \end{equation}
 The functional dependence $\boldsymbol{\Upsilon}(\boldsymbol{X})$ which solves eq.s (\ref{pergolato}) is generically non-linear yet it can be 
 always determined because of the solvable structure of the Lie algebra. Indeed eq.s (\ref{pergolato}) can be solved iteratively in every explicit 
 case of admissible matrix $W^{i}_\alpha$. We show examples of this in Appendix \ref{app:maurercartan} and \ref{patagonia}. 
 \par 
 Let us give a name to the solution of equations (\ref{pergolato}) by writing:
 \begin{equation}\label{carriola}
  \boldsymbol{\Upsilon}\cdot \mathrm{T} \, = \,  \boldsymbol{\Phi}\left[ \mathcal{W} \, | \,  \boldsymbol{X}\cdot \mathrm{t} \right]
 \end{equation} where we have named $\mathrm{t}$ the basis of solvable Lie generators of the first Lie algebra and $\mathrm{T}$ the same for the 
 target Lie algebra. Formally we can set: 
 \begin{equation}\label{pagnottagreca} \Upsilon^i(\boldsymbol{X}) \, = \, \langle T^{i\, \star} \, , \, \boldsymbol{\Phi}\left[ \mathcal{W} \, | 
 \,  \boldsymbol{X}\cdot \mathrm{t} \right] \rangle 
 \end{equation} 
 where $\{T^{i\star}\}$ are the duals of the generators $T_i$.
 With this understanding the non linear map $\boldsymbol{\Phi}\left[ \mathcal{W} \, | \,  \boldsymbol{X} \right]$ becomes
 the core of the map from a solvable Lie group to another one since we can write:
 \begin{eqnarray}\label{prato}
   \boldsymbol{\mho}_{\mathcal{W}} & : & \mathcal{S}_a \, \longrightarrow \, \mathcal{S}_b \nonumber\\
   \forall \mathfrak{s}\in \mathcal{S}_a & : & \boldsymbol{\mho}_{\mathcal{W}} \left( \mathfrak{s}\right) \, \equiv \, \Sigma\left(
   \boldsymbol{\Phi}\left[ \mathcal{W} \, | \,  \Sigma^{-1}(\mathfrak{s}) \right]\right) \in \mathcal{S}_b
 \end{eqnarray}
 \par
 In this way the map from the solvable Lie group $\mathcal{S}_{i}$ in the $i$-th layer to the solvable Lie group $\mathcal{S}_{i+1}$ in the $(i+1)$-th layer can be written as follows:
 \begin{eqnarray}\label{embryo}
   \mathcal{K}_{\mathcal{W}_{i}} & \quad : \quad & \mathcal{S}_{i} \, \longrightarrow \, \mathcal{S}_{i+1} \nonumber\\
   \mathcal{K}_{\mathcal{W}_{i}} & \quad = \quad & \boldsymbol{\mho}_{\mathcal{W}_i}
 \end{eqnarray}
 \subsection{The general map between layers} After the map between solvable Lie groups has been obtained the map can be extended to a more general 
 map between the two corresponding symmetric spaces using their metric equivalence with the corresponding solvable Lie groups. Indeed we can act 
 on the points of the symmetric space $\mathrm{U}_{i+1}/\mathrm{H}_{i+1}$ with any element $g \in \mathrm{U}_{i+1}$ which maps the symmetric space 
 into itself: 
 \begin{equation}\label{carburo}
    g \quad : \quad \dfrac{\mathrm{U}_{i+1}}{\mathrm{H}_{i+1}} \, \longrightarrow \, \dfrac{\mathrm{U}_{i+1}}{\mathrm{H}_{i+1}}
 \end{equation}
naming $\varpi_{n}$ the equivalence map of the $n$-th symmetric space with its corresponding solvable Lie group shown in Equation \ref{tazzone}:
\begin{eqnarray}\label{ganno}
  \varpi_{n} &\quad : \quad &\dfrac{\mathrm{U}_{n}}{\mathrm{H}_{n}} \, \longrightarrow \, \mathcal{S}_{n}\nonumber\\
  \varpi_{n}^{-1} &\quad : \quad &\mathcal{S}_{n} \, \longrightarrow \, \dfrac{\mathrm{U}_{n}}{\mathrm{H}_{n}}
\end{eqnarray}
we can write the complete form of the map from the $i$-th layer to the $(i+1)$-layer in the following concise and very elegant manner:
\begin{eqnarray}\label{grandebellezza}
 \hat{ \mathcal{K}}^{i}_{[\mathcal{W}_i,\Psi_{i+1}]} &\quad : \quad & \dfrac{\mathrm{U}_{i}}{\mathrm{H}_{i}}  \longrightarrow \dfrac{\mathrm{U}_{i+1}}{\mathrm{H}_{i+1}} \nonumber \\
\hat{ \mathcal{K}}^{i}_{[\mathcal{W}_i,\Psi_{i+1}]} &\quad = \quad &\mathrm{ Adj}_{g(\Psi_{i+1})} \circ \varpi_{i+1}^{-1} \circ \boldsymbol{\mho}_{\mathcal{W}_i}\circ \varpi_i
\end{eqnarray}
where $g(\Psi_{i+1})$ denotes the element of the isometry group $\mathrm{U}_{i+1}$ identified by a convenient set of parameters $\Psi_{i+1}$ in any adopted parameterization of that Lie group.  The beauty of eq. (\ref{grandebellezza}) resides in the following properties:
\begin{description} 
\item[a)] All the utilized operations have an intrinsic meaning and do not depend on a specific choice of parameterization or 
of a coordinate basis. 
\item[b)] The structure of eq.(\ref{grandebellezza}) is fully general within the class of chosen manifolds, namely the 
non-compact symmetric spaces. 
\item[c)] Formula (\ref{grandebellezza}) is general, yet it is rooted in the specific features of the chosen class 
of manifolds (selected as we already said from first principles and not "ad hoc") and does not make sense without them. In particular, it relies 
on the universal metric equivalence of non-compact symmetric spaces with appropriate solvable Lie groups. 
\item[d)] Formula (\ref{grandebellezza}) 
provides a clearcut geometric interpretation of the items advocated by classical neural networks namely the linear maps $\mathcal{W}$  and the 
bias translations. The former ones are the respective cores of the homomorphisms between solvable Lie groups, each being dictated by a linear 
homomorphism of solvable Lie algebras, while the latter ones are just generic isometries of the target symmetric spaces.   
\end{description} 
\subsubsection{The formula in practice}
 The master formula (\ref{grandebellezza}) can be made thoroughly explicit. To begin with, we have a generic point in the $i$-th layer which means 
 in the manifold $\mathrm{U}_i/\mathrm{H}_i$. It is represented by the symmetric matrix $\mathcal{M}_i(\boldsymbol{\Upsilon}_i)$ parameterized by 
 the solvable coordinates $\boldsymbol{\Upsilon}_i$ that identify an element of the solvable Lie algebra $Solv_{\mathrm{U}_i/\mathrm{H}_i}$ as 
 $\boldsymbol{\Upsilon}_i \cdot \mathrm{T}_i$. The element of the solvable Lie algebra and the solvable coordinate vector $\boldsymbol{\Upsilon}_i 
 $ are extracted from the symmetric matrix $\mathcal{M}_i(\boldsymbol{\Upsilon}_i)$ executing in sequence the operation $\varpi_i$ which gives as 
 output the upper triangular matrix $\mathbb{L}(\boldsymbol{\Upsilon}_i)$ representing an element of the solvable Lie group $\mathcal{S}_i$ and 
 then the operation $\Sigma^{-1}$ which gives as output the solvable Lie algebra element $\boldsymbol{\Upsilon}_i \cdot \mathrm{T}_i$. Next, we 
 have the core of the map which is appropriately determined, as we explained above, from the chosen instance of the solvable Lie algebra 
 homomorphism (\ref{generico}) 
 \begin{equation}\label{iesimohomo}
 \mathcal{W}_i \quad : \quad   Solv_i \, \longrightarrow \, Solv_{i+1}
 \end{equation}
 Once we have the solvable Lie algebra element from the solution of eq.s (\ref{pergolato})
\begin{equation}\label{caterpiller}
   \boldsymbol{\Upsilon}_{i+1} \cdot \mathrm{T}_{i+1} \, = \, \boldsymbol{\Phi}\left[\mathcal{W}_i \, \mid \, \boldsymbol{\Upsilon}_{i} \cdot \mathrm{T}_{i} \right] \, \in \, Solv_{i+1}
 \end{equation}
 we can uplift it to the solvable Lie group with the map $\Sigma$ (all together this is the map $\boldsymbol{\mho}_{\mathcal{W}_i}$) and then transform the output $\mathbb{L}_{i+1}$ of such operation to a symmetric matrix $\mathcal{M}_{i+1}$  which can be further transformed using the adjoint action of any group element $g(\Psi_{i+1})$ of the isometry group $\mathrm{U}_{i+1}$, yielding a new symmetric matrix 
 $\tilde{\mathcal{M}}_{i+1}$ wherefrom we can start once again the map to a new layer $i+2$.
 \par
 The above-detailed illustration clearly shows that the linear solvable Lie algebra homomorphism (\ref{iesimohomo})  can change the dimensions of the next layer (to a smaller or larger one). This map encapsulates the relevant parameters to be learned by the neural network algorithm. The non-linearity of the map between two layers has two combined sources: on one side the parametrization $\Sigma$ on the other the map $\Phi[\mathcal{W}\mid \boldsymbol{\Upsilon}_{i}]$ originating from the solution of the first order equations (\ref{pergolato}). This map can introduce both exponentials and polynomial non-linearity with a degree of the polynomials that grows rapidly with the number of dimensions. Hence, while preserving covariance at all steps of the procedure, the structure of the function that maps the input into the output acquires a lot of complexity with free parameters for learning. 
 \par

\section{The \texorpdfstring{$r=1$}{r=1} class, or hyperbolic spaces}
\label{architettoTS}
The various manifolds in the same \textbf{TS universality class} provide \textit{a natural, yet not exclusive, conceptual model} of the various 
possible \textbf{layers} in a geometric Deep Learning architecture. Indeed as we have seen in the previous section and will discuss in detail in 
App. \ref{patagonia}, maps from one TS universality class to another one are perfectly possible and admissible. This opens a large landscape of 
possible geometric learning architectures that have to be carefully explored and numerically tested. 
\par In this ample landscape, an interesting family of architectures is precisely that where all the layers belong to the same Tits-Satake 
$r$-class, namely where the $i$-th map is of the following form: 
\begin{equation}\label{ithmap}
 \hat{\mathcal{K}}^{i}_{[r]} \quad : \quad  \mathcal{M}^{r,q_i} \, \longrightarrow \,  \mathcal{M}^{r,q_{i+1}}
\end{equation}
from the symmetric space $\mathcal{M}^{r,q_i}$ sitting in layer $i$-th to the symmetric space $\mathcal{M}^{r,q_{i+1}}$ sitting in layer $(i+1)$-th.
\par
An even more restrictive choice corresponds to choosing the family where $r=1$, namely where all the layers are hyperbolic spaces. This 
constitutes the topic of our work \cite{naviga}. In that context, we have explored the 
problem of defining \textbf{separators}, already highlighted in this context and tackled by \cite{ganea_hyperbolic_2018, shimizu_2021}, essential for the implementation of the \textbf{logistic regression} algorithm (see sect.\ref{cicero}).  
Similarly to these previous works, for manifolds $\mathcal{M}^{1,q}$ of the $r=1$ class we have found a geometrically natural identification of the separator with a 
\textbf{codimension one totally geodesic submanifold $\mathfrak{S}\subset \mathcal{M}^{1,q}$}, that intersects the space boundary 
$\partial\mathcal{M}^{1,q}$ and halves $\mathcal{M}^{1,q}$  into two disjoint parts. On the basis of general theorems $\mathfrak{S}$ is 
\textit{necessarily} a symmetric space and indeed $\mathfrak{S} \, = \,\mathrm{Adj}_g \left[\mathcal{M}^{1,q-1}\right]$   is the conjugate for 
some $g\in \mathrm{SO(1,1+q)}$ of the standard submanifold $\mathcal{M}^{1,q-1}\subset \mathcal{M}^{1,q}$. Being a symmetric space, the separator 
is in particular a \textbf{homogeneous space}. We  leave the complete discussion of this issue to our next coming theoretical work 
\cite{tassella}, yet we mention that for  $r\geq 2$ separators $\mathfrak{S}$ are \textbf{codimension one homogeneous spaces}, yet they are 
neither symmetric nor totally geodesic.

\subsection{Solvable parametrization of hyperbolic spaces}
\label{r1mapposo}
In this section, we apply the general framework of section \ref{generalemapposo} to the case $r=1$.
For a manifold $\mathrm{SO(1,2+q)/SO(2+q)}$, using the solvable coordinates defined in the way
we are going to show, the metric takes the following form:

\begin{equation}\label{metrulla}
  ds^2 \, = \, {dw_1^2}+
   \sum_{i=1}^q\,\frac{1}{4} (w_{1+i} dw_1 +dw_{1+i})^2
\end{equation}
and the Tits-Satake projection (see \cite{pgtstheory} for a completed discussion of all related concepts) corresponds to setting $w_{2+i} \, = \, 0$,  ($i=1,\dots, q$). We name $w_i$ the solvable coordinates, and the standard  generators of the solvable Lie algebra satisfy the following  commutation relations:
\begin{equation}\label{commuti}
  \left[T^1\, , \, T^{1+a}\right] \, = \, T^{1+a} \quad , \quad a=1,\dots,q+1 \quad ; \quad \left[ T^{1+a} \, , \, T^{1+b}\right] 
  \, = \, 0 \quad , \quad 
  a,b \, = \, 1,\dots, q+1
\end{equation}
\par 

We also note that setting $w_{2+q}\, = \, 0$ yields the standard symmetric subspace $\mathcal{M}^{[1,q-1]}\subset\mathcal{M}^{[1,q]}$ whose conjugate $g$ image can act as separator as stated above.
To derive the simple form of the metric (\ref{metrulla}), we have to discuss the choice of coordinates on the solvable Lie manifold which amounts to an explicit rule for the parametrization $\Sigma$. This opens a digression.

\subsubsection{The parameterization \texorpdfstring{$\Sigma$}{Sigma}}
\label{sigmone} Given the general framework of triangular embeddings of all solvable Lie algebras into the solvable subalgebra of the maximally 
split algebra $\mathfrak{sl}(\mathrm{N}, \mathbb{R})$, it is appropriate to begin our analysis with this maximal case. Here, the solvable Lie algebra is  
the complete Borel subalgebra $\mathbb{B}_\mathrm{N} \subset \slal(\mathrm{N},\mathbb{R})$ that is generated as follows: 
\begin{equation}\label{Tgeneratori}
 \mathbb{B}_\mathrm{N}  \, = \, \text{span}\left[\mathrm{T}\right]\quad ; \quad \mathrm{T} \, = \, \left\{
 \mathcal{H}_{i=1,\dots,N-1}, \quad\mathcal{E}^\alpha , \quad \alpha >0\right\}
\end{equation}
where $\mathcal{H}_i$ are the Cartan generators and $\mathcal{E}^\alpha $ are the step operators associated with the positive roots of the Lie algebra. The important thing to stress is that the positive roots $\alpha$ of a Lie algebra have an intrinsic, base-independent, partial ordering given by their height. At height $ht=1$ we have the simple roots $\alpha_{i}$, at $ht=2$ we have the sums of two simple roots, and so on up to the maximal root that is unique. In the case of the $\mathfrak{a}_{N-1}$ Lie algebras the partial ordering can be easily transformed into an absolute order since given an order to the simple ones according to the Cartan matrix, then all the other roots are necessarily of the form $\alpha_i +\alpha_{i+1}+\dots + \alpha_{i+p}$ where 
$p+1$ is the height. Furthermore, the Cartan generators can be paired one-to-one with simple roots. This means there is an intrinsic order of the generators for the Borel solvable Lie algebra starting from the first Cartan and going up to the maximal root. This means that we can define a convenient parametrization $\Sigma$ in the following way:
\begin{equation}\label{sigmadefillo}
  \Sigma\left ( \sum_{i=1}^{\frac{N(N+1)}{2} -1} \Upsilon_i \, \mathrm{T}_i \right)\, \equiv \, \prod_{i=1}^{\frac{N(N+1)}{2} -1} \exp\left[ \Upsilon_i \, \mathrm{T}_i \right]
\end{equation}
The reason why this parametrization is optimal is that it provides a simple expression that can be inverted iteratively without ever requiring the \textit{matrixlog} operation. Furthermore, it respects the intrinsic order of the generators and of the corresponding one-parameter subgroups that are algebraically characterized by roots. Subalgebras and hence subgroups are visible in this way (see \cite{pgtstheory} for further details).
\par 
Matrix-wise we can choose a matrix basis for the solvable Lie algebra generators in such a way that in 
$\boldsymbol{\Upsilon}\cdot \mathrm{T}$ the Cartan coordinates $\Upsilon_{1,\dots, N-1}$ sit on the central diagonal, 
the coordinates corresponding to the simple roots $\Upsilon_{N,\dots, 2N-2}$ sit on the first principle diagonal right of the central one, 
and so on up to the coordinate corresponding to the highest root $\Upsilon_{\frac{N(N+1)}{2}-1}$ that is located  in the leftmost upper corner 
as one sees below:

\begin{equation}\label{pittura}
\begin{pmatrix}
-\displaystyle\sum_{i=1}^{N-1}\Upsilon_{i} 
& \Upsilon_{N} & \Upsilon_{2N-1} & \dots & \cdots & \Upsilon_{\frac{N(N+1)}{2}-1} \\[1em]
0 & \Upsilon_{1} & \Upsilon_{N+1} & \Upsilon_{2N} & \cdots & \Upsilon_{\frac{N(N+1)}{2}-2} \\[0.5em]
0 & 0 & \Upsilon_{2} & \Upsilon_{N+2} & \ddots & \vdots \\[0.5em]
0 & 0 & 0 & \Upsilon_{3} & \ddots & \vdots \\[0.5em]
\vdots & & & \ddots & \ddots & \Upsilon_{2N-2} \\[0.5em]
0 & \cdots & 0 & 0 & 0 & \Upsilon_{N-1}
\end{pmatrix}
\end{equation}

For the non maximally split algebras the ordering of the solvable Lie algebra generators is also done on the basis of the height
of the corresponding roots but one has also the phenomenon of the roots with multiplicity that has to be taken into account
(see \cite{pgtstheory} for all the notions concerning the multiplicity of roots and the Paint Group). For the $r=1$ case the explicit form of $\Sigma$ from the solvable Lie algebra to the solvable group 
\begin{equation}\label{sigmaexp}
  \Sigma \, : \, Solv \, \stackrel{\Sigma}{\longrightarrow} \, \mathcal{S} \, \equiv \, \exp \left[Solv\right]
\end{equation}
was defined in general in \cite{pgtstheory} (see eq.s (3.53)-(3.54) of that paper). Our choice which optimizes the metric  and  the Laplacian to  their simplest possible forms is the following (see details in \cite{pgtstheory} for the concepts of long roots that are Paint singlets and short  roots with Paint Group multiplicity): 
\begin{eqnarray}\label{SigmaExp1}
  \Sigma\left[\mathfrak{s}\left(\boldsymbol{\Upsilon} \right)\right]
  & = & \exp\left[\mathfrak{s}_{C}\left(\boldsymbol{\Upsilon} \right)\right]
  \cdot \exp\left[\mathfrak{s}_{long}\left(\boldsymbol{\Upsilon} \right)\right]\cdot\exp\left[\mathfrak{s}_{short}\left(\boldsymbol{\Upsilon}
  \right)\right]\nonumber\\
  \mathfrak{s}_{C}\left(\boldsymbol{\Upsilon} \right) &=& \text{projection onto the
  Cartan subalgebra of $Solv$} \nonumber\\
  \mathfrak{s}_{long}\left(\boldsymbol{\Upsilon} \right) & =& \text{projection onto the
  subspace  spanned by the long root generators}\nonumber\\
  \mathfrak{s}_{short}\left(\boldsymbol{\Upsilon} \right) & =& \text{projection onto the
  subspace  spanned by the short root generators}\nonumber\\
  \mathfrak{s}\left(\boldsymbol{\Upsilon} \right) &=&\mathfrak{s}_{C}\left(\boldsymbol{\Upsilon} \right)\oplus
  \mathfrak{s}_{long}\left(\boldsymbol{\Upsilon} \right) \oplus \mathfrak{s}_{short}\left(\boldsymbol{\Upsilon} \right)
\end{eqnarray}
In the $r=1$ case there are no long roots and the short roots (associated with the solvable group generators $T^{1+a}$) are $1+q$ so that the 
map $\Sigma$ takes the ultra simple form:
\begin{equation}\label{ladonico}
  \Sigma \left(\underbrace{\sum_{A=1}^{2+q}\,w_A\,T^A}_{\in Solv_{1,1+q}}\right) \, = \, 
  \underbrace{\exp\left[w_1 \, T^1\right]}_{\in \exp[\mathbf{C}]}\cdot\exp\left[\sum_{a=1}^{q+1} \, w_{1+a} \, T^{1+a}\right] \, 
  \equiv \mathbb{L}\left(\mathbf{w}\right)
  \,\in \, \mathcal{S}_{1,1+q}
\end{equation}
where, as before, we have used the shorthand notation $Solv_{[r,r+q]}$ and $\mathcal{S}_{[r,r+q]}$ to denote the solvable Lie algebras and the solvable groups associated with the symmetric spaces defined in eq.(\ref{Mrq}).

We have also emphasized the notion of the maximal non-compact torus, namely the exponential of the non-compact Cartan subalgebra $\exp[\mathbf{C}]$ that for $r=1$ is precisely one-dimensional. There is an intrinsic difference between the coordinates associated with $\mathbf{C}$ and those associated with the other generators of the solvable Lie algebra. The Cartan subalgebra is represented by \textit{simple matrices}, while all the other generators are nilpotent. This means that, in the exponential map, the Cartan coordinates enter as arguments in true exponential functions, while all the others enter only as arguments of polynomial functions. For instance in the case of \textit{supergravity soft cosmic billiards} (see \cite{Fre:2003ep,sashaebog,arrowtime} ) which associates all the fields of the theory to different generators of the solvable Lie algebra, the Cartan ones correspond to the diagonal elements of the space-time metric, while the nilpotent ones correspond to all the rest, namely to off-diagonal metric components and to matter fields. In time evolution along geodesics at large values of time $t$ (both in the past and the future) only the Cartan fields/coordinates survive. In applications to data science, the Cartan generators might represent the most important non-linear interactions of the data while features associated with nilpotent generators will be less relevant.

In the $r=1$ case, the hierarchy is dichotomic because there is just one Cartan and all the other generators are nilpotent and even abelian. With larger 
non-compact rank $r>1$ the maximal torus is ampler and there is room for more than one exponentially dominant combination of features; furthermore, the nilpotent generators have a non-trivial grading as well, since there is a spectrum of nilpotency grades, yielding polynomial dependence of various degrees for the various non-Cartan coordinates/features. 
In the maximally split case of $\slal(\mathrm{N},\mathbb{R})$ the maximal polynomial degree is $N-1$ and the number of Cartan coordinates is $N-1$, achieving universal approximation in the limit $N \rightarrow \infty$. 
\subsection{Isometries and homomorphisms}
This being explained, we continue to lay the ground for the explicit definition of the maps $K^{i}_{[1]}$ defined in eq.(\ref{ithmap}), by writing the explicit form of 
$\mathbb{L}\left(\mathbf{w}\right)$ as presented  in eq.(3.55) of \cite{pgtstheory}. For a generic value of $q$ it reads as follows:
\begin{equation}\label{kramer}
 \Sigma(\mathbf{w}) \, = \, \mathbb{L}(\mathbf{w})\, = \, \left(
\begin{array}{cccccc}
\exp[w_1] & \exp[w_1]\frac{w_2}{\sqrt{2}} & \exp[w_1]\frac{w_3}{\sqrt{2}} & \cdots & \exp[w_1]\frac{w_{q+2}}{\sqrt{2}} & -\frac{1}{4} \left(\sum_{i=1}^{q+1} w_{1+i}^2\right) \\
0 & 1 & 0 & \cdots & 0 & -\frac{w_2}{\sqrt{2}} \\
\vdots & 0 & 1 & \, & 0 &  -\frac{w_3}{\sqrt{2}} \\
\vdots & \, & \, & \ddots & \, & \, \\
0 & 0 & \cdots & 0 & 1 & -\frac{w_{q+2}}{\sqrt{2}} \\
0 & 0 & \cdots & 0 & 0 & \exp[-w_1] \\
\end{array}
  \right)
\end{equation}
The inverse operation $\Sigma^{-1}$ can be formalized by stating:
\begin{equation}\label{comancho}
  \Sigma^{-1} \left[ \mathbb{L}(\mathbf{w})\right] \, = \, \mathbf{w} \, = \, \left\{ w_1,w_2, \dots ,w_{2+q} \right\}
\end{equation}
\subsubsection{Isometries}
Within the general scheme outlined in section \ref{generalemapposo} and subsection \ref{Liegiocaflipper}, there are three fundamental operations we can perform on the solvable group element $\mathbb{L}(\mathbf{w})$ while remaining in the same layer namely at fixed $\mathcal{M}^{1,q} \, = \, \mathbb{H}^{(2+q)}$):
\begin{description}
  \item[a)] The first operation is the action with an element of the compact Paint Group $\mathrm{G_{Paint}}\subset \mathrm{H} \subset \mathrm{U}$ which, as explained in 
      \cite{pgtstheory}, is the external automorphism group of the solvable subgroup $\mathcal{S}_{r,r+q}$ inside 
      $\mathrm{U}=\mathrm{SO(r,r+q)}$. This notion is general for all non-compact symmetric spaces. In the case $r=1$, we simply have 
      $\mathrm{G_{Paint}} \, = \, \mathrm{SO(1+q)}$ and the action of an ortogonal matrix $\mathcal{O}_{Paint}$ on $\mathbb{L}(\mathbf{w})$ is a 
      rotation of the Paint vector $\mathfrak{w}=w_{1+a}$ ($a=1,\dots,q+1$). Hence writing:
      \begin{equation}\label{turiddu}
        \mathbf{w} \, = \, \left\{w_1, \mathfrak{w}\right\}
      \end{equation}
      The paint rotation, which introduces as many parameters as there are in the $\mathrm{SO(1+q)}$ rotation group, is described as follows:
      \begin{equation}\label{stoppables}
       \forall\mathcal{O} \in \mathrm{G_{Paint}} \quad : \quad   
       R_{\mathcal{O}}\, \left[\mathbb{L}\left(\{w_1, \mathfrak{w}\}\right)\right] \, = \, 
       \mathbb{L}\left(\{w_1, \mathcal{O}\mathfrak{w}\}\right)
      \end{equation} 
      \item[b)] The second operation is provided by the left (or right) action of the solvable Lie group on itself. The parameters introduced by this operation is the analogue of what in classical neural networks is named the \textbf{bias}. Given a solvable group element 
      $\mathbb{L}(\mathbf{u})\in \mathcal{S}_{r,r+q}$ identified by the solvable coordinate vector $\mathbf{u}$  we can perform the left translation: 
      \begin{eqnarray}\label{pantofole}
       \forall \,\mathbb{L}(\mathbf{u}) \in \mathcal{S}_{r,r+q} \quad\quad \quad\quad \ell_{\mathbf{u}}&  : & \mathcal{S}_{r,r+q} \, 
       \longrightarrow \, \mathcal{S}_{r,r+q} \nonumber\\
       \forall \,\mathbb{L}(\mathbf{w}) \in \mathcal{S}_{r,r+q} \quad \ell_{\mathbf{u}} \left[\mathbb{L}(\mathbf{w})\right]& = &
        \mathbb{L}(\mathbf{u})\cdot \mathbb{L}(\mathbf{w})\, = \, \mathbb{L}(\mathbf{u}\cdot\mathbf{w}) \nonumber\\
        \mathbf{u}\cdot\mathbf{w} & \equiv & \Sigma^{-1} \left[\mathbb{L}(\mathbf{u})\cdot \mathbb{L}(\mathbf{w})\right]
      \end{eqnarray}
      The definition of the product law in the last line of the above equation (\ref{pantofole}) is general. For higher non-compact ranks $r>1$ it will be rather involved. In the case $r=1$ it is simple and is given by equation (5.5) of \cite{pgtstheory}. We repeat it below:
  \begin{equation}\label{prunello1}
  \begin{array}{rcl}
 (u\cdot w)_1& = & w_1\, +\, u_1 \\
 (u\cdot w)_{1+i}& = & w_{1+i}\, +\, \exp\left[-u_1 \, - \, w_1\right]\, u_{1+i} \quad ; \quad i=1,\,   \dots, q+1\\
\end{array} 
\end{equation}  
  \item[c)] The third operation is the most complex, yet crucial for the neural network algorithm construction. It is named the fiber rotation and its theory was developed in a full-fledged way in section 6.3 of \cite{pgtstheory}. Summarizing the discussion of that paper, for all non-compact symmetric spaces $\mathrm{U/H}$ of the series in eq.(\ref{Mrq}) there exists a compact subgroup $\mathrm{G_F}\subset \mathrm{H}$ isomorphic to $\mathrm{SO(r+q-1)}$ and a Grassmaniann compact coset manifold:
  \begin{equation}\label{crinolina}
  \mathcal{F}_{{TS}} \, = \,  \frac{\mathrm{SO}(r+q-1)}{\mathrm{ SO}(r) \times \mathrm{SO}(q-1)} \, = \, \frac{\mathrm{G_F}}{\mathrm{H_F}}
  \end{equation}
whose tangent vector fields, namely the coset generators $\mathbb{F}_{\mathrm{TS}}$ in the orthogonal decomposition of the corresponding Lie algebras: 
\begin{equation}\label{corsarobianco}
 \mathbb{G}_{\mathrm{F}} \,\equiv \, \mathbb{H}_{\mathrm{F}}\oplus \mathbb{F}_{\mathrm{TS}} \sim \so(r+2s-1)
\end{equation} 
are as many as the dimension of the Tits-Satake fiber vector space and in fact generate it while acting on a solvable group manifold reduced to the Tits-Satake submanifold. The fiber rotation generators also act on any solvable group element and have the distinctive property of mixing the Cartan coordinates with the fiber ones. In the case of $r=1$ manifolds 
the Grassmannian (\ref{corsarobianco}) reduces to a sphere $\mathbb{S}^{q-1}$, that has the same dimension as the \textbf{subPaint vector} 
$\mathfrak{w}$. Therefore the elementary rotations of  $\mathbb{F}_{\mathrm{TS}}$ that generate the whole sphere $\mathbb{S}^{q-1}$ can be chosen in such a way that each time they mix the Cartan coordinate $w_1$ with one of the components of the subPaint vector $\mathfrak{w}$. In appendix D.2 of the twin paper \cite{naviga} we have presented the derivation of the individual fiber rotations for the case $r=1$  
\end{description}
The three mentioned operations correspond to the first three cases of table (\ref{terravuoto}) and, conceptually, they are all of the same type, namely they all pertain to the adjoint action (viewed as active) of the isometry group $\mathrm{SO(1,1+q)}$ on the points of the $i$-th layer. Note also that the first and second operations of table (\ref{terravuoto}) have been interchanged for a reason that we explain below, although conceptually they are better ordered in eq.(\ref{terravuoto}). For convenience the
adjoint action of the isometry group of each layer has been separated in three components (bias-translation, Paint-rotation, fiber-rotation), and in the machine learning styled paper \cite{naviga} they have been exhibited as formulae of the type:
\begin{equation}\label{codicegenetico}
  \boldsymbol{\Upsilon}^\prime \, = \, \mathit{f}_g(\boldsymbol{\Upsilon})
\end{equation}
yet the specific form of the function $\mathit{f}_g$ has always to be viewed as the result of the abstract map
defined in the r.h.s. of eq. (\ref{tarantola}) for different choices of the group element $g\in \mathrm{SO(1,1+q)}$, that can be,
either in the solvable subgroup, or in the Paint subgroup, or a representative of the Grassmannian coset.
\subsubsection{Homomorphisms}
Indeed the question remains how do we make the transition from one layer to the next one, namely how do we jump from a hyperbolic space $\mathbb{H}^{(1+q)}$ to a next one $\mathbb{H}^{(1+q^\prime)}$? As we know from the general scheme outlined above, the map from one  symmetric space to the next is given by 
\begin{equation*}
\varpi_{i+1}^{-1}  \circ\mho_{\mathcal{W}_i}\circ\varpi_i
\end{equation*}
where $\mathcal{W}_i $ is the linear homomorphism between the $i$-th and the ($i+1$)-th solvable Lie algebra. In the special case where both the $i$-th and $i+1$-th Lie algebras are $r=1$ Lie algebras, the Maurer-Cartan equations are very simple. Equations (\ref{pignatta}-\ref{casseruola}) can be reduced to the following form:
\begin{eqnarray}\label{balordus}
  dE^1 &=&0 \nonumber\\
  dE^{1+i}&=& E^1 \wedge E^{1+i} \quad ; \quad i=1,\dots, n \nonumber\\
  de^1 &=&0 \nonumber\\
  de^{1+\alpha}&=& e^1 \wedge e^{1+\alpha} \quad ; \quad \alpha=1,\dots, m \nonumber.\\
\end{eqnarray}
To impose the group homomorphism conditions, we need to impose that the Maurer-Cartan equations \ref{balordus} are respected by the pullback. Let then $G_i, i=1,2$ two spaces with $r=1$ and consider the map $h: G_1 \rightarrow G_2$. Let the MC forms of $G_1$ be $E^i$ and the MC forms of $G_2$ be $e^i$. What we require is that the pullback of the forms of $G_2$ obeys equations \ref{balordus}, i.e.
\begin{equation}
    d(h^*(e^1)) = 0
\end{equation}
and 
\begin{equation}
    d(h^*(e^{1+j})) = h^*(e^1) \wedge h^*(e^{1+j})
\end{equation}
By developing the computations expressed in \ref{carriola} and carried out in App. \ref{app:maurercartan}, we find the general form in coordinates for a homomorphism of \cite{naviga}:
\begin{equation}\label{eq:homo}
h(\Upsilon)=\begin{bmatrix}  \Upsilon_1 \\ W \boldsymbol{\Upsilon_2} + (1-e^{-\Upsilon_1})\, \boldsymbol{b}\end{bmatrix}  = \begin{bmatrix}1  & 0 \\0 & W\end{bmatrix}\begin{bmatrix}\Upsilon_1 \\\boldsymbol{\Upsilon_2} \end{bmatrix} + (1-e^{-\Upsilon_1})\begin{bmatrix}
    0 \\ \boldsymbol{b}
\end{bmatrix}
\end{equation}
where
$W \, \in \, \mathbb{R}^{(1+q_{i+1}) \times (1+q_{i})}$ and $b \in \mathbb{R}^{1+q_{i+1}}$
Equation \ref{eq:homo} is almost linear in the paint coordinates. This absolutely exceptional feature of the $r=1$ solvable Lie algebras is what provides the possibility of combining a Paint group rotation in the target Lie Algebra of dimension
$n=1+q_{i+1}$ that corresponds to the application of a 
matrix of the following type:
\begin{equation}\label{crinetto}
  W_{Paint}\, = 
  \begin{bmatrix}
      1 & \boldsymbol{0} \\
      \boldsymbol{0} & \mathcal{O}
  \end{bmatrix}
\end{equation}
with a generic matrix of the form
\begin{equation}\label{patrullo}
   W_{homo} \, =  
  \begin{bmatrix}
      1 & \boldsymbol{0} \\
      \boldsymbol{0} & W
  \end{bmatrix}\end{equation}
which is included in \ref{eq:homo}. Indeed we can always set
\begin{equation}\label{cagliostro}
  W_{homo}^\prime \, = \, 
  W_{homo}\cdot W_{Paint} 
\end{equation}
and, as long as the parameters $\mathbf{w}$ and $\mathcal{T}$ are free to be learned by the network during its training, there is no point in performing an additional paint group rotation after the essential $W_{homo}$ which encodes the transformation from a layer to the next one as we have pointed out. This is the essential reason why the \textbf{bias translation} and the \textbf{generalized Paint Group Rotation} have been interchanged with respect to the natural order of table (\ref{terravuoto}).
In all other cases the relation between the solvable coordinates of two non-isomorphic solvable manifolds is non linear and
provided by iterative   solution of eq.s (\ref{pergolato})  as defined in eq.(\ref{carriola}).

Notice that in Equation \ref{eq:homo}, $W$ and $b$ are completely free, and the possibility that entire rows of $W$ might be put to zero implies that the actual dimensions of the target space can be \textit{learned} by the network during its training process. This is a possibility that is much more reduced in higher rank cases where the  \textbf{consistency conditions to be a  homomorphism}  are more 
restrictive, as we illustrate in appendix \ref{patagonia}. On the contrary, the freedom in the matrix $W_{homo}$ is counterbalenced
in the $r=1$ case by the already pointed out existence of a unique truely exponential function which means to decide a priori the existence of a unique dominant feature and by the linearity of the relation between solvable coordinates that further reduces the expressivity of a network free from point-wise activation functions.

\section{The general form of the \texorpdfstring{$\slal(\mathrm{N},\mathbb{R})$}{SL(N,R)} solvable Lie algebra}
\label{sandalo} In this section, we develop the notions required to build homomorphisms between manifolds of the type $\mathrm{SL(N)}/\mathrm{SO(N)}$. In order to study the general form of the $\slal(\mathrm{N},\mathbb{R})$ solvable Lie algebra, namely the Borel
subalgebra $\mathbb{B}(\mathrm{N},\mathbb{R})\subset \slal(\mathrm{N},\mathbb{R})$, and plan a systematic study of its subalgebras and of their immersions, it is
convenient to start from the general form of a simple Lie algebra in the Cartan-Weyl basis, namely from the following equations. 
\begin{eqnarray}
\left[  H_i \, , \, H_j \right]  & = & 0 \nonumber\\
\left[ H_i \, , \, E^\alpha \right]  & = & \alpha_i \, E^\alpha \nonumber\\
\left[  E^\alpha \, , \, E^{-\alpha} \right]   & = & \alpha^i \, H_i \nonumber\\
\left[  E^\alpha \, , \, E^\beta \right]  & = & \left\{\begin{array}{lcl} N(\alpha , \beta ) \, E^{\alpha +
\beta}
& \mbox{if} & \alpha + \beta \in \pmb{\Phi}\\
 0 & \mbox{if}   &  \alpha + \beta \notin
\pmb{\Phi} \\
\end{array}\right.\label{CarWeylform}
\end{eqnarray}
where $H_i$ are a basis $i=1,\dots, \ell$ of Cartan generators, $E^\alpha$ are the step operators associated with the roots $\alpha$
and the latter form a root system $\boldsymbol{\Phi}$ (see the textbook \cite{pietro_discrete} for all definitions and theorems) that can always be split into the subset of the positive roots and that of the negative ones: 
\begin{equation}\label{ruttosistemo}
   \pmb{\Phi} \, = \, \pmb{\Phi}_+ \, \bigcup \, \pmb{\Phi}_- 
\end{equation}
each containing the same number of roots since every negative root is just $-\alpha$ where $\alpha >0$ is positive and viceversa.
In eq.(\ref{CarWeylform})  $N(\alpha,\beta)$ is a coefficient that has to be determined
using Jacobi identities. 
\par
Then in commutator presentation the solvable Borel subalgebra contains all the Cartan generators and all the step operators $E^\alpha$ associated with positive roots $\alpha >0$. Furthermore let us recall that for the simple Lie algebra $\slal(\mathrm{N},\mathbb{R})$, which is maximally split, the non-compact rank $r$ equals the rank $ \ell = \mathrm{N}-1$:
\begin{equation}\label{granulato}
  r \, = \, \ell \, = \, \mathrm{N}-1
\end{equation}
\par
In order to write the general form of the Maurer-Cartan equations for the Borel algebra we have to recall the general form of the root system for 
the Lie algebra $\mathfrak{a}_\ell \, = \, \slal(\ell+1,\mathbb{R})$. We do this summarizing  the exposition in section 11.5.1 of \cite{pietro_discrete}. 
\par The Dynkin diagram that encodes the very definition of the simple Lie algebra by means of its Cartan matrix  is that recalled in 
fig.\ref{Aldiag}. The explicit construction of a root system that admits a basis corresponding to such a diagram is done in the following way. We 
consider the $\ell +1$--dimensional Euclidean space $ \mathbb{R}^{\ell +1} $ and let $ {\epsilon} _1$, $\dots$ , $ {\epsilon} _{\ell +1}$ denote 
the unit vectors along the $\ell + 1$ axes: 
\begin{equation} 
   {\epsilon}_1 = \left( \begin{array}{c}
    1 \\
    0 \\
    \dots \\
    \dots \\
    0 \
  \end{array}\right)  \quad , \quad  {\epsilon} _2 = \left( \begin{array}{c}
    0 \\
    1 \\
    0 \\
    \dots \\
    0 \
  \end{array}\right)  \quad \dots \quad  {\epsilon}_{\ell+1} = \left( \begin{array}{c}
    0 \\
    0 \\
    \dots \\
    \dots \\
    1 \
  \end{array}\right)
\label{unitvecteps}
\end{equation}
\par
Given  the vector $ {v}=\epsilon _1 + \epsilon _2 + \dots +  {\epsilon} _{\ell+1}$ 
we define $\mathbb{I} \subset \mathbb{R}^{\ell +1} $ to be the $\ell + 1$ --dimensional cubic lattice immersed
in $ \mathbb{R}^{\ell +1} $:
\begin{equation}
  \mathbb{I}= \left\{  {x} \in \mathbb{R}^{\ell +1 } \, /
  \,  {x} = n^i \,  {\epsilon} _i \, ,
  \quad n^i \in \mathbb{Z} \right\}
\label{Ilatdefi}
\end{equation}
In the cubic lattice $ \mathbb{I}$ we consider
the sublattice:
\begin{equation}
  \mathbb{I}^\prime = \mathbb{I} \, \bigcap \, E
\label{sublatt}
\end{equation}
where $E$ is the hyperplane of vectors orthogonal to the vector $ {v}$:
\begin{equation}
  E = \left\{  {y} \in \mathbb{R}^{\ell +1 } \, / \, \left(
   {v} \, , \,  {y} \right) =0 \right\}
\label{Edefiort}
\end{equation}
Finally in the sublattice $\mathbb{I}^\prime$ we consider the finite set of vectors whose norm is $\sqrt{2}$:
\begin{equation}
  \pmb{\Phi}=\left\{   {\alpha} \in \mathbb{I}^\prime \, / \, \left(  {\alpha}
  \, , \,  {\alpha} \right) =2 \right\}
\label{Phisystdef}
\end{equation}
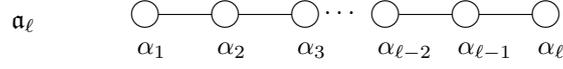
\begin{figure}
\centering
\begin{picture}(100,200)
\put (-70,185){$\mathfrak{a}_\ell$} \put (-20,190){\circle {10}} \put (-23,175){$\alpha_1$} \put
(-15,190){\line (1,0){20}} \put (10,190){\circle {10}} \put (7,175){$\alpha_2$} \put (15,190){\line
(1,0){20}} \put (40,190){\circle {10}} \put (37,175){$\alpha_3$} \put (47,190){$\dots$}
\put (70,190){\circle {10}} \put (67,175){$\alpha_{\ell-2}$} \put (75,190){\line (1,0){20}} \put
(100,190){\circle {10}} \put (97,175){$\alpha_{\ell-1}$} \put (105,190){\line (1,0){20}} \put
(130,190){\circle {10}} \put (127,175){$\alpha_\ell$}
\end{picture}
\vskip -6cm \caption{\label{Aldiag} The Dynkin diagram\index{Dynkin
diagram} of $\mathfrak{a}_\ell$ type}
\end{figure}
As proved in \cite{pietro_discrete} the above defined set $\pmb{\Phi}$ is a root system and it corresponds to the $\mathfrak{a}_\ell$ Dynkin diagram. 
We have:
\begin{equation}
   {\alpha}  \, \in \, \pmb{\Phi} \, \Rightarrow \, \left\{  \begin{array}{crclcrcl}
    1) &  {\alpha } & = & n^i \,  {\epsilon} _i & \null & n^i & \in & \mathbb{Z} \\
    2) & \left(  {\alpha} \, , \, { v}\right)  & = & 0 & \Leftrightarrow &
    \sum_{i=1}^{\ell +1} n^i& = & 0 \\
    3) & \left(  {\alpha} \, , \,  {\alpha} \right)  & = & 2 &
    \Leftrightarrow & \sum_{i=1}^{\ell + 1} \left( n^i\right) ^2& = & 2 \
  \end{array} \right.
\label{properteni}
\end{equation}
and the above diophantine equations have the following solutions:
\begin{equation}
   {\alpha} =  {\epsilon} _i -  {\epsilon}
  _j \quad \left( i \ne j\right)
\label{diofeqsol}
\end{equation}
The number of such solutions is equal to twice the number of pairs $(ij)$ in $(\ell +1)$--dimensions, namely 
\begin{equation}\label{numerokorni}
  \text{\# of roots} \, = \, \ell \left( \ell + 1\right)  \, = \, \mathrm{N(N-1)}
\end{equation}
\noindent In this way, in $ \mathbb{R}^{\ell +1}$ we have constructed the
root system of the $\slal\mathrm{(N,\mathbb{R})}$ Lie algebra.  The  simple  roots are the following ones
\begin{equation}
  \alpha_i =  \epsilon _i - \epsilon _{i+1}  \quad ,
  \quad \left( i=1,\dots , \ell = \mathrm{N -1}\right)
\label{simplerootte}
\end{equation}
The set of roots is split into the subsets of  positive and negative ones according to the following rule:
\begin{equation}
  \begin{array}{rclcc}
    \pmb{\Phi} & = & \pmb{\Phi}_+ \, \bigcup \, \pmb{\Phi}_- & \null & \null \\
    \alpha  & \in  & \pmb{\Phi}_+ & : & \left\{ \alpha = \epsilon _i -\epsilon _j \quad , \quad i<j\right\} \\
    \alpha  & \in & \pmb{\Phi}_- & : & \left\{ \alpha = \epsilon _j -\epsilon _i \quad , \quad i<j\right\} \
  \end{array}
\label{fiplusminus}
\end{equation}
Positive roots that are our main target to study the structure of solvable Lie algebras are written as follows:
\begin{equation}
  \alpha = \epsilon _i - \epsilon _j = \alpha _i + \alpha _{i+1} +
  \dots + \alpha _{j-1}
\label{positrootte}
\end{equation}
The above equation (\ref{positrootte}) is the most relevant one to encode in an iterative algorithmic way all the roots. Indeed because of the 
very simple general structure of the roots they can be identified by a pair of integer numbers in the following way. First, we define the 
\textbf{height} of a root which is the number of addends in the sum (\ref{positrootte}), as below: 
\begin{equation}\label{altezza}
  h[\alpha _i + \alpha _{i+1} +
  \dots + \alpha _{i+m-1}]\, = \, m 
\end{equation}
Next we introduce an integer $k \in \mathbb{N}$ whose possible values are:
\begin{equation}\label{grappolo}
  k \, = \, 1, \dots , \ell-h+1
\end{equation}
The number $k$ tells us the simple root from which we have to start the sum. Hence we can write the correspondence:
\begin{equation}\label{tertulliano}
 \left[h,k\right] \, \Rightarrow \, \alpha_k + \alpha_{k+1} + \dots + \alpha_{k+h-1}
\end{equation}
The $\left[h,k\right] $ labeling identifies all the roots but the nice thing is that we can extend it to identify also the Cartan generators 
considering them the roots of height $h=0$. Since the Cartan generators can be put into one-to-one correspondence with the simple roots in such a way that
\begin{equation}\label{cortoloso}
  \alpha_i \left(H_j\right) \, = \, \delta_{ij}
\end{equation}
we can introduce a uniform set of $1$-form generators $\mathfrak{E}^{[h,k]}$ with:
\begin{equation}\label{remagio}
  h \, = \, 0, \dots, \ell \quad ; \quad k\, = \, \left\{ \begin{array}{lcrcl}
  1, \dots, \ell & \text{if} & h & = & 0 \\
  \, 1, \dots , \ell-h+1 & \text{if} & h & > & 0\\
  \end{array}\right.
\end{equation}
Then the rules to write the Maurer-Cartan equations that follow from eq.s (\ref{CarWeylform}) are very simple.
The quadratic contributions $\mathfrak{E}^{[h_2,k_2]} \wedge \mathfrak{E}^{[h_3,k_3]} $ to the differential 
$\mathrm{d}\mathfrak{E}^{[h_1,k_1]} $ in the equation:
\begin{equation}\label{generatus}
  \mathrm{d}\mathfrak{E}^{[h_1,k_1]} + \dots + \mathfrak{E}^{[h_2,k_2]} \wedge \mathfrak{E}^{[h_3,k_3]} + \dots \, = \, 0
\end{equation}
must fulfill the following integer constraints:
\begin{equation}\label{fargnacco}
    h_1 \, = \, h_2+h_3 \quad ; \quad k_1 \, = \, k_2 \quad ; \quad  k_3 \, = \, k_2 + h_2
\end{equation}
It follows that all the generators of height $h=0$, namely the Cartan generators, are closed:
\begin{equation}\label{primula}
  \mathrm{d}\mathfrak{E}^{[0,k]} \, = \, 0 \quad ; \quad k \, = \, 1,\dots,\ell
\end{equation}
For those of height $h=1$, namely the simple roots, we have:
\begin{equation}\label{azalea}
  \mathrm{d}\mathfrak{E}^{[1,k]} + \mathfrak{E}^{[0,k]} \wedge \mathfrak{E}^{[1,k]} \, = \, 0 \quad ; \quad k \, = \, 1,\dots,\ell
\end{equation}
and so on. Clearly for large $\ell$ the general form of the Maurer-Cartan equations is quite long, yet it can be iteratively constructed 
as the solution of the combinatorial problem (\ref{fargnacco}) which is very much suited for computer calculations. The study,
for generic $\ell = N-1$, of the nested immersion of Borel algebras one into the other is the next problem in our agenda to explore the expressivity of Cartan neural networks.  

\section{The Separators and the Classification task}
\label{cicero} One of the main tasks in the context of supervised deep learning is \textbf{classification}, namely assigning one of several 
classes to an input. In our previous work on Cartan networks, we tested our architecture for Cartan networks both on classification and regression 
tasks (see section 4 of \cite{naviga} entitled Results). In this section, we present a conceptual analysis of the classification task, focusing on 
the interplay between the two fundamental components of deep learning: the differential geometric and the statistical-probabilistic aspects. We 
begin with a general overview of the logistic regression used in machine learning, focusing on its probabilistic and geometric interpretation 
(already extensively discussed in the literature, see \cite{bridle_ce, bishop_pattern_2006, murphy_ml} and the explanations provided in chapter 4 of the book \cite{daniwitten}), aimed at the critical structural analysis, for the benefit of math/phystheo readers, of the practices commonly 
employed in the ML literature. Next, we will generalize such procedures to our manifolds. Classification in non-Euclidean spaces has been 
discussed in the context of hyperbolic deep learning \cite{ganea_hyperbolic_2018,shimizu_2021,chen-etal-2022-fully, peng_hyperbolic_2022, bdeir_fully_2024}. 
\par
For the math/phystheo readers to get a full appreciation of the above mentioned elements,  it is quite important to recall, 
in a dedicated appendix of the present paper, the  basic principles of probability theory, with the explicit purpose of highlighting, within the general mathematical theory,  those critical points on which the current 
ML constructions rely on. Such points may not be obvious to mathematically oriented minds, and they require due emphasis. Reciprocally, the ML readers will benefit from our discussion by reviewing what is often taken for granted.  
\par  
So let us recall that in the modern systematic formulation of probability theory, as presented in a textbook such as \cite{sinaikarolo} or in 
the lectures \cite{lychaginlecture}, the starting point is the definition of a probability measure on a space of outcomes $\Omega$. From a mathematical point of view, there is a cross-link with the general theory of \textit{measure}, as introduced in standard Functional Analysis textbooks (see, for instance, the textbook by Guido Fano \cite{fanometodo})  when discussing Lebesgue integrals, but there are two possible sources of confusion. One is that the space $\Omega$ can be a mixed object, for example, a product of a finite discrete set and a continuous manifold. The second caveat is that the notion of $\sigma$-algebra is more general than the notion of a basis of neighborhoods defining a Hausdorff separable topology, and, for this reason, it can flexibly accommodate such mixed outcome spaces $\Omega$. Hence, before proceeding further, we refer the reader to appendix \ref{probaprincipe} for a concise summary of the basic principles of probability theory propedeutic to the next section. 
\subsection{Conceptual analysis of the classification task}
As we stressed in the introduction to the present section \ref{cicero}, the reason why we inserted the previously mentioned appendix \ref{probaprincipe} is the peculiarity of the \textit{space of outcomes} that underlies both the procedures respectively named \textit{logistic 
regression} and its generalization \textit{softmax}. The peculiarity of the space $\Omega$ in this formulation is precisely its composite nature as the direct product of a finite set with a differentiable manifold, still consistent with the construction of a suitable $\sigma$-algebra.     
\subsubsection{The classification task in general}
\label{classtask} Consider a stochastic vector $\mathbf{X}\in \mathbb{R}^n$ which represents the coordinate of a point $p \in \mathcal{M}$ in a 
differentiable manifold $\mathcal{M}$ of dimension $n$, and another stochastic variable $Y\in \mathfrak{C}$, where $\mathfrak{C}$ is a finite set 
containing $K$ abstract elements that we call \textbf{classes} $\mathfrak{c}[a]$ ($a=1,\dots K$). 
\par
The space of \textbf{outcomes} is the direct product:
\begin{equation}\label{probspace}
  \Omega \, \equiv \, \mathfrak{C} \times \mathcal{M}
\end{equation}
namely every possible outcome is a pair $\{\mathfrak{c}[a], p\in\mathcal{ M}\}$. One ought to construct a suitable $\sigma$-algebra on such a mixed space, in order 
to introduce a \textbf{probability measure} for \textbf{events} that satisfies the axioms of definition \ref{misaproba} and is 
constructed from some suitable elementary function defined over individual outcomes. 
\par
For instance,  $\mathbf{X}: \Omega \rightarrow \mathbb{R}^n$  (with $n=28\times28$)  can be the vector of pixels for a photo displaying a 
hand-written digit from 0 to 9 and the correlated stochastic variable $Y$ can assume precisely $K=10$ different values (like in the popular MNIST 
database \cite{lecun1998gradient}). In this example, the classification task is to determine the probability that a point $p\in \mathcal{M}$ 
corresponds to the image of a given digit. Once these class probabilities are established for each data point, the predicted class will be the one 
with the highest probability. 
\subsubsection{The logistic regression in the \texorpdfstring{$K=2$}{K=2} case and the sigmoid}
\label{sigmarecedi}
 Logistic regression is the name of a widely used statistical model parameterizing the \textit{probability measure} discussed in the previous 
 subsection when there are only two classes,  coded by a single binary dependent variable assuming values 0 or 1. (see appendix \ref{historsigma}
 for a historical note on the sigmoid and on the logistic regression).
 To this end, we introduce the sigmoid function: 
 \begin{eqnarray}
   \sigma(x) & \equiv & \frac{1}{1\, + \, e^{-x}} 
\end{eqnarray}
 that maps the whole real line $\mathbb{R}$ into the open interval $]0,1[$ and has the following two obvious properties:
 \begin{equation}\label{sigma_properties}
 \sigma(-x) \, = \, 1\, - \, \sigma(x)  \quad ; \quad \sigma(x) \, = \, \frac{e^{x}}{1+ e^x} 
 \end{equation} 
 The former property is important for a \textbf{categorical probabilistic interpretation} since in binary classification the sigmoid can be used to model the probability of one of two possible outcomes, with the complementary being $\sigma(-x)$. The sign of the argument can be taken as a discriminant so that the probability of the outcome \textbf{0} ($x<0$)  is the complement of the probability of the outcome \textbf{1} ($x>0$). 

Let us consider the general framework introduced at the beginning of Section \ref{classtask}, and, for now, assume that the manifold $\mathcal{M}$ referenced in Equation \ref{probspace} coincides with the Euclidean space $\mathbb{E}^n = \mathbb{R}^n$. Given an outcome $(\mathbf{x}, \,y)$, with $\mathbf{x} \in \mathbb{R}^n$ and $y \in \{0, 1\}$, we model its probability using the sigmoid function. This probability depends only on the input $\mathbf{x}$ and a set of parameters $\mathbf{w}$ and $b$, as shown below:
\begin{eqnarray}\label{ramarro}
\mathfrak{p}\left(Y=1 \mid \mathbf{X}=\mathbf{x},\,\mathbf{w},\, b \right) \equiv \mathfrak{p}\left( \mathbf{x};\,\mathbf{w},\, b \right) \equiv \sigma\left( \mathbf{w} \cdot \mathbf{x} + b \right)
\end{eqnarray}
where $\mathbf{w} \in \mathbb{R}^n$ and $b \in \mathbb{R}$. Notice that this corresponds to assigning each point based on the sign of the argument of the sigmoid:
 \begin{equation}\label{commodoro}
   \begin{array}{ccccccccccc}
      \mathbf{w} \cdot\mathbf{x} + b & > & 0 & \Leftrightarrow & \mathfrak{p}\left( \mathbf{x};\,\mathbf{w},\, b\right) 
      & > & \frac{1}{2} & \Rightarrow & \hat y  & = &1 \\ 
       \mathbf{w} \cdot\mathbf{x} + b & < & 0 & \Leftrightarrow & \mathfrak{p}\left( \mathbf{x};\,\mathbf{w},\, b\right)
      & < & \frac{1}{2} & \Rightarrow & \hat y  & = &0 \\ 
      \end{array}
\end{equation}

where $\hat y$ is the class predicted for point $\mathbf{x}$.

The problem is the following:  given a set of known outcomes (or observations) $\mathcal{D} = \bigcup_{i=1}^N\left\{(\mathbf{x}_i, \,y_i)\right\}$, \textit{i.e.} the training set, can we optimize the choice of the parameters $\mathbf{w},b$ in such a way that the probability measure (\ref{ramarro}) agrees as much as possible with these known outcomes and enables us to make sound predictions on the future outcome $y$ in terms of the available choices of $\mathbf{x}$? 
We do this by extremizing the likelihood functional.

\paragraph{The likelihood functional.} 
We subdivide the set of known outcomes $\mathcal{D}$  in the two classes:
\begin{equation}\label{cronogatto}
\begin{array}{rclcrcl}
  \mathcal{D} & = &  \bigcup_{i=1}^N\left\{ (\mathbf{x}_i,\,y_i)\right\}  &  ;  &\mathcal{D} & = &\mathcal{C}_1 \bigcup \mathcal{C}_2 \nonumber\\
  \mathcal{C}_1 & = & \langle (\mathbf{x},\,y) \in \mathcal{D}  \, \mid \, y \, = \, 1 \rangle & ; &
   \mathcal{C}_2 & = & \langle (\mathbf{x},\,y) \in \mathcal{D}  \, \mid \,y \, = \, 0 \rangle 
   \end{array}
\end{equation}
and we introduce the likelihood functional defined as follows:
\begin{equation}\label{gelsominobrutto}
  \mathcal{L}(\mathbf{w},b) \, \equiv \,  \prod_{(\mathbf{x},\,y)\in \mathcal{C}_1} \mathfrak{p}\left( \mathbf{x};\,\mathbf{w},\, b\right)\, \times \, \prod_{(\mathbf{x},\,y)\in \mathcal{C}_2}  \,\left[1-\mathfrak{p}\left( \mathbf{x};\,\mathbf{w},\, b\right)\right]
\end{equation}
The possible values of the functional $\mathcal{L}(\mathbf{w},b) $ are in the interval $[1,0]$ since it is a product of probabilities. If the 
predicted probability of all outcomes in class $\mathcal{C}_1$ were $1$ and the predicted probability of all outcomes in class $\mathcal{C}_2$ were $0$, which is the ideal correct result, we would obtain $\mathcal{L}(\mathbf{w},b)\, = \, 1 $. Instead, if the predicted probabilities were maximally wrong, then we would obtain $\mathcal{L}(\mathbf{w},b)\, = \, 0 $. Hence, extremizing $\mathcal{L}(\mathbf{w},b) $ (or its logarithm) is a convenient strategy to optimize the parameters $\mathbf{w},b$:
\begin{eqnarray}\label{amatriciana}
  \log \left[\mathcal{L}(\mathbf{w},b)\right] & = &  \sum_{(\mathbf{x},\,y)\in \mathcal{C}_1} 
  \log \left[\mathfrak{p}\left( \mathbf{x};\,\mathbf{w},\, b\right)\right] \, + \,\sum_{(\mathbf{x},\,y)\in \mathcal{C}_2} 
  \log \left[1-\mathfrak{p}\left( \mathbf{x};\,\mathbf{w},\, b\right)\right]\nonumber\\
  0 & = & \partial_\mathbf{w} \log \left[\mathcal{L}(\mathbf{w},b)\right] \, = \, \partial_b 
  \log \left[\mathcal{L}(\mathbf{w},b)\right]
\end{eqnarray}

\subsection{Geometric interpretation of the argument of the  logistic function}
\label{logigeometra}
Let us consider the argument of the sigmoid in eq.(\ref{ramarro}) 
\begin{equation}\label{carciofone}
  \mathfrak{arg}(\mathbf{x})\, =\, \mathbf{w} \cdot\mathbf{x} + b
\end{equation}
The discriminant between the two classes corresponds to the vanishing of $\mathfrak{arg}(\mathbf{x})$. What does it mean in Euclidean geometry the equation $\mathfrak{arg}(\mathbf{x}) \, = \, 0$? It is the equation of a hyperplane $\mathfrak{S}$:
\begin{equation}\label{carciofino}
  \mathfrak{S}(\mathbf{w},b) \, = \, \langle \mathbf{x}\in \mathbb{R}^n \, \mid \,  \mathfrak{arg}(\mathbf{x}) \, = \, 0 \rangle
\end{equation}
The hyperplane $\mathfrak{S}(\mathbf{w},b) $ is a separator since it produces a partition of the full manifold $\mathbb{R}^n$ into two  halves:
\begin{equation}\label{separatologo}
  \mathbb{R}^n = \mathbb{R}^n_+ \bigcup\mathbb{R}^n_-
\end{equation}  

given that

\begin{equation}
\mathbb{R}^n_+ \bigcap \mathbb{R}^n_- =
  \mathfrak{S}(\mathbf{w},b) \subset \mathbb{R}^n  \quad \text{with}\quad \text{dim}\mathfrak{S}(\mathbf{w},b) \, = \, n-1 \nonumber 
\end{equation}

and

\begin{equation}
  \mathbb{R}^n_+ = \langle \mathbf{x}\in \mathbb{R}^n \mid \mathfrak{arg}(\mathbf{x}) \geq 0 \rangle \quad\text{and}\quad
  \mathbb{R}^n_- = \langle \mathbf{x}\in \mathbb{R}^n \mid \mathfrak{arg}(\mathbf{x}) \leq 0 \rangle \nonumber
\end{equation}

Furthermore for those $\mathbf{x}$ such that $\mathfrak{arg}(\mathbf{x}) \neq 0$ the function $\mathfrak{arg}(\mathbf{x})$ is proportional to the \textbf{oriented distance} of the point $\mathbf{x}$ from the separator hyperplane $\mathfrak{S}(\mathbf{w},b)$. We consider the point $\mathbf{u} \in \mathfrak{S}(\mathbf{w},b)$  that is at the minimal distance from
$\mathbf{x}$ and the oriented geodesic going from $\mathbf{u}$ to $\mathbf{x}$, whose length is the absolute value of such a distance while the sign is the orientation with respect to the normal vector to the separator hyperplane that is decided implicitly by equation  $\mathfrak{arg}(\mathbf{x}) \, = \, 0$ as in (\ref{carciofino}). Then the absolute value of the distance is $\mathfrak{arg}(\mathbf{x})/\|\mathbf{w}\|$. 

\subsection{Generalization of the separator in a smooth manifold with distance function}
\label{separatoincasa}
It is now the occasion to point out again the role played by \textbf{the four principles} listed in the introduction in choosing non-compact symmetric spaces to formulate new neural network architectures. Indeed, it is essential to require the existence of a distance function on the manifolds used for the network layers. This requirement excludes all Riemannian manifolds not endowed with a notion of absolute distance. The existence of a well-defined absolute distance is an exception in Riemannian geometry, and the \textbf{non-compact symmetric spaces} claimed by principles 1 and 2 are the main body of such an exception among \textbf{homogeneous spaces}. They are all \textbf{normal homogeneous spaces} according to Alekseevsky's definition (see \cite{pgtstheory}). The need for a distance becomes clear in the classification task when we address the issue of separators. The stochastic vector $\mathbf{X}$ corresponds to the coordinates of a point $\mathbf{p}\in \mathcal{M}_{N}$, the manifold $\mathcal{M}_{N}$ being the \textbf{last hidden layer of a Cartan neural network} that we need to subdivide into classes. The full scheme of the Cartan neural network algorithm in the case of the $K=2$ classification task can be viewed as follows:
\begin{equation}\label{formaggio}
 \mathbb{V}_{\text{input}} \, \underbrace{\stackrel{\iota_{[\mathcal{Q},\Lambda]}}{\hookrightarrow}}_{\text{injection}} \,\underbrace{\mathcal{M}_1 \, \stackrel{\hat{\mathcal{K}}^1_{[\mathcal{W}_1,\Psi_2]}}{\longrightarrow} \, \mathcal{M}_2\, \longrightarrow \,\dots \, 
 \longrightarrow \,\mathcal{M}_{N}}_{\text{Cartan neural network hidden layers}} \, \underbrace{\stackrel{\mathfrak{S}}{\longrightarrow}\, \mathcal{M}_+\cup \mathcal{M}_-}_{\text{partition}} \underbrace{\, \stackrel{\sigma}{\longrightarrow}\,\left[0,1\right]_{\text{out}}}_{\text{log. regr.}}
\end{equation}
The first step consists of the injection of the datum vector into the first manifold that is parameterized by a matrix $\mathcal{W}_0$ that associates the input vector $\boldsymbol{\Xi}$ with the solvable coordinates $\boldsymbol{\Upsilon}$ of the first layer:
\begin{equation}\label{imputto}
  \Upsilon_A \, = \, \mathcal{Q}_{A}^{i} \, \Xi_i 
\end{equation}
followed by an isometry transformation of $\mathcal{M}_1$ parameterized by $\Lambda$. Then we have the sequence of maps between the inner layers according to the scheme discussed in previous sections, until the image of the original $\boldsymbol{\Xi}$ reaches 
$\mathcal{M}_{N}$. For binary classification, one needs to generalize the notion of hyperplane separator to the manifold $\mathcal{M}_{N}$ and needs the be able to compute the (unique) distance of a given point from this separator.
\begin{definizione}
\label{separatoni}
Let $\mathcal{M}=\mathrm{U/H}$ be any non-compact symmetric space. A separator of $\mathcal{M}$ is a \textit{codimension 
one} submanifold $\mathfrak{S} \subset \mathcal{M}$, with the following properties:
\begin{description}
  \item[a)] $\mathfrak{S}$ is a homogeneous space.
  \item[b)] $\mathfrak{S}$ intersects the boundary at infinity of $\mathcal{M}$ : $\partial \mathcal{M}\cap \mathfrak{S} \neq \emptyset$
  \item[c)] $\mathfrak{S}$ induces a binary partition of $\mathcal{M}$ defined as follows.
  There exists two submanifolds $\mathcal{M}_\pm \subset \mathcal{M}$ such that
  \begin{enumerate}
    \item $\mathcal{M} \, = \, \mathcal{M}_+\bigcup\mathcal{M}_-$
    \item $\text{dim}\,\mathcal{M}_\pm  \, = \, \text{dim}\, \mathcal{M}$ 
    \item $\mathcal{M}_+\bigcap\mathcal{M}_- \, = \, \mathfrak{S} $
  \end{enumerate}
\end{description}
\end{definizione}
The explicit construction of separators for generic symmetric manifolds $\mathcal{M}^{[r,s]}$ of the series (\ref{Mrq}) is addressed and presented in the forthcoming paper \cite{tassella} while for the case $r=1$, as we already anticipated at the beginning of section \ref{architettoTS}, we have that the separator 
\begin{equation}\label{arciseparo}
\mathcal{M}^{1,q}  \subset \mathfrak{S} \, = \,\mathrm{Adj}_g \left[\mathcal{M}^{1,q-1}\right]
\end{equation}
is the conjugate for some $g\in \mathrm{SO(1,1+q)}$ of the standard symmetric submanifold $\mathcal{M}^{1,q-1}\subset \mathcal{M}^{1,q}$  and hence not only homogeneous but even symmetric. 
\par
In any case, given a separator $\mathfrak{S}\left[\boldsymbol{\theta}\right] \subset \mathcal{M}_{N}$, labeled by its immersion parameters $\boldsymbol{\theta}$, the logistic regression immediately follows. First, we introduce the notion of oriented distance of a point from the separator.
\begin{definizione}
\label{oriadistanza} Let $\mathcal{M}$ be a non-compact symmetric space and $\mathfrak{S}\left(\boldsymbol{\theta}\right) \subset \mathcal{M}$ a parameterized separator. We define the  \textit{unoriented distance} of any point $\mathbf{p}\in \mathcal{M}$ from the separator as:
\begin{equation}\label{curtalengo}
 \delta\left(\mathbf{p},\mathfrak{S}\left[\boldsymbol{\theta}\right] \right) \, \equiv \, \underbrace{\text{\rm inf}}_{\mathbf{u} \in \mathfrak{S}\left[\boldsymbol{\theta}\right]} \, \mathrm{d}\left(\mathbf{p},\mathbf{u}\right)
\end{equation}
where $\mathrm{d}( \, , \, )$ is the distance function in $\mathcal{M}$. 
\end{definizione} 
By definition if $\mathbf{u}_0$ is the point in the separator that corresponds to the minimum distance, then $\delta\left(\mathbf{p},\mathfrak{S}\left[\boldsymbol{\theta}\right] \right)\, = \, \mathrm{d}\left(\mathbf{p},\mathbf{u}_0\right)$
and the latter is the length of the geodesic arc from $\mathbf{u}_0$ to $\mathbf{p}$. Given such a geodesic curve, unique in a non-compact symmetric space, one can consider its tangent vector $\mathfrak{t}_0$  in $\mathbf{u}_0$ and evaluate the projection of the latter onto the normal bundle to the separator in the same point. Since the separator has codimension one, we can always split the tangent bundle $T\mathcal{M} \, =\, T\mathfrak{S}\left[\boldsymbol{\theta}\right]  \oplus T\mathcal{N}$ and with respect to any standard frame of the normal bundle the geodesic tangent vector can have a positive or negative projection. This sign converts the unoriented distance $\delta\left(\mathbf{p},\mathfrak{S}\left[\boldsymbol{\theta}\right] \right) $ into the \textbf{oriented} one that we denote with a hat:
\begin{equation}\label{corretopino}
  \widehat{\delta}\left(\mathbf{p},\mathfrak{S}\left[\boldsymbol{\theta}\right] \right) \, = \, \underbrace{\pm}_{\text{if } \mathfrak{t}_0\cdot T\mathcal{N} = \pm}   \, \delta\left(\mathbf{p},\mathfrak{S}\left[\boldsymbol{\theta}\right] \right)
\end{equation}

Having introduced the oriented distance, we can replace eq.(\ref{ramarro}) with the following one:
\begin{equation}\label{sonnambulo} 
\forall \mathbf{p}\in \mathcal{M}_{N} \quad ; \quad 
\mathfrak{p}\left(Y=1\mid\mathbf{X} = \mathbf{p},\boldsymbol{\theta}\right) \, = \mathfrak{p}\left( \mathbf{p};\boldsymbol{\theta}\right) \, = \, 
 \sigma\left( 
 \widehat{\delta}\left(
 \mathbf{p},\mathfrak{S}\left[\boldsymbol{\theta} \right]\right)\right)
\end{equation}
and we might construct the \textbf{likelihood functional} as in eq.(\ref{gelsominobrutto}) to be regarded as a function of the immersion parameters $\boldsymbol{\theta} $ of the separator. We have to recall at this point that the entire chain of equation (\ref{formaggio}) from the input space $\mathbb{V}^{0}_{in}$ to $\mathcal{M}_{N}$ can be seen as a unique map (see eq.(\ref{fisarmonica}))
\begin{equation}
\tilde{\boldsymbol{\mathcal{U}}}_{[\boldsymbol{\mathfrak{W}},\boldsymbol{\Psi}]}=\boldsymbol{\mathcal{U}}_{[\boldsymbol{\mathfrak{W}},\boldsymbol{\Psi}]}\circ \iota_{[\mathcal{Q},\Lambda]} \quad : \quad \mathbb{V}^0_{in} \, \longrightarrow \, 
\mathcal{M}_{N}  
\end{equation}
where the injection parameters $\mathcal{Q}$ and $\Lambda$ have been included in the collections $\boldsymbol{\mathfrak{W}}$ and 
$\boldsymbol{\Psi}$, respectively. 
Hence, given an original datum $\Xi_i$, the points $\mathbf{p}_i$ involved in the calculation of the likelihood function are the images through the map 
$\boldsymbol{\mathcal{U}}_{[\boldsymbol{\mathfrak{W}},\boldsymbol{\Psi}]} $ so that we can write
\begin{equation}\label{caranello}
  \mathbf{p}_i \, = \, \tilde{\boldsymbol{\mathcal{U}}}_{[\boldsymbol{\mathfrak{W}},\boldsymbol{\Psi}]}\left(\Xi_i\right)
\end{equation}
Correspondingly eq.  (\ref{sonnambulo}) can be rewritten as
\begin{equation}\label{rattatuia}
  \forall\, \Xi \in \mathbb{V}^0_{in} \quad ; \quad 
\mathfrak{p}\left( \Xi;\,\underbrace{\boldsymbol{\theta}, \boldsymbol{\mathfrak{W}},\boldsymbol{\Psi} }_{\boldsymbol{\zeta}}\right) \, = \, 
 \sigma\left( 
 \widehat{\delta}\left(
\tilde{\boldsymbol{\mathcal{U}}}_{[\boldsymbol{\mathfrak{W}},\boldsymbol{\Psi}]}\left(\Xi\right) ,\mathfrak{S}\left[\boldsymbol{\theta} \right]\right)\right)
\end{equation}
where we have collected all the parameters as  $\boldsymbol{\zeta}$. 
\par 
Using the probability function
$$ \mathfrak{p}\left( \Xi;\boldsymbol{\zeta}\right) $$
we can now construct the likelihood functional as in eq.(\ref{gelsominobrutto}).
\begin{equation}\label{gelsominobello}
  \mathcal{L}(\boldsymbol{\zeta}) \, \equiv \,  \prod_{(y,\,\mathbf{\Xi})\in \mathcal{C}_1} \mathfrak{p}\left(\Xi;\boldsymbol{\zeta}\right)\, \times \, \prod_{(y,\,\boldsymbol{\Xi})\in \mathcal{C}_2} \left(1\, - \,\mathfrak{p}\left(\Xi;\boldsymbol{\zeta}\right) \right)
\end{equation}
In the case of the binary classification task, training corresponds to maximizing the likelihood functional \ref{gelsominobello}
with respect to all parameters $\boldsymbol{\zeta}$.
\paragraph{An alternative for \texorpdfstring{$r=1$}{r=1}.} In the case of $r=1$, we recall equations (16), (17), and (48) of \cite{naviga} to place the actual procedures of the twin experimental paper \cite{naviga} into the general mathematical perspective of the present discussion. As it appears from
the comparison with such formulae in particular with eq.(16) one sees that in \cite{naviga}, as argument of the sigmoid, we have taken 
not the oriented distance $\hat{\delta}$, rather a monotonic function of the same, namely the quantity $h_{\boldsymbol{\theta}}(\boldsymbol{p})$ whose vanishing corresponds to an algebraic equation for the separator hypersurface. The relationship between the two is given by:
\begin{equation}
\label{cosmicomica}
d(\boldsymbol{p}, \mathfrak{S}\left[\boldsymbol{\theta} \right]) = \mathrm{arcsinh}\left(|\tilde h_{\boldsymbol{\theta}}(\boldsymbol{p})|\right)
\end{equation}
where $\tilde h_{\boldsymbol{\theta}}(\boldsymbol{p})=h_{\boldsymbol{\theta}}(\boldsymbol{p})/\sqrt{\|\boldsymbol{w}\|^2-\alpha \beta}$ with $\boldsymbol{w},\alpha,\beta$ being the parameters composing $\boldsymbol{\theta}$ described in details in \cite{naviga}, and:

\begin{equation}\label{citrullo}
 h_{\boldsymbol{\alpha,\beta,\boldsymbol{w}}}(\boldsymbol{p}) \, = \alpha\, e^{-\Upsilon_1} + \langle\boldsymbol{w} ,\,\boldsymbol{\Upsilon}_2\rangle + \beta\, e^{\Upsilon_1} \left(1+|\boldsymbol{\Upsilon}_2|^2\right)
\end{equation}

with with $|\boldsymbol{w}|^2 - 4\alpha\beta > 0$, $\alpha,\,\beta \in \mathbb{R}$, $\boldsymbol{w}\in \mathbb{R}^q$.
Hence, using $\tilde h$ as an argument of the sigmoid corresponds in terms of the distance to the use of another sigmoid-like function:
\begin{equation}
  \tilde{\sigma}(x) \equiv \sigma\left(\sinh(x)\right) 
\end{equation}
that, thanks to $\sinh(-x) \, = \, -\sinh(x)$, has exactly the same properties as those of $\sigma(x)$ displayed in eq.(\ref{sigma_properties}) and equally maps the whole real line $\mathbb{R}$ into the interval $[0,1]$.
\par
The use of $\tilde{\sigma}$ instead of $\sigma$ is favorable when $r=1$ since, formulating the logistic regression in this way, the 
argument of the true sigmoid $\sigma$ looks like its argument in the Euclidean case, namely the equation of a hyperplane. However, it is important 
to stress that the notion of the oriented distance is more general, while the algebraic equation of the separator changes in different spaces. As the reader will see in \cite{tassella} also in the general $r$ case the formulation of the logistic regression can be done in terms of the function $\tilde{\sigma}$ since a formula completely analogous to eq.(\ref{cosmicomica}) written in terms of a single solvable coordinate can be derived also for $r\geq 2$. 
\subsection{Generalization to the case of several classes and softmax}
What we described above applies to binary classification. This method can be easily generalized when the number of classes is larger than two. To 
do so, we introduces a set of $K$ separators $\boldsymbol{\mathfrak{S}}[\boldsymbol{\theta}_k]$, $k=1,\dots, K$ and we consider the 
\textit{softmax}: 
\begin{equation}\label{colloni}
  \boldsymbol{p} \left(Y=k \, \mid\mathbf{X} =  \mathbf{p},\boldsymbol{\theta}_1,\dots,\boldsymbol{\theta}_K, \right) \, \equiv \,
  \frac{\exp\left[\widehat{\delta}\left(
 \mathbf{p},\mathfrak{S}\left[\boldsymbol{\theta}_k  \right]\right)\right]}{\sum_{i}^K\exp\left[\widehat{\delta}\left(
 \mathbf{p},\mathfrak{S}\left[\boldsymbol{\theta}_i  \right]\right)\right]}\quad ; \quad k=1,\dots, K
  \end{equation}
  expressing the probability that the point $\mathbf{p}\in \mathcal{M}_{N} $ is in the $k$-th chamber of the partition 
  defined by the $K$ partition of the full space in $K$ hulls that we identify with the classes. The formulation of the resulting likelihood is analogous to the previous binary case. The simulations conducted in the twin paper \cite{naviga} have utilized $K=10$ and the softmax model. We refer to it for all details concerning the numerical results and the evaluation of their robustness. 

\section{Conclusions and Perspectives}
This paper elucidates the mathematical mechanisms underpinning the construction of Cartan Neural Networks, a new class of learning algorithms introduced in the twin paper \cite{naviga}, which was aimed at the Machine Learning Community. Together, these two papers constitute the first step in the development of the long-term PGTS project  laid down in \cite{pgtstheory}, whose mission 
is that of reformulating neural networks in a covariant setup, where the non-linearity 
necessary for expressivity is provided by a sequence of geometric and group-theoretic operations. The key mathematical ingredient is a general scheme whereby all non-compact symmetric spaces are \textbf{metrically equivalent}  to as many \textbf{solvable Lie 
subgrops} $\mathcal{S} \subset \mathrm{U}$ precisely determined for each $\mathrm{U/H}$ 
\cite{Alekseevsky1975,Alekseevsky:2003vw,tittusnostro,pancetta1,pancetta2,Trigiante:1998vu,pgtstheory}.
As we have already emphasized in this paper, the choice of non-compact symmetric spaces logically follows from the recognition that the existence of a uniquely defined \textbf{distance function}, which reduces the layers to be \textbf{Cartan-Hadamard manifolds}, is vital for machine learning algorithms. Furthermore, the requirement that only relative distances should matter imposes homogeneity, and this reduces the choice to non-compact coset manifolds with non-positive curvature. The use of Euclidean spaces corresponds to the limiting case of null-curvature, and this is what kills spontaneous non-linearity, thus requiring the use of point-wise activation functions. Restoring negative curvature is what opens the perspective of disposing of point-wise activation functions and introducing covariance and geometrical interpretability.
\par
In the twin paper, the explicit examples that have been constructed correspond to the non-compact rank $r=1$ case, namely to the hyperbolic spaces
$\mathbb{H}^n$. While the results reported were already competitive, we argue here that they might be significantly improved by raising the non-compact 
rank and even consider cases with variable rank. 
\par
In particular, we consider it very important to make a general study of the nested homomorphisms between the solvable Lie subalgebras of $\slal(N,\mathbb{R})$ at various values of $N$. The examples of homomorphisms included  in the present paper 
are to be regarded just as motivational toy models in preparation for a more extensive and systematic study to be performed elsewhere. 
We expect to find a quite higher degree of expressivity by using $\mathrm{SL(N,\mathbb{R})/SO(N)}$ manifolds as a modeling of 
interior layers. 

The Cartan Networks framework allows a reappraisal of the geometrical formulation of the classification algorithms popular in data science under the names of \textit{logistic regression}, \textit{multinomial logistic regression}, and \textit{softmax}. It appears that geometrically, the classification problem can be viewed as a chamber partition of manifolds that are symmetric spaces. It is clear that a more in-depth mathematical study of such chamber structures is needed, and a first set of results will be presented in the forthcoming paper
\cite{tassella}. 

Finally, an important future direction will be to leverage the geometric interpretability of Cartan Networks to obtain insights into the specific patterns recognised by the networks. \textit{A posteriori}, it should be possible to extract the matrices $W_{homo}$ learned by the network and put them in a normal canonical form, which should reveal the characteristics of the learned map from one layer to the next one. We hypothesize that this will constitute an instrument of study of great value to understand the mathematical essence of the learned patterns, leading both to optimization of architectures and opening roads to grasp the \textbf{hidden laws} of the processes studied.
\subsection{Perspectives for further steps} As we stated above, the present pair of twin papers is to be regarded, together with the immediately forthcoming one \cite{tassella}, as the first step in the development of the PGTS programme. The next steps in such development are the following:
\begin{description}
    \item[1)] Relying on a new explicit formulation for separators (to appear in \cite{tassella}) to extend Cartan networks, introduced in \cite{naviga} for $r=1$, to manifolds with $r\geq 2$.
    \item[2)] Formulating convolutional-like Cartan Neural Networks based on the modeling of data structure as Tits Satake vector bundles (see \cite{tassella}).
    \item[3)] Studying the class of functions that a shallow \textit{Cartan Neural Network} can approximate with a $\mathrm{SL(N,\mathbb{R})/SO(N)}$ hidden layer. Here, a particularly relevant role is going to be played by the harmonic analysis, as described in the foundational paper \cite{pgtstheory} and as rediscussed in \cite{tassella}.
    \item[4)] Last but not least, inspecting the possible relation of the ML techniques of Principal Component Analysis with the Tits Satake Paint Group Structure of the hidden layers. Of course, this introduces the challenge of \textit{unsupervised learning}, and relates to the \textit{a priori} optimization of an injection map that should match the statistical relevance of the various features with the \textbf{intrinsic hierarchical grading} of the solvable coordinates.
\end{description}

\section*{Acknowledgements}
P. G. Fr\'e would like to acknowledge very much inspiring discussions with his close friend and habitual coauthor Mario Trigiante. Furthermore, we are all grateful to our friend Sauro Additati,  main promoter and cofinancer of the long-term project of which this and its twin paper constitute, 
after the first \cite{pgtstheory}, the second step. It is always a pleasure to share with Sauro the updates on our advances and discuss the 
perspectives. The authors thank Ugo Bruzzo for the clarifications and the reformulation of Lemma \ref{omomorfista}.
\newpage 
\appendix
\section{Homomorphisms in the \texorpdfstring{$r=1$}{r=1} case}\label{app:maurercartan}
We develop here the computations that lead to Equation \ref{eq:homo}. The pullback is defined by a linear action on the basis $e_i$ and $E_i$. We write $$h^*(e^i) = W^i_j E^j$$
Therefore, the MC equations are
\begin{equation}\label{eq:mc1}
    \mathrm{d}(W_k^1 E^k) = 0
\end{equation}
\begin{equation}\label{eq:mc2}
\mathrm{d}(W^{1+j}_k E^k) = W^1_l W^{1+j}_m E^l \wedge E^m\end{equation}
From equation \ref{eq:mc1} we derive 
\begin{equation}
W_k^1 E^1 \wedge E^k = 0
\end{equation}
hence $W_{k+1}^1 = 0$.
Equation \ref{eq:mc2} is rewritten as 
\begin{equation}W_{1+k}^{1+j}E^1 \wedge E^{1+k} = W^1_l W^{1+j}_m E^l \wedge E^m = W^1_{1} W^{1+j}_m E^{1} \wedge E^m \end{equation}
from which $W_1^1 = 1$, and the $W_m^{1+j}$ are free. Therefore the action of $h^*$ on $e^i$ can be written as
\begin{equation}h^*\begin{pmatrix} e^1 \\ \boldsymbol{e}^2\end{pmatrix} = \begin{pmatrix} 1 & \boldsymbol{0} \\ \boldsymbol{W}_1^2 &\boldsymbol{W} \end{pmatrix} \begin{pmatrix} E^1 \\ \boldsymbol E^2\end{pmatrix}
\end{equation}
We now write the 1-forms in coordinates, assuming solvable parametrization $Y_i$ for the forms $E^i$ and $X_i$ for the forms $e^i$. We have
\begin{equation}
E^1 = \mathrm{d}Y_1
\end{equation}
\begin{equation}E^{1+j} = \frac{1}{\sqrt{2}} Y_{1+j} \mathrm{d}Y_1 + \frac{1}{\sqrt{2}}\mathrm{d}Y_{1+j}
\end{equation}
and
\begin{equation}
e^1 = \mathrm{d}X_1
\end{equation}
\begin{equation}
e^{1+j} = \frac{1}{\sqrt{2}} X_{1+j} \mathrm{d}X_1 + \frac{1}{\sqrt{2}}\mathrm{d}X_{1+j}
\end{equation}
Then the above condition becomes
\begin{equation}
h^*(e^1) = E^1 = \mathrm{d}Y_1
\end{equation}
\begin{equation}
h^*(e^1) = f^*(\mathrm{d}X_1) = \frac{\partial f_1}{\partial Y_j}\mathrm{d}Y_j
\end{equation}
hence 
\begin{equation}
\frac{\partial f_1}{\partial Y_j} = \delta_{j1}
\end{equation}
so $X_1 = f_1(Y) = Y_1$.
For the other coordinates, we have 
\begin{equation}h^*(e^{j+1}) = W_i^{j+1} E^i = (W_1^{j+1} + \frac{1}{\sqrt{2}} W_{1+i}^{1+j} Y_{1+i})\mathrm{d}Y_1 + \frac{1}{\sqrt{2}} W_{1+i}^{1+j} \mathrm{d}Y_{1+i}
\end{equation}
but also 
\begin{equation}
h^*(e^{j+1}) = h^* (\frac{1}{\sqrt{2}} X_{1+j} \mathrm{d}X_1 + \frac{1}{\sqrt{2}}\mathrm{d}X_{1+j}) = \frac{1}{\sqrt{2}}\left(f_{1+j}(Y)\frac{\partial {f_1}}{\partial Y_k} + \frac{\partial f_{1+j}}{\partial Y_k}\right) \mathrm{d}Y_k
\end{equation}
Hence we are left with the system:
\begin{equation}
\frac{\partial h_{1+j}}{\partial Y_1} = (\sqrt{2} W_1^{j+1} + W_{1+i}^{1+j} Y_{1+i}) - h_{1+j}(Y)
\end{equation}
\begin{equation}
\frac{\partial h_{1+j}}{\partial Y_{1+i}} = W_{1+i}^{1+j} \quad i = 0, \dots, n
\end{equation}
so
\begin{equation}
X_{1+j} = h_{1+j}(Y) =  Ce^{-Y_1} +(\sqrt{2} W_1^{j+1} + W_{1+i}^{1+j} Y_{1+i})
\end{equation}
which is in agreement with Theorem 3.1 of \cite{naviga}.
\section{Examples of non-trivial solvable Lie algebra homomorphisms with TS class-switching}
\label{patagonia} To illustrate the ample spectrum of possibilities that opens up for network architectures when we employ the general scheme of solvable Lie algebra homomorphisms, without restriction to a fixed non-compact rank 
$r$ (namely when, instead of navigating just one Tits-Satake universality class, we allow class-switching) we can consider a toy model given by the deformations of the canonical triangular embedding of $\mathbb{H}^3$ into  $\mathrm{SL(4,\mathbb{R})/SO(4)}$. We will denote by $Solv^{\mathfrak{t}}_{[d|n]}$ the algebra of class $\mathfrak{t}$, with $d$ Cartan generators and of dimension $n$.
\par
Focusing  our attention on the two solvable Lie algebras $Solv^{\mathfrak{a}}_{[3|9]} \subset \slal(4,\mathbb{R})$ 
and $Solv^{\mathfrak{d}}_{[1|3]} \subset \so(1,3)$, we will consider the available injection and projection maps:
\begin{equation}\label{cordaro}
  \mathcal{I}_{\mathcal{W}} \quad : \quad Solv^{\mathfrak{d}}_{[1|3]} \, \stackrel{\iota}{\hookrightarrow} \, Solv^{\mathfrak{a}}_{[3|9]}
  \quad ; \quad \mathcal{P}_{\mathcal{W}} \quad : \quad Solv^{\mathfrak{a}}_{[3|9]} \stackrel{\pi}{\rightarrowtail} Solv^{\mathfrak{d}}_{[1|3]} 
\end{equation}
and discuss in a couple of cases the corresponding $\boldsymbol{\Phi}[\mathcal{W} \mid \mathbf{X}\cdot\mathrm{t}]$ maps
between solvable coordinates. 
\paragraph{The larger solvable algebra}
The solvable Lie subalgebra $Solv^{\mathfrak{a}}_{[3|9]} \subset \slal(4,\mathbb{R})$ that corresponds to the symmetric
space $\mathrm{SL(4,\mathbb{R})/SO(4)}$ is abstractly and uniquely defined by the following Maurer-Cartan equations:
\begin{equation}\label{birillo}
\begin{array}{rcl}
 dE^{1} & \text{ = } & 0 \\
 dE^{2} & \text{ = } & 0 \\
 dE^{3} & \text{ = } & 0 \\
 dE^{4}-2 E^{1}\wedge E^{4}-E^{2}\wedge E^{4}-E^{3}\wedge E^{4} & \text{ = } & 0 \\
 dE^{5}+E^{1}\wedge E^{5}-E^{2}\wedge E^{5} & \text{ = } & 0 \\
 dE^{6}+E^{2}\wedge E^{6}-E^{3}\wedge E^{6} & \text{ = } & 0 \\
 dE^{7}-E^{1}\wedge E^{7}-2 E^{2}\wedge E^{7}-E^{3}\wedge E^{7}+E^{4}\wedge E^{5} & \text{
   = } & 0 \\
 dE^{8}+E^{1}\wedge E^{8}-E^{3}\wedge E^{8}+E^{5}\wedge E^{6} & \text{ = } & 0 \\
 dE^{9}-E^{1}\wedge E^{9}-E^{2}\wedge E^{9}-2 E^{3}\wedge E^{9}+E^{4}\wedge
   E^{8}-E^{6}\wedge E^{7} & \text{ = } & 0 \\
\end{array}
\end{equation}
The solubility of the algebra is manifest in a very nice and systematic way from eq.(\ref{birillo}) that  also shows a general
pattern holding true for any value of $N$ (see sect.\ref{sandalo}).
\paragraph{The smaller solvable algebra}
Let us next consider the  solvable Lie subalgebra $Solv^{\mathfrak{d}}_{[1|3]} \subset \so(1,3)$ that corresponds to the symmetric space $\mathrm{SO(1,3)/SO(3) }\sim \mathbb{H}^{3}$. It is abstractly and uniquely defined by the following Maurer-Cartan equations:
\begin{equation}\label{carmelitano}
  \begin{array}{rcl}
 \text{d}e^1 & = & 0\\
 \text{d}e^2+e^1\wedge e^2 & = & 0\\
 \text{d}e^3+e^1\wedge  e^3& = & 0 \\
\end{array}
\end{equation}
\par
We are interested in discussing the general form of both the enlargement and restriction maps between these two solvable algebras, and hence both the injection maps:
\begin{equation}\label{crallus}
  \dfrac{\mathrm{SO(1,3)}}{\mathrm{SO(3)}} \, \stackrel{\iota}{\hookrightarrow} \, \dfrac{\mathrm{SL(4,\mathbb{R})}}{\mathrm{SO(4)}}
\end{equation}
and the projection maps:
\begin{equation}\label{sullarc}
  \dfrac{\mathrm{SL(4,\mathbb{R})}}{\mathrm{SO(4)}}
  \stackrel{\pi}{\rightarrowtail } \dfrac{\mathrm{SO(1,3)}}{\mathrm{SO(3)}}
\end{equation}
Before exploring the \textbf{consistency conditions to be a homomorphism} for the algebraically defined $W$ matrices, we review the solvable coordinate parameterization of the two solvable group manifolds.
\subsection{Parameterization of \texorpdfstring{$\mathcal{S}^{\mathfrak{a}}_{[3|9]}$}{Sa3|9}}
The generic form of an element of the solvable Lie algebra $Solv^{\mathfrak{a}}_{[3|9]}$, in the basis of generators that
we utilize is the following
\begin{equation}\label{ganimede}
 \boldsymbol{\Upsilon}\cdot \mathrm{T} \, = \, \left(
\begin{array}{cccc}
 -\sum_{i=1}^3 \,\Upsilon_i  & \Upsilon _4 & \Upsilon _7 & \Upsilon _9 \\
 0 & \Upsilon _1 & \Upsilon _5 & \Upsilon _8 \\
 0 & 0 & \Upsilon _2 & \Upsilon _6 \\
 0 & 0 & 0 & \Upsilon _3 \\
\end{array}
\right)
\end{equation}
which exactly corresponds, for the choice $N=4$, to the expression mentioned in eq. \ref{pittura}.
Then, according with the definition (\ref{sigmadefillo}) of the exponential map $\Sigma$, the generic element
of the solvable Lie algebra $Solv^{\mathfrak{a}}_{[3|9]}$,  displayed in eq.(\ref{ganimede}), maps to the generic
element $\mathbb{L}_{\mathfrak{a}}\left(\boldsymbol{\Upsilon}\right)$ of the solvable Lie group $\mathcal{S}^{\mathfrak{a}}_{[3|9]}$:
\begin{equation}\label{giasone}
  \Sigma \left(\boldsymbol{\Upsilon}\cdot \mathrm{T}\right) \, = \, \mathbb{L}_{\mathfrak{a}}\left(\boldsymbol{\Upsilon}\right) 
\end{equation}
where the latter  takes  the following explicit expression:
\begin{eqnarray}\label{birichino}
 &\mathbb{L}_{\mathfrak{a}}\left(\boldsymbol{\Upsilon}\right) \, =\,   &\nonumber\\
 &\left(
\begin{array}{cccc}
 e^{\frac{1}{2} \left(\Upsilon _1+\Upsilon _2+\Upsilon _3\right)} & -e^{\frac{1}{2}
   \left(\Upsilon _1+\Upsilon _2+\Upsilon _3\right)} \Upsilon _4 & e^{\frac{1}{2}
   \left(\Upsilon _1+\Upsilon _2+\Upsilon _3\right)} \left(\Upsilon _4 \Upsilon
   _5-\Upsilon _7\right) & -e^{\frac{1}{2} \left(\Upsilon _1+\Upsilon _2+\Upsilon
   _3\right)} \left(\Upsilon _4 \left(\Upsilon _5 \Upsilon _6-\Upsilon
   _8\right)+\Upsilon _9\right) \\
 0 & e^{-\frac{\Upsilon _1}{2}} & -e^{-\frac{\Upsilon _1}{2}} \Upsilon _5 &
   e^{-\frac{\Upsilon _1}{2}} \left(\Upsilon _5 \Upsilon _6-\Upsilon _8\right) \\
 0 & 0 & e^{-\frac{\Upsilon _2}{2}} & -e^{-\frac{\Upsilon _2}{2}} \Upsilon _6 \\
 0 & 0 & 0 & e^{-\frac{\Upsilon _3}{2}} \\
\end{array}
\right)&\nonumber\\
\end{eqnarray}
The reader is invited to take notice that the structure of the solvable group element is the same as that of the Lie algebra element, with effective new entries that are exponential and polynomial functions of the entries of the Lie algebra element. The number of different exponents is $N-1$ (in our case $3$), and the polynomials span the various degrees from linear to $d_{max} \, = \, N-1$ as the rank.
\par
Calculating the left invariant $1$-form:
\begin{equation}\label{lasciamistare}
  \Theta_{\mathfrak{a}} \, \equiv \, \mathbb{L}^{-1}\left(\boldsymbol{\Upsilon}\right)\cdot \mathrm{d }\mathbb{L}\left(\boldsymbol{\Upsilon}\right)
\end{equation}
and projecting it onto the basis of generators $\mathrm{T}$:
\begin{equation}\label{thetanove}
  \Theta_{\mathfrak{a}} \, = \, \sum_{i=1}^9 \, \mathcal{E}^i \, T_i
\end{equation}
we obtain the explicit form of the left invariant $1$-forms $\mathcal{E}^i$ that satisfy the abstract Maurer-Cartan eq.s (\ref{birillo}):
\begin{equation}\label{corridori}
\begin{array}{rcl}
 \mathcal{E}^1 & \text{ = } & -\frac{\text{d$\Upsilon $}_1}{2} \\
 \mathcal{E}^2 & \text{ = } & -\frac{\text{d$\Upsilon $}_2}{2} \\
 \mathcal{E}^3 & \text{ = } & -\frac{\text{d$\Upsilon $}_3}{2} \\
 \mathcal{E}^4 & \text{ = } & -\frac{1}{2} \left(2 \text{d$\Upsilon $}_1+\text{d$\Upsilon
   $}_2+\text{d$\Upsilon $}_3\right) \Upsilon _4-\text{d$\Upsilon $}_4 \\
 \mathcal{E}^5 & \text{ = } & \frac{1}{2} \left(\text{d$\Upsilon $}_1-\text{d$\Upsilon
   $}_2\right) \Upsilon _5-\text{d$\Upsilon $}_5 \\
 \mathcal{E}^6 & \text{ = } & \frac{1}{2} \left(\text{d$\Upsilon $}_2-\text{d$\Upsilon
   $}_3\right) \Upsilon _6-\text{d$\Upsilon $}_6 \\
 \mathcal{E}^7 & \text{ = } & \frac{1}{2} \left(2 \text{d$\Upsilon $}_4 \Upsilon _5+2
   \text{d$\Upsilon $}_1 \Upsilon _4 \Upsilon _5+\text{d$\Upsilon $}_2 \Upsilon _4
   \Upsilon _5+\text{d$\Upsilon $}_3 \Upsilon _4 \Upsilon _5-\text{d$\Upsilon $}_1
   \Upsilon _7-2 \text{d$\Upsilon $}_2 \Upsilon _7-\text{d$\Upsilon $}_3 \Upsilon
   _7-2 \text{d$\Upsilon $}_7\right) \\
 \mathcal{E}^8 & \text{ = } & \frac{1}{2} \left(2 \text{d$\Upsilon $}_5 \Upsilon
   _6-\text{d$\Upsilon $}_1 \Upsilon _5 \Upsilon _6+\text{d$\Upsilon $}_2 \Upsilon
   _5 \Upsilon _6+\text{d$\Upsilon $}_1 \Upsilon _8-\text{d$\Upsilon $}_3 \Upsilon
   _8-2 \text{d$\Upsilon $}_8\right) \\
 \mathcal{E}^9 & \text{ = } & \frac{1}{2} \left(-2 \text{d$\Upsilon $}_1 \Upsilon _4 \Upsilon
   _5 \Upsilon _6-\text{d$\Upsilon $}_2 \Upsilon _4 \Upsilon _5 \Upsilon
   _6-\text{d$\Upsilon $}_3 \Upsilon _4 \Upsilon _5 \Upsilon _6-2 \text{d$\Upsilon
   $}_6 \Upsilon _7+\text{d$\Upsilon $}_2 \Upsilon _6 \Upsilon _7-\text{d$\Upsilon
   $}_3 \Upsilon _6 \Upsilon _7\right.\\
   \null&\null&\left.+2 \text{d$\Upsilon $}_1 \Upsilon _4 \Upsilon
   _8+\text{d$\Upsilon $}_2 \Upsilon _4 \Upsilon _8\right.\\
   \null&\null&\left.+\text{d$\Upsilon $}_3 \Upsilon
   _4 \Upsilon _8+\text{d$\Upsilon $}_4 \left(2 \Upsilon _8-2 \Upsilon _5 \Upsilon
   _6\right)-\text{d$\Upsilon $}_1 \Upsilon _9-\text{d$\Upsilon $}_2 \Upsilon _9-2
   \text{d$\Upsilon $}_3 \Upsilon _9-2 \text{d$\Upsilon $}_9\right) \\
\end{array}
\end{equation}
\subsection{Parameterization of \texorpdfstring{$\mathcal{S}^{\mathfrak{d}}_{[1|3]}$}{Sd1|3}}
Let us now consider the smaller solvable group. The generic form of an element of the solvable Lie algebra $Solv^{\mathfrak{d}}_{[1|3]}$, in our basis of generators, is the following:
\begin{equation}\label{aspirina}
\boldsymbol{w}\cdot \mathrm{t} \, = \, \left(
\begin{array}{cccc}
 w_1 & \frac{w_2}{\sqrt{2}} &
   \frac{w_3}{\sqrt{2}} & 0 \\
 0 & 0 & 0 & -\frac{w_2}{\sqrt{2}} \\
 0 & 0 & 0 & -\frac{w_3}{\sqrt{2}} \\
 0 & 0 & 0 & -w_1 \\
\end{array}
\right)
\end{equation}
Then, according with the definition (\ref{ladonico}) of the exponential map $\Sigma$, the generic element
of the solvable Lie algebra $Solv^{\mathfrak{d}}_{[1|3]}$,  displayed in eq.(\ref{aspirina}), maps to the generic
element $\mathbb{L}_{\mathfrak{d}}\left(\boldsymbol{w}\right)$ of the solvable Lie group 
$\mathcal{S}^{\mathfrak{d}}_{[1|3]}$:
\begin{equation}\label{pinolosiberiano}
  \mathbb{L}_{\mathfrak{d}}\left(\boldsymbol{w}\right) \, = \, \left(
\begin{array}{cccc}
 e^{w_1} & \frac{e^{w_1} w_2}{\sqrt{2}} &
   \frac{e^{w_1} w_3}{\sqrt{2}} & -\frac{1}{4}
   e^{w_1} \left(w_2^2+w_3^2\right) \\
 0 & 1 & 0 & -\frac{w_2}{\sqrt{2}} \\
 0 & 0 & 1 & -\frac{w_3}{\sqrt{2}} \\
 0 & 0 & 0 & e^{-w_1} \\
\end{array}
\right)
\end{equation}
Note that the solvable group element (\ref{pinolosiberiano}) is a particular case of the generic group element (\ref{birichino}) upon the following substitution
\begin{equation}\label{triaembed}
 \boldsymbol{\Phi}\left[\mathcal{W}_{can} \mid \mathbf{w}\cdot \mathrm{t}\right]\, =\, \left\{ \begin{array}{rcl}
 \Upsilon _1& = & 0 \\
 \Upsilon _2& = & 0 \\
 \Upsilon _3& = & 2 w_1 \\
 \Upsilon _4& = & -\frac{w_2}{\sqrt{2}} \\
 \Upsilon _5& = & 0 \\
 \Upsilon _6& = & \frac{w_3}{\sqrt{2}} \\
 \Upsilon _7& = & -\frac{w_3}{\sqrt{2}} \\
 \Upsilon _8& = & \frac{w_2}{\sqrt{2}} \\
 \Upsilon _9& = & \frac{1}{4}
   \left(w_3^2-w_2^2\right) \\
\end{array}\right.
\end{equation}
Equation (\ref{triaembed}) shows that the solvable Lie group $\mathcal{S}^{\mathfrak{d}}_{[1|3]}\subset \mathcal{S}^{\mathfrak{a}}_{[3|9]}$ is a subgroup of the solvable Lie group metrically equivalent to the symmetric space
$\mathrm{\mathrm{SL(4,\mathbb{R})}/\mathrm{SO(4)}}$ and indeed it corresponds to an injection map (\ref{crallus}) actually to the canonical one advocated by statement \ref{statamento}. It is our interest to derive the canonical embedding as a particular case of all the possible enlargements homomorphisms:
\begin{equation}\label{allargato}
  \mathcal{W}_{enl} \quad ; \quad Solv_{[1,3]}^{\mathfrak{d}} \, \hookrightarrow \,Solv^{\mathfrak{a}}_{[3|9]}
\end{equation}
To this effect, we proceed as we did in the previous case. Calculating the left invariant $1$-form:
\begin{equation}\label{nosinistra}
  \Theta_{\mathfrak{d}} \, \equiv \, \mathbb{L}^{-1}_{\mathfrak{d}}\left(\boldsymbol{w}\right)\cdot \mathrm{d }\mathbb{L}_{\mathfrak{d}}\left(\boldsymbol{w}\right)
\end{equation}
and projecting it onto the basis of generators $\mathrm{t}$:
\begin{equation}\label{theta3}
  \Theta_{\mathfrak{d}} \, = \, \sum_{\alpha=1}^3 \, {\varepsilon}^\alpha \, t_\alpha
\end{equation}
we obtain the explicit form of the left invariant $1$-forms $\mathcal{\varepsilon}^i$ that satisfy 
the abstract Maurer-Cartan eq.s (\ref{carmelitano}):
\begin{equation}\label{boscaglia}
  \begin{array}{rcl}
     \varepsilon^1  & = & dw_1 \\
     \varepsilon^2  & = & dw_2 + w_2 \, dw_1 \\
     \varepsilon^3  & = & dw_3 + w_3 \, dw_1 \\
   \end{array}
\end{equation}
Let us stress before proceeding that equations (\ref{corridori}) and (\ref{boscaglia}) have been derived from the 
respective left-invariant one forms obtained by the exponentiation map $\Sigma$  appropriately defined
for each algebra, which is not a pure matrix exponential of the solvable Lie algebra element written in matrix form. The 
immersion map provided in eq.(\ref{triaembed}) is not a linear map of the solvable coordinates $\boldsymbol{w}$ into
the solvable coordinates $\boldsymbol{\Upsilon}$. In this case, the $\boldsymbol{\Upsilon}$ are linear or quadratic functions
of the $\boldsymbol{w}$. However, if we insert eq.(\ref{triaembed}) into eq.(\ref{corridori}) we obtain a linear relation between the one-forms $\mathcal{E}^i$ and the one-forms $\varepsilon^\alpha$, precisely the following one:
\begin{equation}\label{carpazi}
  \mathcal{E}^i \, = \, \mathcal{W}^i_\alpha \, \varepsilon^\alpha
\end{equation}
where the $9\times 3$ matrix $\mathcal{W}^i_\alpha$ is the following one:
\begin{equation}\label{perdelio}
  \mathcal{W}_{can} \, = \, \left(
\begin{array}{c|c|c}
 0 & 0 & 0 \\
 \hline
 0 & 0 & 0 \\\hline
 0 & 0 & -1 \\
 \hline\hline
 0 & \frac{1}{\sqrt{2}} & 0 \\
 \hline
 0 & 0 & 0 \\
 \hline
 0 & 0 & -\frac{1}{\sqrt{2}} \\
 \hline
 0 & 0 & \frac{1}{\sqrt{2}} \\
 \hline
 0 & -\frac{1}{\sqrt{2}} & 0 \\
 \hline
 0 & 0 & 0 \\
\end{array}
\right)
\end{equation}
This illustrates the general principle discussed in section \ref{Liegiocaflipper} that any solvable group homomorphism, independently from the basis of solvable coordinates that is utilized and hence from the employed form of the exponential map is defined by a linear homomorphism of the involved solvable Lie algebras described by a matrix $\mathcal{W}$ that relates the left-invariant $1$-forms of the target group to those of the first group.
The corresponding relation between the solvable coordinates, that is not necessarily linear, can be worked out, iteratively and unambiguously, from the linear relation among the two sets of 1-forms. As we stated in section \ref{Liegiocaflipper}, the solvable structure allows us to solve the system of equations one after the other with recursive algorithms. 
\subsection{The general form of the injection map}
In order to study all the possible enlargement maps alternative to the canonical one, we introduce a  $9\times 3$ $W$-matrix containing 21 parameters. The first three lines have the structure shown below because in force of the Maurer-Cartan equations
(\ref{birillo}) the first three $1$-form generators $E^{1,2,3}$ can only be proportional to the unique generator $e^1$ of the other algebra that is simply closed:
\begin{equation}\label{lulu}
\mathcal{W}gen \,= \,  \left(
\begin{array}{ccc}
 \alpha  & 0 & 0 \\
 \beta  & 0 & 0 \\
 \gamma  & 0 & 0 \\
 a_1 & a_2 & a_3 \\
 a_4 & a_5 & a_6 \\
 a_7 & a_8 & a_9 \\
 a_{10} & a_{11} & a_{12} \\
 a_{13} & a_{14} & a_{15} \\
 a_{16} & a_{17} & a_{18} \\
\end{array}
\right)
\end{equation}
In this way, we obtain an ansatz for the $1$-forms $E^i$ in terms of the one forms $e^\alpha$ that is the following:
\begin{equation}\label{ansazzo}
  E^i \, = \, \mathcal{W}gen^{i}_{\alpha} \, e^\alpha
\end{equation}
Next we insert equation (\ref{ansazzo}) into the Maurer-Cartan equations (\ref{birillo}) and we use the Maurer Carta equations
(\ref{carmelitano}). Imposing that (\ref{birillo}) should be zero, we obtain the set of 15 quadratic equations displayed below:
\begin{equation}\label{equatoni}
\begin{array}{rcl}
 a_2 a_6-a_3 a_5 & \text{ = } & 0 \\
 a_9 a_{11}-a_8 a_{12}-a_3 a_{14}+a_2 a_{15} &
   \text{ = } & 0 \\
 a_5 a_9-a_6 a_8 & \text{ = } & 0 \\
 a_5 (\alpha -\beta -1) & \text{ = } & 0 \\
 a_6 (\alpha -\beta -1) & \text{ = } & 0 \\
 a_8 (\beta -\gamma -1) & \text{ = } & 0 \\
 a_9 (\beta -\gamma -1) & \text{ = } & 0 \\
 a_2 (-(2 \alpha +\beta +\gamma +1)) & \text{ =
   } & 0 \\
 a_3 (-(2 \alpha +\beta +\gamma +1)) & \text{ =
   } & 0 \\
 -\alpha  a_{11}-2 a_{11} \beta -a_{11} \gamma
   -a_2 a_4+a_1 a_5-a_{11} & \text{ = } & 0 \\
 -\alpha  a_{12}-2 a_{12} \beta -a_{12} \gamma
   -a_3 a_4+a_1 a_6-a_{12} & \text{ = } & 0 \\
 \alpha  a_{14}-a_{14} \gamma -a_5 a_7+a_4
   a_8-a_{14} & \text{ = } & 0 \\
 \alpha  a_{15}-a_{15} \gamma -a_6 a_7+a_4
   a_9-a_{15} & \text{ = } & 0 \\
 -\alpha  a_{17}-a_{17} \beta -2 a_{17} \gamma
   +a_8 a_{10}-a_7 a_{11}-a_2 a_{13}+a_1
   a_{14}-a_{17} & \text{ = } & 0 \\
 -\alpha  a_{18}-a_{18} \beta -2 a_{18} \gamma
   +a_9 a_{10}-a_7 a_{12}-a_3 a_{13}+a_1
   a_{15}-a_{18} & \text{ = } & 0 \\
\end{array}
\end{equation}
Studying the system (\ref{equatoni}), we have realized that there are two main branches of the solution system, each of which has several sub-branches. 
\paragraph{The 11-parameter embedding.}
The first branch keeps the possibility of having three non-trivial Cartan forms (each proportional to the unique $e^1$), but this requires the condition:
\begin{equation}\label{calindro}
  a_2 \, =\, a_3 \, = \, a_5 \, = \, a_6\, =  \, a_8\, = \, a_9 \, = \, 0
\end{equation}
With such a condition, for the remaining parameters we obtain $19$ different solutions (branches) with a respective number of free parameters  as follows:
\begin{equation}\label{parlunghi}
 \{11,10,10,10,10,10,10,10,9,9,9,9,9,9,9,9,9,9,9\}
\end{equation}
Renaming the free parameters as $\delta_{1,\dots,11}$, we display only the principal sub-branch with the largest number of parameters, omitting the exceptional sub-branches with fewer parameters. The corresponding $W$ matrix 
that we name $\mathcal{W}_{11} $ is the following one:
\begin{equation}\label{Wtype1}
 \mathcal{W}_{11} \, = \,  \left(
\begin{array}{ccc}
 \delta _{11} & 0 & 0 \\
 -\delta _{11} & 0 & 0 \\
 \delta _{11}-1 & 0 & 0 \\
 \delta _1 & 0 & 0 \\
 \delta _9 & 0 & 0 \\
 \delta _{10} & 0 & 0 \\
 \delta _2 & \delta _3 & \delta _4 \\
 \delta _5 & \frac{2 \delta _{11} \delta _7-\delta _7+\delta _3 \delta _{10}}{\delta
   _1} & \frac{2 \delta _{11} \delta _8-\delta _8+\delta _4 \delta _{10}}{\delta _1}
   \\
 \delta _6 & \delta _7 & \delta _8 \\
\end{array}
\right)
\end{equation}
Comparing eq.(\ref{Wtype1}) with eq.(\ref{perdelio}), we see that the canonical embedding is not contained in the $11$-parameter
embedding, which therefore is a genuinely different homomorphism. Shortly, we will show the explicit form of such a homomorphism at the level of solvable coordinate mapping.
\paragraph{The 12 parameter embedding.} The second main branch of solutions of the quadric system (\ref{equatoni}) is obtained by fixing a priori the identification of one of the three Cartan $1$-form generators of the $Solv^{\mathfrak{a}}_{[3|9]}$
algebra with the unique Cartan generator of $\mathcal{S}^{\mathfrak{d}}_{[1|3]}$. Explicitly, our conventional choice is the following:
\begin{equation}\label{cynar}
  \alpha \, = \, \beta \, = \, 0 \quad \gamma \, = \, -1
\end{equation} 
Inserting the constraint (\ref{cynar}) into (\ref{equatoni}), we obtain a quadric system for the remaining $18$ parameters, of which we can find the general solution. It consists of $5$ sub-branches with  respective numbers of free parameters  as follows:
\begin{equation}\label{parlunghi1}
 \{12,11,10,10,9\}
\end{equation} 
As we did for the previous main branch, we display only the principal sub-branch with the maximal number of parameters
namely $12$:
\begin{equation}\label{principal12}
 \mathcal{W}_{12} \, = \,  \left(
\begin{array}{ccc}
 0 & 0 & 0 \\
 0 & 0 & 0 \\
 -1 & 0 & 0 \\
 \delta _1 & \delta _8 & \delta _9 \\
 0 & 0 & 0 \\
 \delta _{10} & \delta _{11} & \delta _{12} \\
 \delta _2 & \delta _3 & \delta _4 \\
 \delta _5 & \delta _6 & \frac{\delta _6 \delta
   _9+\delta _4 \delta _{11}-\delta _3 \delta
   _{12}}{\delta _8} \\
 \delta _7 & -\delta _1 \delta _6+\delta _5
   \delta _8+\delta _3 \delta _{10}-\delta _2
   \delta _{11} & \frac{-\delta _1 \delta _6
   \delta _9+\delta _5 \delta _8 \delta
   _9+\delta _4 \delta _8 \delta _{10}-\delta
   _1 \delta _4 \delta _{11}+\delta _1 \delta
   _3 \delta _{12}-\delta _2 \delta _8 \delta
   _{12}}{\delta _8} \\
\end{array}
\right)
\end{equation}
In this branch, we find the canonical embedding (\ref{perdelio}). Indeed, it suffices to choose the $12$ parameters as follows:
\begin{equation}\label{cartongesso}
  \delta _8\to \frac{1}{\sqrt{2}},\delta
   _1\to 0,\delta _9\to 0,\delta _{12}\to
   -\frac{1}{\sqrt{2}},\delta _{10}\to 0,\delta
   _{11}\to 0,\delta _2\to 0,\delta _3\to
   0,\delta _4\to \frac{1}{\sqrt{2}},\delta
   _5\to 0,\delta _6\to
   -\frac{1}{\sqrt{2}},\delta _7\to 0
\end{equation}
\subsubsection{The general functional form of the \texorpdfstring{$11$}{11}-embedding}
To appreciate the richness of the variegated possibilities that we mentioned above, we present  the general solution of
the first order differential equations (\ref{pergolato}) in the case of the $11$-embedding (\ref{Wtype1}):
\begin{equation}\label{bicoccola}
  \mathcal{E}^i(\boldsymbol{\Upsilon}) \, = \, \mathcal{W}_{11\mid \alpha}^{i} \, \varepsilon^\alpha (\boldsymbol{w})
\end{equation}
As is to be expected, the general solution of a first-order differential system introduces new free parameters that are integration constants. Disregarding the trivial constants simply corresponding to constant shifts of the solvable coordinates, three are the new relevant parameters so that the general solution contains a total of $14$-parameters that, in a hypothetical neural network involving, as layers, the two considered manifolds, would be all parameters to be fixed by learning.
\par Explicitly, we find:
\begin{equation}\label{barbabietola}
  \boldsymbol{\Phi}\left[\mathcal{W} \mid \boldsymbol{w}\cdot \mathrm{t}\right]\, = \, \left\{\begin{array}{rcl}
 \Upsilon _1& = & -2 \delta _{11} w_1 \\
 \Upsilon _2& = & 2 \delta _{11} w_1 \\
 \Upsilon _3& = & -2 \left(\delta _{11}-1\right)
   w_1 \\
 \Upsilon _4& = & \frac{\delta _1}{2 \delta
   _{11}-1}+\delta _{12} e^{\left(2 \delta
   _{11}-1\right) w_1} \\
 \Upsilon _5& = & \delta _{13} e^{-2 \delta _{11}
   w_1}-\frac{\delta _9}{2 \delta _{11}} \\
 \Upsilon _6& = & \frac{\delta _{10}}{2 \delta
   _{11}-1}+\delta _{14} e^{\left(2 \delta
   _{11}-1\right) w_1} \\
 \Upsilon _7& = & -\delta _2+\frac{\delta _1
   \delta _9}{2 \delta _{11}}-\delta _3
   w_2-\delta _4 w_3+\frac{\delta _1 \delta
   _{13} e^{-2 \delta _{11} w_1}}{2 \delta
   _{11}-1} \\
 \Upsilon _8& = & \frac{-2 \delta _{11} \delta
   _5+\delta _5-\delta _9 \delta _{10}}{2
   \delta _{11}-1}+\frac{\left(-2 \delta _{11}
   \delta _7+\delta _7-\delta _3 \delta
   _{10}\right) w_2}{\delta _1}+\frac{\left(-2
   \delta _{11} \delta _8+\delta _8-\delta _4
   \delta _{10}\right) w_3}{\delta
   _1}-\frac{\delta _9 \delta _{14} e^{\left(2
   \delta _{11}-1\right) w_1}}{2 \delta _{11}}
   \\
 \Upsilon _9& = & \frac{-\delta _1 \delta _9
   \delta _{10}+\delta _2 \delta _{11} \delta
   _{10}-\delta _1 \delta _5 \delta
   _{11}+\delta _6 \delta _{11}}{2 \left(\delta
   _{11}-1\right) \delta _{11}}+\delta _7
   \left(-w_2\right)-\delta _8 w_3-\frac{\delta
   _1 \delta _{10} \delta _{13} e^{-2 \delta
   _{11} w_1}}{\left(1-2 \delta
   _{11}\right){}^2}+\frac{\delta _1 \delta
   _{13} \delta _{14} e^{-w_1}}{1-2 \delta
   _{11}} \\
\end{array}\right.
\end{equation}
Let us comment on the above result. The original parameters contained in the $W$-matrix are the $\delta_{1,\dots,11}$ while the additional parameters 
introduced by the integration are $\delta_{12,13,14}$. If we put these latter equal to zero, we see that the dependence of the 
$\boldsymbol{\Upsilon}$ on the $\boldsymbol{w}$ becomes linear. The integration, however, provides the possibility of introducing exponentials of 
$w_1$ also in the $\boldsymbol{\Upsilon}$ coordinates corresponding to nilpotent generators. This very small and simple example reveals the 
general fact: in the class-switching case, Cartan fields of the initial manifold can enter into non-Cartan fields of the target manifold in the 
case of the enlargement injection map, although the reverse is not possible. In the restriction map, as we are going to see, the reverse is instead 
the case. Hence, combining enlargement and restriction maps, features originally assigned to Cartans can migrate to nilpotents and vice versa. 
Focusing on the above example, imagine that we were in much higher dimensions. The analogue of equation (\ref{barbabietola}) would introduce the 
available exponentials of Cartan fields into the nilpotent coordinates that, after the exponential map, produce polynomials up to the $N-1$-th degree. 

\subsection{The general form of the restriction map}
 In this case, the general $W$-matrix we have to start from is $3\times 9,$ and we can parameterize it with 27 parameters as follows:
 \begin{equation}\label{MrWu}
\mathcal{W}proj \, = \,   \left(
\begin{array}{ccccccccc}
  {a_1} &  {a_2} &  {a_3} &  {a_4}
   &  {a_5} &  {a_6} &  {a_7} &
    {a_8} &  {a_9} \\
  {b_1} &  {b_2} &  {b_3} &  {b_4}
   &  {b_5} &  {b_6} &  {b_7} &
    {b_8} &  {b_9} \\
  {c_1} &  {c_2} &  {c_3} &  {c_4}
   &  {c_5} &  {c_6} &  {c_7} &
    {c_8} &  {c_9} \\
\end{array}
\right)
 \end{equation}
 Inserting the ansatz:
 \begin{equation}\label{projansaz}
 e^\alpha \, = \, \mathcal{W}proj^\alpha_i \, E^i
 \end{equation}
 into the Maurer-Cartan eq.s (\ref{carmelitano}) and using the Maurer-Cartan eq.s (\ref{birillo}), we obtain a set of 86 quadratic equations that we do not display since the system is too large. Such a system has a general solution composed of $13$ branches with respective numbers of parameters as follows:
 \begin{equation}\label{brancini}
  \{6,5,4,4,4,4,4,4,3,3,3,3,3\}
 \end{equation}
Of these solutions, we show only a few:
\begin{equation}\label{W1rest}
  \mathcal{W}_1 \, = \, \left(
\begin{array}{ccccccccc}
 0 & 0 & 0 & 0 & 0 & 0 & 0 & 0 & 0 \\
 \alpha _1 & \alpha _2 & \alpha _3 & 0 & 0 & 0
   & 0 & 0 & 0 \\
 \alpha _4 & \alpha _5 & \alpha _6 & 0 & 0 & 0
   & 0 & 0 & 0 \\
\end{array}
\right)
\end{equation}
\begin{equation}\label{W2rest}
  \mathcal{W}_2 \, = \, \left(
\begin{array}{ccccccccc}
 \alpha _1 & \alpha _2 & \alpha _3 & 0 & 0 & 0
   & 0 & 0 & 0 \\
 \alpha _4 & \frac{\alpha _2 \alpha _4}{\alpha
   _1} & \frac{\alpha _3 \alpha _4}{\alpha _1}
   & 0 & 0 & 0 & 0 & 0 & 0 \\
 \alpha _5 & \frac{\alpha _2 \alpha _5}{\alpha
   _1} & \frac{\alpha _3 \alpha _5}{\alpha _1}
   & 0 & 0 & 0 & 0 & 0 & 0 \\
\end{array}
\right)
\end{equation}
\begin{equation}\label{W3rest}
  \mathcal{W}_3 \, = \, \left(
\begin{array}{ccccccccc}
 -2 & -1 & -1 & 0 & 0 & 0 & 0 & 0 & 0 \\
 \alpha _1 & \frac{\alpha _1}{2} & \frac{\alpha
   _1}{2} & \alpha _2 & 0 & 0 & 0 & 0 & 0 \\
 \alpha _3 & \frac{\alpha _3}{2} & \frac{\alpha
   _3}{2} & \alpha _4 & 0 & 0 & 0 & 0 & 0 \\
\end{array}
\right)
\end{equation}
\begin{equation}\label{W7rest}
  \mathcal{W}_7 \, = \, \left(
\begin{array}{ccccccccc}
 1 & -1 & 0 & 0 & 0 & 0 & 0 & 0 & 0 \\
 \alpha _1 & -\alpha _1 & 0 & 0 & \alpha _2 & 0
   & 0 & 0 & 0 \\
 \alpha _3 & -\alpha _3 & 0 & 0 & \alpha _4 & 0
   & 0 & 0 & 0 \\
\end{array}
\right)
\end{equation}
\begin{equation}\label{W10rest}
  \mathcal{W}_{10} \, = \, \left(
\begin{array}{ccccccccc}
 0 & 0 & \alpha _1 & 0 & 0 & 0 & 0 & 0 & 0 \\
 0 & 0 & \alpha _2 & 0 & 0 & 0 & 0 & 0 & 0 \\
 0 & 0 & \alpha _3 & 0 & 0 & 0 & 0 & 0 & 0 \\
\end{array}
\right)
\end{equation}
Let us make a few remarks about the above displayed result. The restriction map is always a projection map. Indeed, there is always a non-trivial kernel of the homomorphism which constitutes a subalgebra of the original algebra. In the present case, we see that the kernel always contains the nilpotent generators $E^{6,7,8,9}$ and alternatively other generators
that are either nilpotent or Cartan. The features that in the $i$-th layer were associated with the generators of the kernel are projected out in the restriction map from the $i$-th to the $(i+1)$-th layer. On the other hand, we see that in all cases the nilpotent generators of the target algebra contain Cartan generators of the original larger algebra in various combinations.
\subsubsection{The general functional form of the third restriction map}
As an illustration, we have worked out the general form of the map between solvable coordinates for the case of the
$W$-matrix in eq.(\ref{W3rest}). The map is fairly simple and linear:
\begin{equation}\label{W3Phi}
  \begin{array}{rcl}
 w_1& = & \Upsilon _1+\frac{\Upsilon
   _2}{2}+\frac{\Upsilon _3}{2} \\
 w_2& = & \frac{1}{2} \left(-2 \alpha _2 \Upsilon
   _4-\alpha _1\right) \\
 w_3& = & \frac{1}{2} \left(-2 \alpha _4 \Upsilon
   _4-\alpha _3\right) \\
\end{array}
\end{equation}
It shows that in the projection the unique Cartan field of the rank $r=1$ $(i+1)$-th layer is a precise linear combination of all the Cartan fields of original maximal rank $i$-th layer and the nilpotent features of the target layer are diverse linear combination of a smaller set of nilpotent features of the $i$-th layer (in this case the smaller set is $1$-dimensional). Al the other nilpotent features are projected out.

\section{The Basic Principles of Modern Probability Theory}
\label{probaprincipe} Having clarified in the main body of the paper the goals pursued by our recollection of the basic principles of modern 
probability theory in this section we present them. 
\subsection{The Basic Notions}
\label{prodomosua}     
\subsubsection{\texorpdfstring{$\sigma$}{Sigma}-Algebras and  Probability Measures}
\begin{definizione}
\label{sigmalgeb} Given a set $\Omega$ one defines $\sigma$-algebra on $\Omega$ a family $\mathcal{A}$ of subsets $A_i \subset \Omega$ such that: 
\begin{enumerate}
  \item $\emptyset \in \mathcal{A}$ and $\Omega \in \mathcal{A}$. 
  \item If $A \in \mathcal{A}$ then its complement $A^c \equiv \Omega - A$ also belongs to the same family $A^c\in \mathcal{A}$. 
  \item If the elements of a denumerable family  of sets $\left\{ A_i\right\}_{i\in \mathbb{N}}$ belong to $\mathcal{A}$ then also their union 
      belongs to it: 
  $$ A \, = \, \bigcup_{i=1}^{\infty} A_i \, \in  \mathcal{A}$$
\end{enumerate}
\end{definizione}
\begin{definizione}
\label{spazmisur} The pair $\left(\Omega\, , \, \mathcal{A}\right)$ where $\Omega$ is a set and $\mathcal{A}$ is a $\sigma$-algebra on $\Omega$ is 
named a measurable space. 
\end{definizione}
In probability theory, the elements of the $\sigma$-algebra $X\in \mathcal{A}$ are named \textbf{events}, while the points  $p\in \Omega$ are 
named \textbf{outcomes} or \textit{experiments}. This is a very important conceptual distinction.
\begin{definizione}
\label{poweralg} One names \textbf{Part Algebra} of a set 
 $\Omega$  and denotes it with the symbol $\mathfrak{P}(\Omega)$ the set of all subsets
 equipped with the boolean algebraic operations of  \textit{union,
intersection and complement}. 
\end{definizione}
Let us now consider two sets $\Omega_1$ and $\Omega_2$; any map 
\begin{equation}\label{mappatrainsiemi}
    \phi \, : \,\Omega_1 \, \rightarrow \,\Omega_2
\end{equation}
induces a \textit{pullback map} $\phi_\star^{-1}$ on the corresponding Part Algebras 
\begin{equation}\label{pullobacco}
    \phi_\star^{-1} \, : \,\mathfrak{P}(\Omega_2) \,\rightarrow \,\mathfrak{P}(\Omega_1)
\end{equation}
One can verify that the map $\phi_\star^{-1}$ satisfies the following properties 
\begin{eqnarray}
\label{morfisema}
  \phi_\star^{-1}\left(X\bigcup Y \right) &=& \phi_\star^{-1}\left(X\right)\bigcup \phi_\star^{-1}\left(Y\right)\nonumber \\
  \phi_\star^{-1}\left(X\bigcap Y \right) &=& \phi_\star^{-1}\left(X\right)\bigcap \phi_\star^{-1}\left(Y\right)\nonumber \\
  \phi_\star^{-1}\left(X^c \right) &=& \phi_\star^{-1}\left(X \right)^c
\end{eqnarray}
Hence $\phi_\star^{-1}$ \`{e} is a morphism of boolean algebras. 
\par
Once these general concepts have been established, one can introduce the following  notion of \textit{probability measure}: 
\begin{definizione}
\label{misaproba} Let $\left(\Omega\, , \, \mathcal{A}\right)$ be a measurable space. A probability measure on $\left(\Omega\, , \, 
\mathcal{A}\right)$ is a map 
\begin{equation}\label{cuccusprobrpus}
    \mathfrak{p} \, : \, \mathcal{A} \, \rightarrow \, [0,1] \subset \mathbb{R}
\end{equation}
that satisfies the following properties: 
\begin{itemize}
  \item $\mathfrak{p}(\emptyset) \, = \, 0$ and $\mathfrak{p}(\Omega)=1$ 
  \item $\mathfrak{p}\left ( \bigcup_{i} X_i \right)\, = \, \sum_i \mathfrak{p}\left(X_i\right)$  for all denumerable unions of disjoint parts, 
      i.e. such that $X_i\bigcap X_j \, =\, \emptyset$ se $i \neq j$. 
\end{itemize}
\end{definizione}
When we have a triplet $\left(\Omega,\mathcal{A},\mathfrak{p}\right)$ we say that we have a stochastic  space and the value $\mathfrak{p}(X)$ is 
the probability of the event $X$. 
\subsubsection{Stochastic Functions and Stochastic Vectors} Let us begin with: 
\begin{definizione} If $\mathcal{T}$ is a separable topological space,  one names \textbf{Borel Algebra} of $\mathcal{T}$, denoted
$\mathcal{B}(\mathcal{T})$  the $\sigma$-algebra made by all denumerable unions,  intersections and complements of open subsets 
$U\subset\mathcal{T}$. 
\end{definizione}
In particular, on all varieties $\mathbb{R}^n$ we have the ball-topology which, for the real line $\mathbb{R}$, reduces to the topology of open 
intervals $]x,y[ \subset \mathbb{R}$, and the correspondent Borel algebra is clearly defined. Hence, using as $\sigma$-algebra the natural Borel 
algebra $\mathcal{B}(\mathbb{R})$,  we have that the pair $\left( \mathbb{R}\, ,\,\mathcal{B}(\mathbb{R})\right)$, makes a measurable space 
\par
Let us then consider a stochastic space $\left(\Omega,\mathcal{A},\mathfrak{p}\right)$ and a map: 
\begin{equation}\label{coronafine}
    \psi \, : \, \Omega \, \rightarrow \, \mathbb{R}
\end{equation}
which to any point   $p \in \Omega$  of the set $\Omega$ associates a real number (its coordinate). Because of what we discussed above, the 
pullback map 
\begin{equation}\label{eq:pullback1}
    \psi^{-1}_\star \, : \, \mathcal{B}(\mathbb{R}) \, \rightarrow
    \, \mathcal{A}
\end{equation}
is a morphism of  boolean algebras that  explicitly associates an element $X\in \mathcal{A}$ to every element of the Borel algebra of the real 
line. Thus, composing the maps we define: 
\begin{equation}\label{cornutella}
    \mathfrak{p}_{\psi} \, \equiv \, \mathfrak{p}\circ\psi^{-1}_\star
\end{equation}
which is a map from the  Borel algebra of the real line to the interval $[0,1]$: 
\begin{equation}\label{cromatogno}
    \mathfrak{p}_\psi \, : \, \mathcal{B}\left(\mathbb{R}\right) \, \rightarrow \,
    [0,1]
\end{equation}
This is what we name a \textbf{stochastic function}. In practice  to every open interval $]x,y[$ the stochastic function associates a number 
between $0$ ed $1$  which is the probability that while doing a measuring experiment the measured value happens to be in the considered interval. 
In this way one can consider stochastic functions that are discontinuous, step-wise and the like, yet they are Lebesgue integrable thanks to the 
measurability of the support space. 
\subsubsection{Probability Density} An interesting case is when the stochastic function can be described in 
terms of a probability density given by an integrable function $\rho_\psi(q)$ on the real line such that: 
\begin{equation}\label{coriaceo}
    \mathfrak{p}_\psi\left(\left[a,b\right]\right) \, =
    \,\int_{a}^{b} \rho_\psi(q) \,dq
\end{equation}
For the probability density to be well defined, it is necessary that the probability density $\rho_\psi(q)$ be properly normalized: 
\begin{equation}\label{ranierus}
    \int_{-\infty}^{+\infty} \rho_\psi(q) \,dq \, = \, 1
\end{equation}
Under these conditions, one can define the average value of any function $f(q)$ of the  stochastic variable $q \in \mathbb{R}$ writing 
\begin{equation}\label{romildo}
    \langle f\rangle \, \equiv \,  \int_{-\infty}^{+\infty}\, f(q)\, \rho_\psi(q) \,dq \,
\end{equation}
\subsubsection{Stochastic Vector} In a similar way we can define stochastic vectors. 
\par
Consider a finite dimensional vector space $\mathbb{V}$: 
\begin{equation}\label{lollobrigga}
    \mathrm{dim}_\mathbb{R} \,\mathbb{V}\, = \, r < \infty
\end{equation}
and a basis $\mathbf{e}_i$ ($i=1\dots,r$) of vectors such that 
\begin{equation}\label{contellus}
    \forall \mathbf{X} \in \mathbb{V} \, : \, \mathbf{X} \, =\,
    \sum_{i=1}^r X^i(\chi)\mathbf{e}_i
\end{equation}
Where the components of the vector are thought of as functions of $\chi \in \Omega$,  the space of outcomes over which we defined the probability 
measure.  By reasoning entirely analogous to that above, each component $X^i(\chi)$ can be thought of as a probability density $X^i_\psi\left( 
\mathbf{q}\right)$ on a space $\mathbb{R}^n$ where $n$ is the number of coordinates necessary to identify a point $\chi\in\Omega$, namely the 
dimension of the set $\Omega$, if this latter can be thought of as a differentiable manifold. As in the previous case what we are constructing is, 
for each compenent $w^i$ of the stochastic vector, a map: 
\begin{equation}
    \psi^i \, : \, \Omega \, \rightarrow \, \mathbb{R}^n
\end{equation}
which, by pullback, induces a map: 
\begin{equation}\label{eq:pullback2}
    \psi^{-1|i}_{\star} \, : \, \mathcal{B}(\mathbb{R}^n) \, \rightarrow
    \, \mathcal{A}
\end{equation}
By composition of maps we obtain 
\begin{equation}\label{scortabella}
    \mathfrak{p}_{\psi^i} \, \equiv \, \mathfrak{p}\circ\psi^{-1|i}_\star
\end{equation}
which is a map from the Borel algebra  of $\mathbb{R}^n$ to the interval $[0,1]$: 
\begin{equation}\label{cromatogno1}
    \mathfrak{p}_{\psi^i} \, : \, \mathcal{B}\left(\mathbb{R}^n\right) \, \rightarrow \,
    [0,1]
\end{equation}
In this way we have defined  a stochastic vector, namely a map: 
\begin{equation}\label{coronaspessa}
    \pmb{\Psi} \, : \, \Omega \, \rightarrow \, \mathbb{V}
\end{equation}
Also, for  stochastic vectors, the most favorable and smooth situation occurs when each of the vector components is substituted by an integrable 
probability density: 
\begin{equation}
    \mathbf{X}(\mathbf{q}) \, = \, \sum_{i=1}^r \,
    \rho_\Psi^i(\mathbf{q})\,  \mathbf{e}_i \quad ; \quad
    \int\int\dots\int_{\mathbb{R}^n}\, \rho_\Psi^i(\mathbf{q}) \,
    \underbrace{d^n\mathbf{q} }_{\equiv d\mu(\mathbf{q})}\, = \, 1
\end{equation}
where with  $\mathrm{d}\mu(\mathbf{q})$ we have denoted  the integration measure on the space $\Omega$ which might be more elaborate and contain a 
factor $\sqrt{\text{det}g}$ when $\Omega$ is a Riemannian manifold endowed with a metric. 

\section{A brief history of the logistic regression and of the sigmoid}
\label{historsigma} We provide here a historical note on the origin of the logistic function and 
of the utilized names, which is always an inspiring starting point for understanding. 
 \paragraph{Historical Note.}
 Why the name \textbf{regression} for a statistical optimization? It appears that such a name is due to Francis Galton (1822-1911), the inventor of eugenetics, who, studying the distribution through generations of human statures, or the size of vegetable seeds, noticed and explained the phenomenon of the \textbf{regression of deviations towards the mean}. The qualifier \textbf{logistic} was instead invented (for not too clear reasons) by the Belgian mathematician Pierre Francois Verhulst (1804-1849), who studied the evolution of population and in 1838 arrived at the following differential equation:
 \begin{equation}\label{populazia}
   \mathcal{P}'(t)-r \mathcal{P}(t) \left(1-\frac{\mathcal{P}(t)}{K}\right)\, = \, 0
 \end{equation}
 for the function $\mathcal{P}(t) $  expressing the \textit{the magnitude of population} at each time $t$. The two constants
 appearing in (\ref{populazia}) are respectively $r$, the rate of growth, and $K$, the maximum number of individuals that the environment can support. The general integral of eq.(\ref{populazia}) involves an extra integration constant $\mathcal{P}_0$ that can be interpreted as the magnitude of population at time $t=0$. The integral is as follows:
 \begin{equation}\label{plebaglia}
  \mathcal{P}(t) \, = \, \frac{K}{\frac{\left(K-P_0\right) e^{-r t}}{P_0}+1} \, = \, K \,\frac{1}{1+\exp\left[-r \, t \, + \, \log\left[\frac{K-\mathcal{P}_0}{\mathcal{P}_0}\right]\right]}
 \end{equation}
which can be rewritten as we display below:
 \begin{eqnarray}\label{karletto}
   \frac{\mathcal{P}(\mathbf{x})}{K} & = & \sigma(r\,t +b) \quad ; \quad {b} \, = \, - \, \log\left[\frac{K-\mathcal{P}_0}{\mathcal{P}_0}\right]\nonumber\\
   \sigma(x) & \equiv & \frac{1}{1\, + \, e^{-x}} \quad \text{the sigmoid function} 
   \end{eqnarray}

\newpage
\bibliography{allesbiblio}
\bibliographystyle{ieeetr}
\end{document}